\documentclass[lettersize,journal]{IEEEtran}
\usepackage{amsmath,amsfonts}
\usepackage{algorithmic}
\usepackage{array}
\usepackage[caption=false,font=normalsize,labelfont=sf,textfont=sf]{subfig}
\usepackage{textcomp}
\usepackage{stfloats}
\usepackage{url}
\usepackage{verbatim}
\usepackage{graphicx}
\def\BibTeX{{\rm B\kern-.05em{\sc i\kern-.025em b}\kern-0.08em
    T\kern-.1667em\lower.7ex\hbox{E}\kern-.125emX}}
\usepackage{balance}
\usepackage{multirow}
\usepackage{amssymb}
\usepackage{cite} 
\usepackage{threeparttable}
\usepackage[figuresright]{rotating}
\usepackage{xcolor}
\usepackage{makecell}
\usepackage[numbers]{natbib}
\usepackage{color}
\usepackage{pifont}
\usepackage{booktabs}
\newcommand{\etal}{\textit{et al.}}

\begin{document}
\title{A Survey on Physical Adversarial Attack in Computer Vision}
\author{ 
\IEEEauthorblockN{Donghua Wang$^{1,2}$, Wen Yao$^{2*}$, Tingsong Jiang$^{2*}$, Guijian Tang$^{2,3}$, Xiaoqian Chen$^{2}$}

\thanks{*Wen Yao and Tingsong Jiang are the corresponding authors.}

\IEEEauthorblockA{$^1$ College of Computer Science and Technology, Zhejiang University, Hangzhou, China} \\

\IEEEauthorblockA{$^2$ Defense Innovation Institute, Chinese Academy of Military Science, Beijing, China} \\

\IEEEauthorblockA{$^3$ College of Aerospace Science and Engineering, National University of Defense Technology, Changsha, China}

\IEEEauthorblockA{wangdonghua@zju.edu.cn, wendy0782@126.com, tingsong@pku.edu.cn, tangbanllniu@163.com, chenxiaoqian@nudt.edu.cn}
}

\markboth{Journal of \LaTeX\ Class Files,~Vol.~18, No.~9, September~2022}%
{How to Use the IEEEtran \LaTeX \ Templates}

\maketitle

\begin{abstract}
Over the past decade, deep learning has revolutionized conventional tasks that rely on hand-craft feature extraction with its strong feature learning capability, leading to substantial enhancements in traditional tasks. However, deep neural networks (DNNs) have been demonstrated to be vulnerable to adversarial examples crafted by malicious tiny noise, which is imperceptible to human observers but can make DNNs output the wrong result. Existing adversarial attacks can be categorized into digital and physical adversarial attacks. The former is designed to pursue strong attack performance in lab environments while hardly remaining effective when applied to the physical world. In contrast, the latter focus on developing physical deployable attacks, thus exhibiting more robustness in complex physical environmental conditions. Recently, with the increasing deployment of the DNN-based system in the real world, strengthening the robustness of these systems is an emergency, while exploring physical adversarial attacks exhaustively is the precondition. To this end, this paper reviews the evolution of physical adversarial attacks against DNN-based computer vision tasks, expecting to provide beneficial information for developing stronger physical adversarial attacks. Specifically, we first proposed a taxonomy to categorize the current physical adversarial attacks and grouped them. Then, we discuss the existing physical attacks and focus on the technique for improving the robustness of physical attacks under complex physical environmental conditions. Finally, we discuss the issues of the current physical adversarial attacks to be solved and give promising directions.
\end{abstract}

\IEEEpeerreviewmaketitle

Deep neural networks (DNNs) have demonstrated substantial achievements across a broad spectrum of research areas, including but limited to computer vision \cite{simonyan2014very,he2016deep,carion2020end,liu2021swin}, natural language processing \cite{vaswani2017attention,devlin2018bert}, and speech recognition \cite{amodei2016deep,gulati2020conformer}. Applications with the core of these advanced techniques have been widely deployed in the real world, such as facial payment, medical image diagnosis, autonomous driving, and so on. However, DNNs have exposed their security risks to adversarial examples crafted by small noises that are invisible to humane observers but can mislead the DNNs to output wrong results. For instance, maliciously devised eyeglasses frames can fool the facial recognition system \cite{sharif2016accessorize}; the automatic driving car is susceptible to environmental changes (e.g., shadow or brightness), resulting in out-of-control \cite{kong2020physgan} or misdetection\cite{zhang2019camou,wang2021dual,wang2022fca}. Such instability factors impose a potential risk for those deployed DNN-based systems in the real world, especially for security-sensitive applications.

\begin{figure}[t]
	\centering
	\begin{minipage}{.5\linewidth}
		\centering
		\includegraphics[width =1\linewidth]{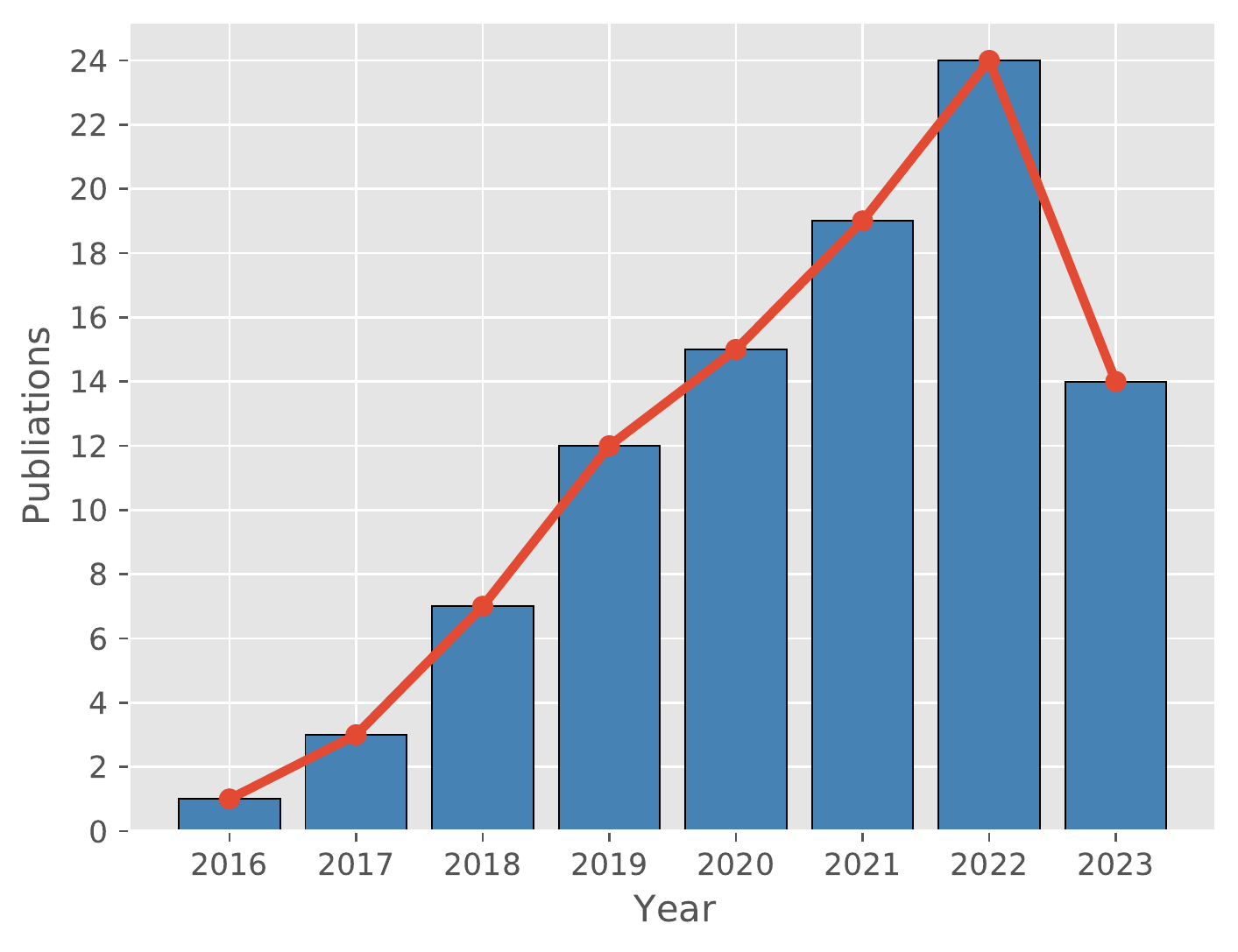}
	\end{minipage}
	\caption{Count of physical attack literature published yearly.}
	\label{fig:paper_count}
\end{figure}

Since Szegedy \textit{et al.} \cite{szegedy2014intriguing} first discovered that the DNN-based image classifier is susceptible to malicious devised noise, a line of works \cite{fgsm2015explaining,bim2016adversarial,pgd2018towards,mim2018BoostingAA,carlini2017towards,Huang2020BlackBox} have been proposed to explore the adversarial examples and the corresponding countermeasure techniques in different research areas, especially in computer vision. Accordingly, many surveys \cite{sun2018survey,wiyatno2019adversarial,qiu2019review,yuan2019adversarial,serban2020adversarial,ren2020adversarial} were proposed to the taxonomy or summarize the development of the adversarial examples. Serban \textit{et al.} \cite{serban2020adversarial} made taxonomy exhaustively for attack and systemically summarized the existing attack methods from machine learning to DNNs. \cite{yuan2019adversarial,zhang2019adversarial,ren2020adversarial,hu2021artificial,aldahdooh2022adversarial} reviewed the development of adversarial attack and defense methods against DNNs at different periods. Some surveys only focus on the specific task, such as image classification \cite{machado2021adversarial}, adversarial patch \cite{sharma2022adversarial}. However, these surveys mainly provide an overall view of the adversarial attack and defense in the specific research area or a roughly overall perspective. Moreover, with the increasingly maturing of DNNs in commercial deployment, exploring the physical attack is more urgent for enhancing the robustness of DNN-based systems. Although some works have reviewed the development of physical attacks \cite{ren2020adversarial,ren2021adversarial}, they are out-of-date as plenty of novel physical attacks have emerged \cite{wang2021universal,wang2021dual,zhu2021fooling,zhu2022infrared,wang2022fca,duan2022learning,suryanto2022dta,pavlitskaya2022feasibility,nemcovsky2022physical,doan2022tnt,wei2022adversarial} in the past two years. Concurrent to our survey, \cite{wei2022physically,wei2022physical} only review 43 and 46 literature the physical attack, respectively, ignoring some literature about physical adversarial attacks. In contrast, we review 95 literature on physical attacks from 2016 to 2023 (until July). Figure \ref{fig:paper_count} illustrates the number of physical attack literature published yearly. Therefore, it is necessary to analyze the newly proposed method for better tracking the latest research direction.

In this survey, we comprehensively review the development of existing physical adversarial attacks in the computer vision task. Given the statistical result of the current literature, the survey of physical attacks is divided into three-fold: image recognition, object detection, and others. This is because current physical adversarial attack algorithms mainly focus on image recognition and object detection, while semantic segmentation, object tracking, and so on are less explored. Specifically, we first propose a taxonomy to categorize physical attack literature in terms of each process of performing physical attacks (i.e., determining the {\bf Victim Model}, {\bf Problem Modeling}, {\bf Optimization}, {\bf Deployment} of physical attack, and {\bf Evaluation}). Then, we detailed discuss different physical adversarial attacks in image recognition, object detection, and others. Finally, we discuss the issues to be solved of physical adversarial attacks at present and the potential worthy research direction. We hope this survey will provide comprehensive retrospect and beneficial information on physical adversarial attacks for researchers.

To cover as many existing physical attacks as possible, we used the keywords ``physical attack" or ``physical adversarial examples" to search for the latest paper from Google Scholar. In addition, we collected the papers from conference proceedings (e.g., CVPR, ICCV, ECCV, and so on) in terms of the keyword and title. Then, we manually checked the references in all selected papers to avoid ignorance of the relevant literature. For completeness, we also briefly introduce the works that do not involve physical attacks but are closely related to them. In addition, we excluded papers that have not relevant to physical adversarial attacks on computer vision. Finally, we gathered 95 physical adversarial attack papers in computer vision. 

In summary, our main contributions are listed as follows.

\begin{itemize}
	\item We comprehensively survey physical adversarial attacks in computer vision tasks, providing an overview and evolvement of this research field.
	\item We propose a taxonomy to categorize the existing literature on physical adversarial attacks according to each process of physical attacks. We discuss different physical attacks and highlight the most frequently used techniques for improving the robustness of physical attacks.
	\item We discuss the challenging and valuable future research directions of physical adversarial attacks in the following aspects: realistic-orient input transformation (e.g., scale change, occlusion), practical-orient attack algorithm (e.g., black-box attack, improving transferability, and attack multi-modal and multi-task), strict and uniform evaluation criterion.
\end{itemize}

The following paper is organized as follows. The background knowledge of adversarial examples and the deep neural network is briefly introduced in Section \ref{sec:preliminarires}. The physical adversarial attacks against the image recognition, object detection and other task are discussed in Section \ref{sec:classify}, Section \ref{sec:object_det}, and Section \ref{sec:other}, respectively. In Section \ref{sec:disccus}, we discuss the challengings to be solved and then provide some possible research directions for constructing better physical adversarial examples. Finally, Section \ref{sec:conclude} conclude this survey.


\begin{table}
\centering
\caption{Terminology}
\begin{tabular}{ll}
\hline
				& \MakeUppercase{Terminology} \\ \hline
$x$     	 	& Clean image.    \\
$y$				& Ground truth of the clean image.\\
$\hat{x}$       & Clean image in the physical world.       \\
$x_{adv}$      	& Adversarial example.      \\
$\hat{x}_{adv}$  & Adversarial example in the physical world.        \\
$\delta$		& Adversarial perturbation. \\
$\epsilon$		& Maximum allowable magnitude of perturbation. \\
$f$		        & Neural network.       \\
$J_f(\cdot)$    & Loss function of the f.       \\
$\nabla J_f(\cdot)$ & The gradient of the $f$. \\
$\|\cdot\|_p$   & $L_p$ norm distance.        \\
$d(\cdot, \cdot)$ & Distance metrics.        \\
$\mathcal{T}$   & Transformation distribution.      \\
$t$             & A specific transformation.      \\ \hline
\end{tabular}
\label{tab:terminology}
\end{table}

\section{Background knowledge}
\label{sec:preliminarires}
In this section, we briefly introduce the foundation concept of deep learning and adversarial examples. Table \ref{tab:terminology} lists the terminology used in this paper.

\subsection{Deep learning} 

Deep learning is known for its powerful capability to extract data patterns and features automatically from enormous data, promoting the development of many traditional tasks, such as automatic driving \cite{sun2020scalability}, and drug design \cite{jumper2021highly}. Deep neural networks (DNNs) are one of the most substantial components of deep learning, which realizes feature extraction from data and decision-making in an end-to-end manner. The architecture of DNN is highly determined by the task; for instance, recurrent neural network (RNN) is adopted for speech recognition \cite{amodei2016deep}. Nonetheless, convolutional neural networks (CNNs) have been the most commonly used architecture in computer vision tasks in the past decade and have been maturely deployed in commercial applications. The representative architecture of CNN in computer vision like ResNet \cite{he2016deep}, VGG \cite{vgg16}, and DenseNet \cite{densenet}, which are employed in image recognition tasks and object detection tasks \cite{redmon2016you,he2016deep}. Recently, the tremendous success of transformer architecture \cite{vaswani2017attention,devlin2018bert} in Natural Language Processing (NLP) tasks has inspired the computer vision community to develop vision-specific variants of the transformer, referred to as the Vision Transformer \cite{dosovitskiy2020image,liu2021swin}. The Vision transformer comprises an encoder and a decoder, which consists of multiple transformer blocks with the same architecture. The encoder projects the inputs to latent feature representations, which are then processed by the decoder to produce the task-specific output, such as the category prediction for a classification task. Despite the differences in architecture across various networks, they share a uniform representation function, i.e., $f(x)$, where the concrete representation of $f(x)$ depends on the given task. However, despite DNNs achieving great success, the underlying mechanism of DNNs is unclear and exposes potential security risks \cite{szegedy2014intriguing}.

\subsection{Adversarial examples}

The definition of the adversarial examples against DNNs was first introduced by Szegedy \textit{et al.} \cite{szegedy2014intriguing}, which refers to a sample containing elaborately crafted malicious noise that is invisible to human observers but can mislead the DNN to output the wrong result. Specifically, for a well-trained network $f$, the adversary generate a adversarial example $x_{adv}$ for a given image $x$ by solving the following problem \cite{yuan2019adversarial}

\begin{equation}
{\rm \min} ~~ \|{x_{adv} - x}\|_p  ~~~ s.t. ~~ f(x_{adav}) \neq f(x),  \\
\label{eq:adv}
\end{equation}
where the $\|\cdot\|_p$ is the $L_p$-norm, $\epsilon$ controls the maximum allowable magnitude of $\delta$ ($\delta = x_{adv} - x$) under the constrain of $L_p$-norm. The common choice of $p$ is 0, 2, and $\infty$. $L_0$ confines the maximum modification pixel number of the clean image, $L_2$ constraints the maximum modification of Euclidean distance between the $\delta$ and zero,  and $L_\infty$ bounds the maximum value of $\delta$. 

Research of adversarial examples in the digital world has long been explored, and a line of works \cite{fgsm2015explaining,pgd2018towards,mim2018BoostingAA,li2023bayesian,sun2023multi,li2023adaptive} has proposed. However, adversarial examples optimized for the digital world may be inefficient in the physical world due to the huge gap between the training and testing environment caused by various factors, which include printed color distortion due to the mismatch between the optimized color and printable color, distortion caused by the environment light changing, and deformability during recapture the adversarial example. Note that, despite some works attempting to attack the online employed DNN-based systems \cite{xie2019improving}, we do not discuss these works as we are focused on the attacks that can apply to the real world. Based on the above discussion, developing effective physical adversarial attacks are more challenging than digital attacks.

\section{Taxonomy of physical adversarial examples}
\begin{figure*}[ht]
	\centering
	\begin{minipage}{.7\linewidth}
		\centering
		\includegraphics[width =1\linewidth]{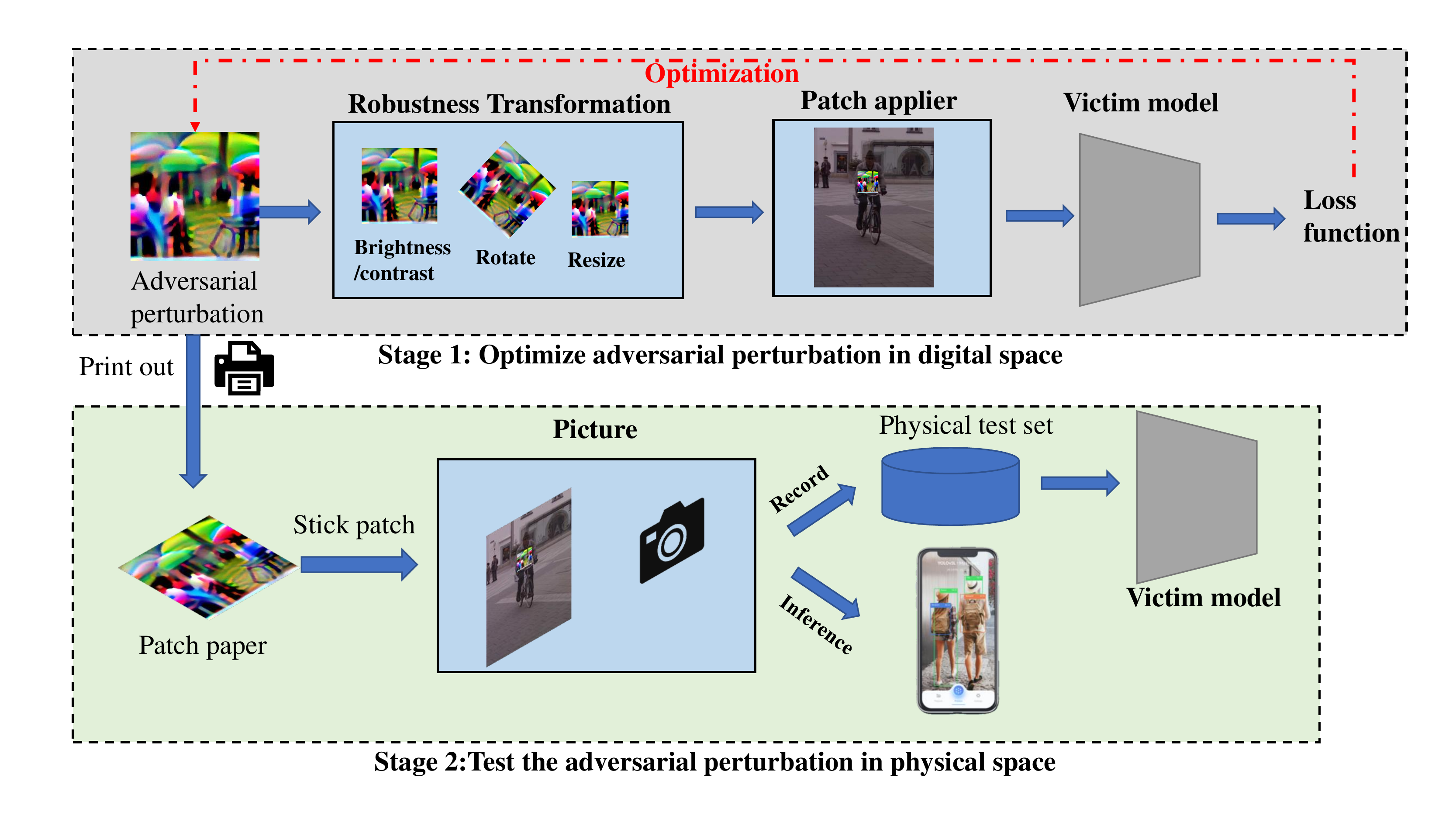}
	\end{minipage}
	\caption{Overview of the physical adversarial attack process.}
	\label{fig:overview}
\end{figure*}

\begin{figure*}[h]
	\centering
	\begin{minipage}{0.7\linewidth}
		\centering
		\includegraphics[width =1\linewidth]{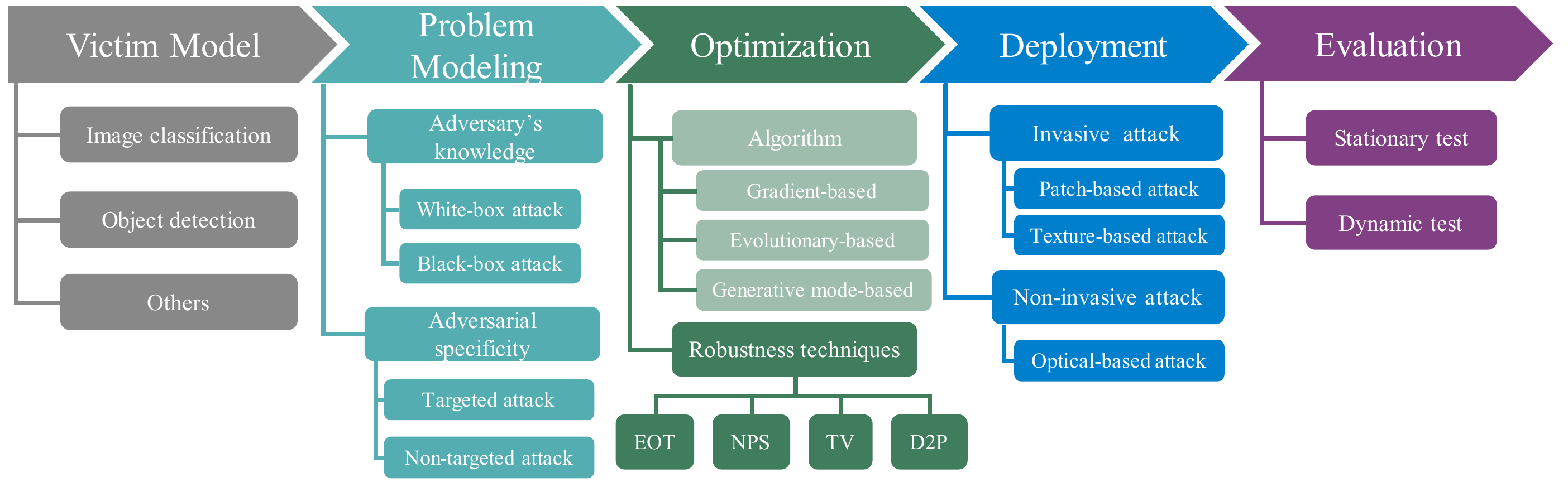}
	\end{minipage}
	\caption{Taxonomy in terms of each process of physical adversarial attacks.}
	\label{fig:taxonomy}
\end{figure*}

The common process of physical adversarial attacks in computer vision tasks is described as follows: first, optimize the adversarial perturbation in the digital world. Then, prepare the adversarial perturbation and paste/paint it over the target object, e.g., printing out the adversarial patch with the color printer and then hanging it on the object. Finally, place the adversarial perturbed object toward the deployed sensor device (e.g., RGB camera sensor) to perform attacks. Figure \ref{fig:overview} provides an overview of the whole process of physical attacks. Accordingly, in this section, we propose a taxonomy criterion to systematically categorize the existing literature on physical adversarial attack methods. We categorize the current literature in terms of each process of the physical adversarial attack. Specifically, when we attempt to attack a DNN system deployed in the real world, we have to determine the following question: \ding{182} Which model do you plan to attack? ({\bf Victim Model}) \ding{183} How much information do you have about the victim model, and what's the expected behavior of the adversarial attack? ({\bf Problem Modeling}) \ding{184} How to optimize the adversarial perturbation? and What robustness technique is introduced? ({\bf Optimization}) \ding{185} How to perform the physical adversarial attacks? ({\bf Deployment}) \ding{186} How to evaluate the attack performance? ({{\bf Evaluation}}). Figure \ref{fig:taxonomy} depicts the detailed taxonomy. Note that in this survey, we focus on robustness techniques for improving the physical attack performance; thus, we ignore the discussion of algorithm in {\bf Optimization}. In the following, we will introduce each category displayed in Figure \ref{fig:taxonomy} in detail.

\subsection{Victim Model}
The victim model refers to the DNN to be attacked that is determined by the adversary first. According to the statistical result of the collected literature on physical adversarial attacks against vision tasks, we group the victim model into three-fold: image recognition, object detection, and others. The detailed description is as follows.

\subsubsection{Image recognition} 
Image recognition is a conventional computer vision task that classifies the input image into a specific class. For a well-trained DNN-based image classifier $f$, one can get the output  $f(x)=y_{cls}$ for given an image $x$, where $y_{cls}=y$, $y_{cls}$ and $y$ is the predicted class and ground truth of the input image, respectively. Common image recognition task includes general image classification \cite{ifgsm2018adversarial}, face recognition (FR) \cite{sharif2016accessorize,yang2023towards}, license plate recognition (LPR) \cite{qian2020spot}, and traffic sign recognition (TSR) \cite{eykholt2018robust,wang2023rfla}. 

\subsubsection{Object detection}
Object detection (OD) is another significant computer vision task designed to locate the object in the image and simultaneously give the class and confidence of the located object. For a well-trained DNN-based object detector $f$, one can get the output $f(x)=y_{obj}$ for given an image $x$, where $y_{obj}$ contain three outputs, including the bounding box (composed of two coordinates), the confidence score, and category. Furthermore, object detection is also extended to other image modalities, such as infrared image object detection (IR OD) \cite{zhu2021fooling} and aerial image object detection (AI OD) \cite{du2022physical}.

\subsection{Others}
With the increasing attention raised on physical adversarial attacks, more vision tasks have reported their vulnerability to adversarial attacks in the physical world. However, the corresponding literature is less compared to image recognition and object detection tasks.  Therefore, we group the physical adversarial attack beside the image recognition and object detection task into others. For example, the attacks include but are not limited to semantic segmentation (SS) \cite{nesti2022evaluating}, object tracking (OT) \cite{wiyatno2019physical,ding2021towards}, re-identification(ReID) \cite{wang2019advpattern}, steering model of automatic system \cite{kong2020physgan}, monocular depth estimation (MDE) \cite{cheng2022physical}, and prohibited item detection \cite{liu2023x}.

\subsection{Problem Modeling}
After determining the victim model, the adversary has to model the problem in terms of the available information of the target model (adversary's knowledge) and decide their requirements (adversarial specificity). The former can be further divided into white-box attacks and black-box attacks. The latter can be divided into targeted and non-targeted attacks. Both of them will be briefly introduced as follows.

\subsubsection{Adversary's knowledge}
\textbf{White-box attacks.} White-box attacks assume that the adversary can access the full knowledge about the target victim network \cite{fgsm2015explaining,bim2016adversarial}, including its architecture, parameters, and training data. In white-box settings, the adversary is permitted to exploit the gradient with respect to input images of the target victim model to generate adversarial examples \cite{fgsm2015explaining}. As a result, white-box attacks are currently the most effective type of adversarial attack \cite{athalye2018obfuscated}.

\textbf{Black-box attacks.} Black-box attacks hypothesize that the adversary has no knowledge about the target victim network but can query the target model \cite{papernot2017practical}. That makes black-box attacks more challenging than white-box attacks \cite{li2022approximate}. Nonetheless, there are some algorithms to perform black-box attacks, such as genetic algorithm \cite{alzantot2019genattack}, particle swarm optimization (PSO) \cite{wang2023rfla}, and differential evolution \cite{li2022approximate,sun2023differential}, or exploit the transferability of adversarial examples \cite{mim2018BoostingAA,wang4332122advops}. 

\subsubsection{Adversarial specificity}
\textbf{Non-targeted attacks.} Non-targeted attacks are designed to fool the target victim model to produce the prediction except the original class \cite{szegedy2014intriguing}. Intuitively, non-target attacks are prone to search the adversarial example along the direction of the nearest decision boundary to the original sample \cite{carlini2017towards}, which makes it can quickly obtain the adversarial examples. 

\textbf{Targeted attacks.} Targeted attacks are tailored to mislead the target victim model to output the specific class assigned by the attacker \cite{papernot2016limitations}. Although object detection has multiple outputs (i.e., bounding boxes, classes, and the confidence score of the detected object), the targeted attack only refers to misleading the detector to produce the specific class. In optimization, targeted attacks hardly search adversarial examples successfully when the relationship between the original and target classes in the high-dimensional decision boundary is complicated \cite{zhao2021success}. Therefore, targeted attacks are more challenging than non-targeted \cite{moosavi2016deepfool,chen2017zoo,Narodytska2017SimpleBA}.

\subsection{Optimization}
Optimization of physical deployable adversarial perturbation involves optimization algorithms and robustness techniques. The former focus on making the perturbation aggressive, and the mainstream method includes gradient-based, generative model-based, and evolutionary-based methods, which have been exhaustly discussed in previous works \cite{wang2020adversarial,hu2021artificial}. Thus, we would not discuss the optimized algorithm. In contrast, the latter concentrates on making perturbation robustness to the complex environments in the real physical world, which has been less explored in previous surveys. Here, we introduce several widely used robustness techniques for improving the physical attack performance, including expectation over transformation (EOT), non-printability score (NPS) loss, total variant (TV) loss, and digital-to-physical (D2P) modeling, which will be introduced in details.

\subsubsection{Expectation Over Transformation (EOT)}

\begin{table}[t]
\centering
\caption{Transformation distribution reported in \cite{chen2018shapeshifter,eykholt2018robust,thys2019fooling,huang2020universal}.}
\begin{tabular}{lll}
\hline
Transform   & Parameters 					& Remark      \\ \hline
Affine      &  $\mu=0, \sigma=0.1$          & Perspective/Deformed Transforms \\ 
Rotation    &  $-15^\circ \sim 15^\circ$    & Camera Simulation \\
Contrast    &  $0.5 \sim 1.5$               & Camera Parameters \\
Scale       &  $0.25 \sim 1.25$             & Distance/Resize \\ 
Brightness  &  $-0.25 \sim 0.25$            & Illumination \\ 
Translation &  $-0.04 \sim 0.04$            & Pattern Location \\ 
Cropping    &  $-0.7 \sim 1.0$              & Photograph/Occlude Simulation  \\
Random Noise&  $-0.15 \sim 0.15$            & Noise \\ \hline

\end{tabular}
\label{tab:transform}
\end{table}

EOT is a general framework for improving the adversarial robustness of physical attack on a given transformation distribution $\mathcal{T}$ \cite{athalye2018synthesizing}. Essentially, EOT is a data augmentation technique for adversarial attacks, which takes potential transformation in the real world into account during the optimization, resulting in better robustness. Table \ref{tab:transform} provides the widely used transformations reported in \cite{chen2018shapeshifter,eykholt2018robust,thys2019fooling,huang2020universal}. Specifically, instead of utilizing the norm of $x - x^{'}$ to bound the solution space, EOT endeavors to impose constraints on the expected effective distance between the adversarial and original inputs under a transformation distribution, which is expressed

\begin{equation}
	\delta = \mathbb{E}_{t \sim \mathcal{T}}[d(t(x_{adv}), t(x))].
\end{equation} 
Recently, Miao \etal \cite{dong2022isometric} pointed out EOT can not cover all select transformations, making the generated adversarial examples not robust. They address this issue by finding a subset of worst-cast (i.e., most harmful) transformations, which could cover all weaker transformations.

In addition, some approaches \cite{wang2021dual,wang2022fca,mathov2022enhancing,yang2022controllable} leverage the physical render to optimize the mesh or texture of the 3D object, allowing them can sample images from varying viewpoints. Therefore, the table of this survey categorizes the methods that used the above transformation as EOT for simplicity.

\subsubsection{Non-Printability Score (NPS)}
\begin{figure}[t]
	\centering
	\begin{minipage}{.4\linewidth}
		\centering
		\includegraphics[width =1\linewidth]{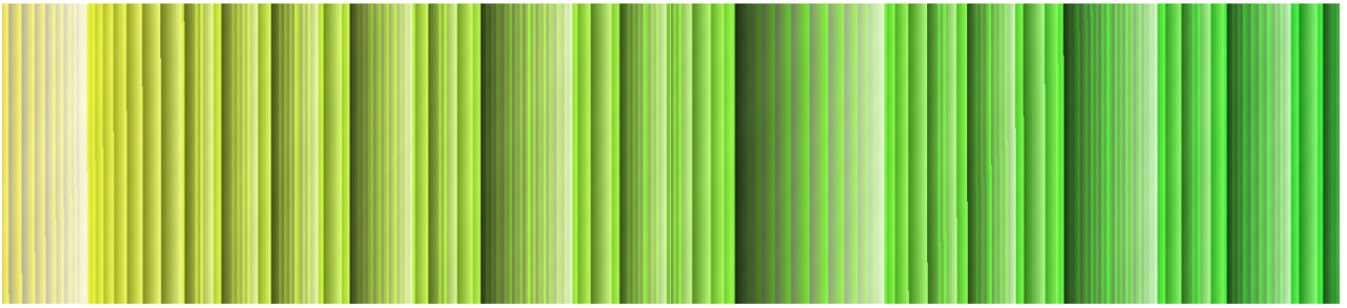}
		\centerline{\footnotesize (a) Digital Image}
	\end{minipage}
	\begin{minipage}{.4\linewidth}
		\centering
		\includegraphics[width =1\linewidth]{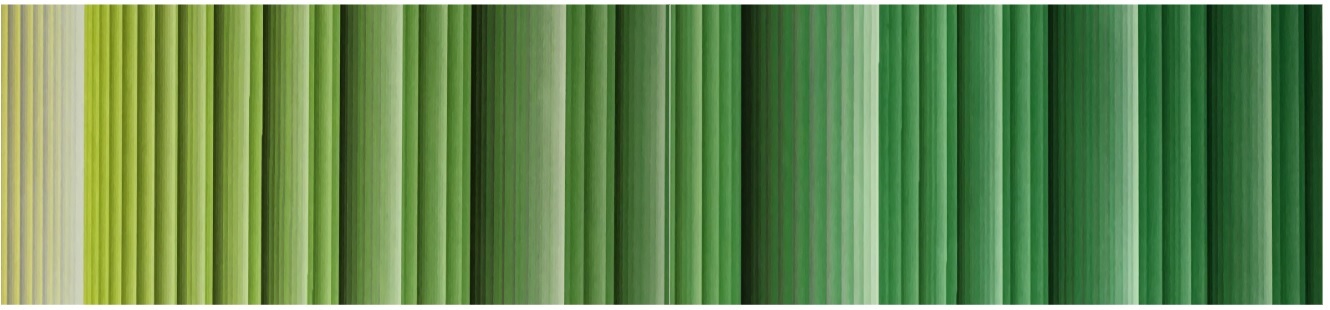}
		\centerline{\footnotesize (b) Printed Image}
	\end{minipage}
	\caption{Visualization discrepancy between the digital image (a) and its printed images (b).\cite{song2018physical}}
	\label{fig:nps}
\end{figure}

In most cases, printing the adversarial perturbation is a necessary step in conducting physical attacks. Thus, the perturbation will inevitably suffer some distortion caused by the printer, resulting in attacks failing in some cases. For instance, Figure \ref{fig:nps} illustrates the digital color and its printed result. To address this issue, non-printability score (NPS) loss \cite{sharif2016accessorize} is devised as a metric to measure the color distance between optimized adversarial perturbation and the common printer. In general, the NPS loss can be represented as the following loss

\begin{equation}
L_{nps} = \sum_{p_{(i,j)} \in x_{adv}} \min_{c_{print} \in C} \lvert p_{(i,j)} - c_{print}\rvert,
\end{equation}
where the $p_{(i,j)}$ is a pixel at coordinate $(i,j)$ in the adversarial example and $c_{print}$ is a color triplet in a set of printable colors $C$. Under the constraints of NPS loss, the optimized adversarial perturbation will gradually close to the predefined printable colors. Recently, \cite{wang2021daedalus} improved NPS loss by adopting a subsampled (0.1\% ~1\% ratio of the original pixel) non-printability score strategy, accelerating the compute efficiency on the large image without loss performance.

\subsubsection{Total variation norm (TV loss) }
Natural images featuring smoothness, in which the color of the patch is progressive and consistent \cite{mahendran2015tvloss}. Hence, finding smooth and consistent perturbations can enhance the plausibility of physical attacks. In addition, extreme differences in perturbation across adjacent pixels are implausible to be exactly captured by cameras due to the sampling noise. As a result, non-smooth perturbations are hard to physically realizable \cite{sharif2016accessorize}. To address the above issues, total variation (TV) \cite{mahendran2015tvloss} loss is presented to maintain the smoothness of perturbation. For a perturbation $\delta$, TV loss is defined as 

\begin{equation}
TV(\delta) = \sum_{i,j}((\delta_{i,j} - \delta_{i+1,j})^2 + (\delta_{i,j} - \delta_{i,j+1})^2)^{\frac{1}{2}}.
\end{equation}
Minimizing $TV(\delta)$ would make the values of neighborhood pixels even, producing smoother perturbation, and vice versa \cite{sharif2016accessorize}. Therefore, under the constraint of $TV(\delta)$ loss, the perturbation is gradually smooth and suitable for physical realizability. Recently, \cite{singh2022powerful} improved the TV loss by only considering the discrepancy of pixel pairs greater than a specific threshold, alleviating the constraints during optimization and resulting in better convergence.

\subsubsection{Digital-to-Physical (D2P)}
In some cases, EOT cannot cover all physical world transformations, such as the discrepancy between the digital image and the corresponding captured physical image caused by environmental factors. Thus, some researchers \cite{jan2019connecting} proposed to 
train a model to learn the potential physical transformation of the digital and physical image pair (namely the D2P model), which is then used to post-process the adversarial examples during optimization to improve its robustness. D2P has been shown the effectiveness in improving the physical robustness of adversarial perturbation. Therefore, D2P can be regarded as an effective technique to boost the physical robustness of adversarial attacks.

\subsection{Deployment}
Once the optimized adversarial perturbation is obtained, the adversary needs to perform physical adversarial attacks with it. In this stage, according to whether it requires the adversary to approach and modify the target object or not at the deployment stage of physical attacks, the physical attacks deployment can be categorized into invasive attacks \cite{zhong2022shadows} and non-invasive attacks. 

\subsubsection{Invasive attacks} 
Invasive attacks require the attacker to approach the target object and modify its appearance with adversarial perturbation, which can be further grouped into patch-based and texture-based attacks according to the form of perturbation.

{\bf Patch-based attack.} Patch-based attacks engender a universal adversarial image patch, which is stuck on the target object's surface to mislead the DNNs. Thus, patch-based attacks are more in 2D image space. In performing patch-based physical attacks, the adversary needs to print out the patch image with a printer and then stick/hang it on the surface of the target, covering its original appearance. The common usage of the adversarial patch is to stick on the clothes, making it wearable \cite{thys2019fooling,hu2021naturalistic}.

{\bf Texture-based attack.} Texture-based attacks generate the adversarial texture, which is wrapped/painted over the 3D model. Thus, texture-based attacks mainly rely on the physical renderer to render the adversarial texture toward the 3D object iteratively. In performing physical attacks, the attacker first makes the adversarial texture physically, then wraps them over the target object's surface, and the original texture is covered. The most representative form is the adversarial camouflage for vehicles \cite{wang2021dual,wang2022fca}.

\subsubsection{Non-invasive attacks} Non-invasive attacks \cite{gnanasambandam2021optical,zhong2022shadows} do not require the attacker to approach and modify the target object. By contrast, the attacker leverages the lighting source to perform physical adversarial attacks, which can be done away from the target object. Therefore, non-invasive attacks feature controllable and concealment. {\bf Optical attacks} is the typical non-invasive attacks. Some devices can perform optical attacks, such as the projector \cite{huang2022spaa,lovisotto2021slap}, the laser emitter \cite{duan2021adversarial}, and the flashlight \cite{gnanasambandam2021optical,wang2023rfla}. Recently, natural phenomena being utilized to perform physical adversarial attacks, such as shadow \cite{zhong2022shadows} and reflect-light \cite{wang2023rfla}.

\subsection{Evaluation}
Based on the correlation between the adversarial object and the sensor device, the physical testing environment can be classified into two categories: stationary physical test and dynamic physical test.

\subsubsection{Stationary physical test} The stationary physical adversarial test experiment means that the adversary object and the sensor device maintain static simultaneously during the experiment. For example, in attacking the automatic checkout system, the attacker sticks the adversarial patch on the commodity's surface and then places it under the sensor input of checkout systems to perform attacks \cite{liu2020bias,wang2021universal}. 

\subsubsection{Dynamic physical test} The dynamic physical test experiment indicates that the adversary and the target object move simultaneously, which contains three different formals 1) the adversarial object is fixed while the sensor device is moving (e.g., drone surveillance); 2) the adversarial object moving while the sensor device is fixed (e.g., video surveillance); 3) both the adversarial object and the sensor device are moving simultaneously (e.g., automatic driving). The attack performance under dynamic physical test may be affected by the following factors a) changing of the environment, b) the size varying of the object, c) motion blur caused by moving. Such factors make realistic physical attacks more challenging.

\begin{figure*}[h]
	\centering
	\begin{minipage}{.7\linewidth}
		\centering
		\includegraphics[width =1\linewidth]{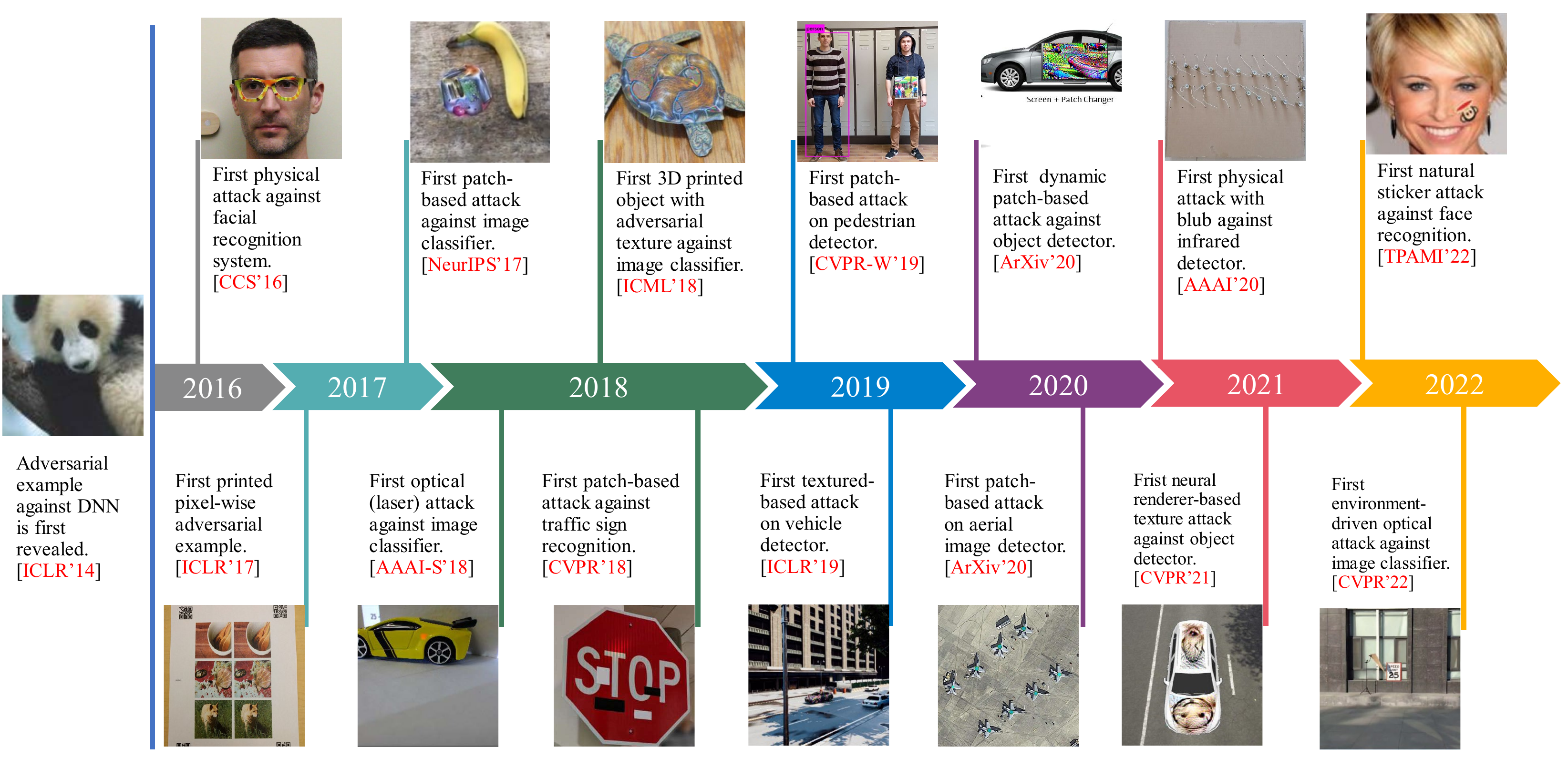}
	\end{minipage}
	\caption{Milestone of physical adversarial attacks on different vision task with different forms of adversarial perturbations.}
	\label{fig:milestone}
\end{figure*}

Finally, we list the statistical result of adversarial attacks against image recognition, object detection, and others in Table \ref{tab:overview}, Table \ref{tab:overview_2}, and Table \ref{tab:overview_3}. We also provide the road map of physical adversarial attacks in Figure \ref{fig:milestone} in terms of their research field and attack type, showing their milestone variant. Due to space limitations, we provide more sophisticated categories, resources (available code), and the follow-up update in \url{https://github.com/winterwindwang/Physical-Adversarial-Attacks-Survey}.

\begin{table*}[htbp]
\centering
\footnotesize
\setlength\tabcolsep{1pt}
\caption{Physical adversarial attacks against image recognition tasks. We list them by time and task, aligning with the discussed order.}
\label{tab:overview}
\begin{threeparttable}
\begin{tabular}{ccccccccc}
\hline
\textbf{Method}                                           & \textbf{Year-Venue}          &  \textbf{\makecell{Adversarial's \\ Knowledge}} & \textbf{\makecell{Threat \\ Model}}  & \textbf{\makecell{Robust \\ Technique}}          & \textbf{\makecell{Physical \\Test Type}}   & \textbf{Space} & \textbf{Remark}         & \textbf{Code} \\ 
\midrule
I-FGSM\cite{ifgsm2018adversarial}              & 2017-ICLR                  & White-box               & Classification      & -                        & Static               & 2D    & Pixel-wise & \checkmark    \\
EOT   \cite{athalye2018synthesizing}           & 2018-ICLR                  & White-box               & Classification      & EOT                      & Static               & 2D    & Pixel-wise & $\times$    \\
D2P\cite{jan2019connecting}                    & 2019-AAAI                  & White-box               & Classification      & EOT,D2P                  & Static               & 2D    & Pixel-wise & \checkmark    \\
MetaAttack\cite{feng2021meta}                  & 2021-ICCV                  & White-box               & Classification      & EOT                      & Static               & 2D    & Pixel-wise & $\times$    \\
ISPAttack\cite{phan2021adversarial}            & 2021-CVPR                  & White-box               & Classification      & -                        & Static               & 2D    & Pixel-wise & $\times$    \\
Attack2Fool\cite{akhtar2021attack}			   & 2021-TPAMI                 & White-box               & Classification      & -                        & Static               & 2D    & Pixel-wise & $\times$    \\
EnchancedPatch\cite{mathov2022enhancing}	   & 2022-Neurocomputing        & White-box               & Classification      & EOT                      & Static               & 3D    & Pixel-wise & $\times$    \\
AdvPatch\cite{brown2017adversarial}            & 2017-NeurIPS               & White-box               & Classification      & EOT                      & Static               & 2D    & Patch      & \checkmark    \\
ACOsAttack\cite{liu2020bias}				   & 2020-ECCV                  & White-box               & Classification      & EOT                      & Static               & 2D    & Patch      & \checkmark    \\
ACOsAttack2\cite{wang2021universal} 		   & 2021-TIP         		    & White-box               & Classification      & EOT                      & Static               & 2D    & Patch      & \checkmark    \\
TnTAttack\cite{doan2022tnt}                    & 2022-TIFS                  & White-box               & Classification      & -                        & Static               & 2D    & Patch      & $\times$    \\
Copy/PasteAttack\cite{casper2022robust}        & 2022-NeurIPS               & White-box               & Classification      &  EOT,TV                 & Static                & 2D    & Patch      &   \checkmark   \\
ViewFool   \cite{dong2022viewfool}             & 2020-NeurIPS               & White-box               & Classification      & -                        & Static               & 3D    & Position   & \checkmark    \\
LightAttack\cite{nichols2018projecting}        & 2018-AAAI-S                & Black-box               & Classification      & EOT                      & Static               & 2D    & Optical    & $\times$    \\
ProjectorAttack\cite{man2019poster}            & 2019-S\&P                  & White-box               & Classification      & -                        & Static               & 2D    & Optical    & $\times$    \\
ABBA\cite{guo2020watch}                        & 2020-NeurIPS               & White-box               & Classification      & EOT                      & Static               & 2D    & Optical & \checkmark     \\
OPAD\cite{gnanasambandam2021optical}           & 2021-ICCV-W                & White-box               & Classification      & EOT                      & Static               & 2D    & Optical    & $\times$    \\
LEDAttack\cite{sayles2021invisible}            & 2021-CVPR                  & White-box               & Classification      & EOT                      & Static               & 2D    & Optical    & \checkmark    \\
AdvLB\cite{duan2021adversarial}                & 2021-CVPR                  & Black-box               & Classification      & -                        & Static               & 2D    & Optical    & \checkmark    \\
SLMAttack\cite{kim2021light}                   & 2021-ArXiv                 & White-box               & Classification      & -                        & Static               & 2D    & Optical    & $\times$    \\
SPAA\cite{huang2022spaa}                       & 2022-VR                    & White-box               & Classification      & D2P                      & Static               & 3D    & Optical    & \checkmark    \\ 
AdvLen\cite{hu2022advlen}                      & 2023-ArXiv                 & Black-box               & Classification      & -                        & Static               & 2D    & Optical    & $\times$    \\ 
AdvFilm\cite{hu2022advfilm}                    & 2023-ArXiv                 & Black-box               & Classification      & -                        & Static               & 2D    & Optical    & $\times$    \\ 
RFLA\cite{wang2023rfla}                        & 2023-ArXiv                 & Black-box               & Classification      & -                        & Static               & 2D    & Optical    & \checkmark    \\ \hline
AdvEyeglass\cite{sharif2016accessorize}        & 2016-CCS                   & White-box               & FR                  & TV,NPS                   & Static               & 2D    & Eyeglasses & $\times$    \\
AdvEyeglass2\cite{sharif2019general}           & 2019-TOPS                  & White-box               & FR                  & Alignment                & Static               & 2D    & Eyeglasses & \checkmark    \\
CLBAAttack\cite{singh2021brightness}           & 2021-BIOSIG                & White-box               & FR                  & EOT                      & Static               & 2D    & Eyeglasses & $\times$    \\
AdvEyeglass3\cite{singh2022powerful}           & 2022-WACV                  & White-box               & FR                  & EOT,TV                   & Static               & 2D    & Eyeglasses & \checkmark    \\
ArcFaceAttack\cite{pautov2019adversarial}      & 2019-SIBIRCON              & White-box               & FR                  & TV                       & Static               & 2D    & Sticker    & $\times$    \\
ReplayAttack\cite{zhang2020adversarial}        & 2020-CVIU                  & White-box               & FR                  & EOT                      & Static               & 2D    & Sticker      & $\times$    \\
AdvHat\cite{komkov2021advhat}                  & 2020-ICPR                  & White-box               & FR                  & TV,Bending               & Static               & 2D    & Sticker    & $\times$    \\
TAP\cite{xiao2021improving}                    & 2021-CVPR                  & White-box               & FR                  & EOT                      & Static               & 2D    & Sticker    & $\times$    \\
AdvSticker\cite{wei2022adversarial}            & 2022-TPAMI                 & Black-box               & FR                  & Blending                 & Static               & 3D    & Sticker    & \checkmark    \\
SOPP\cite{wei2022simultaneously}               & 2022-TPAMI                 & Black-box               & FR                  & -                        & Static               & 2D    & Sticker    & \checkmark    \\
Face3DAdv\cite{yang2022controllable}           & 2022-ArXiv                 & White-box               & FR                  & EOT                      & Static               & 3D    & Sticker    & \checkmark    \\
AdvMakeup\cite{lin2022real}                    & 2023-ICASSP                & White-box               & FR                  & -                        & Static               & 2D    & Sticker    & \checkmark    \\
CAA\cite{zheng2023robust}                      & 2023-PR                    & White-box               & FR                  & D2P,EOT,TV               & Static               & 2D    & Sticker    & \checkmark    \\
AT3D\cite{yang2023towards}                     & 2023-CVPR                  & White-box               & FR                  & EOT                      & Static               & 3D    & Sticker    & \checkmark    \\
AdvMask\cite{zolfi2021adversarial}             & 2022-ECMLPKDD              & White-box               & FR                  & Rendering                & Static               & 3D    & Mask       & \checkmark    \\
LPA\cite{nguyen2020adversarial}                & 2020-CVPR-W                & White-box               & FR                  & Alignment				   & Static               & 2D    & Optical    & $\times$    \\  \hline
$RP_2$\cite{eykholt2018robust}                 & 2018-CVPR                  & White-box               & TSR                 & D2P                      & Static               & 2D    & Patch      & $\times$    \\
RogueSigns\cite{sitawarin2018rogue}            & 2018-ArXiv                 & White-box               & TSR                 & EOT                      & Dynamic              & 2D    & Pixel-wise & \checkmark    \\
PS\_GAN\cite{liu2019perceptual}                & 2019-AAAI                  & White-box               & TSR                 & -                        & Static               & 2D    & Patch      & $\times$    \\
AdvCam\cite{duan2020adversarial}               & 2020-CVPR                  & White-box               & TSR                 & EOT                      & Static               & 2D    & Pixel-wise & \checkmark    \\ 
IAP\cite{bai2021inconspicuous}                 & 2021-IEEE IOT              & White-box               & TSR                 & TV                       & Static               & 2D    & patch      & $\times$    \\ 
ShadowAttack\cite{zhong2022shadows}            & 2022-CVPR                  & Black-box               & TSR			        & EOT                      & Static               & 2D    & Optical    & \checkmark    \\ \hline
LPRAttack\cite{qian2020spot}                   & 2020-C\&S 					& Black-box               & LPR                 & -                       & Static              & 2D    & Patch      & $\times$   \\ \hline
\end{tabular}
\begin{tablenotes}
\footnotesize
\item[*]: Venue with postfix ``-S" and ``-W" indicates the symposium and workshop. \textbf{C\&S }: Computers \& Security.
\item[*]: \textbf{FR}: Face Recognition. \textbf{TSR}: Traffic Sign Recognition. \textbf{LPR}: License Plate Recognition. ``-" in \textbf{Robust Technique} indicates that there are no specific techniques to enhance physical robustness.
\end{tablenotes}
\end{threeparttable}
\end{table*}

\section {Physical Adversarial Attack on Image Recognition Task}
\label{sec:classify}
Advanced deep learning techniques have been increasingly deployed in daily life. For example, facial recognition in automatic payment (e.g., AliPay), image recognition techniques are employed in auto checkout machines in the shopping center \cite{wang2021universal}, and traffic sign recognition is embedded in the computer-aid drive of the automated vehicle. Therefore, a line of literature is proposed to investigate the security of these systems. In this section, we will discuss the existing physical attacks on image recognition tasks from general image recognition, facial recognition, and traffic sign recognition. Table \ref{tab:overview} lists the corresponding physical attacks against image recognition tasks.

\subsection{General image recognition}
The early investigation of physical adversarial attacks against image recognition mainly focuses on verifying the possibility of physical attacks by printing the adversarial image. Kurakin \etal \cite{ifgsm2018adversarial} proposed an iterative version of FGSM \cite{fgsm2015explaining}, resulting in powerful adversarial examples that remain adversarial after printing out to the physical world. Akhtar \etal \cite{akhtar2021attack} indicated that printed adversarial examples constructed by visually salient geometric patterns can mislead DNN. Later, more realistic settings are explored, such as the utilization of adversarial patches\cite{brown2017adversarial} or adversarial viewpoints \cite{dong2022viewfool}.

\subsubsection{EOT}
To construct robust adversarial examples for physical attacks, Athalye \textit{et al.} \cite{athalye2018synthesizing} proposed a general framework to synthesize robust adversarial examples over a set of transformation distributions, which called Expectation Over Transformation (EOT). Specifically, they constructed the the following optimization problem
\begin{equation}
\operatorname*{argmin}_{x_{adv}} \mathbb{E}_{t \sim \mathcal{T}} \left[ \log \Pr(y_t\lvert t(x_{adv})) - \lambda \|LAB(t(x_{adv})) - LAB(t(x))\|_2 \right], 
\label{eq:eot} 
\end{equation}
where the $\Pr(y_t \lvert t(x_{adv}))$ represents the probability of $t(x_{adv})$ be classified to $y_t$, which is used to guarantee the attack performance; $LAB(\cdot)$ is the function that converts the image from RGB space to LAB space, which can represent the color consistent with the human observer better, resulting in imperceptual adversarial perturbation. Note that, they develop an algorithm to optimize the texture of 3D objects in 3D cases scenario. In physical attacks, the author printed the 3D object and the adversarial texture with a 3D printer, then wrapped the adversarial texture over the 3D object. Evaluations were conducted on the captured image, and the results suggested that the multi-view captured images can fool the image recognition model. 

\subsubsection{D2P}
Take further steps than \cite{ifgsm2018adversarial}, Jan \textit{et al.} \cite{jan2019connecting} analyzed the discrepancy between the digital image and the physical adversarial image, then proposed a method to modeling such difference. Specifically, the author first collected enormous digital-physical image pairs by printing digital images and retaking them. Then, the author treated the transformation between the digital and physical worlds as the image-to-image translation and adopted a Pix2Pix model \cite{isola2017image} or a CycleGAN \cite{zhu2017unpaired} to learn the translation mapping. Additionally, the EOT is adopted to boost the robustness of the adversarial perturbation on the simulated physical image (i.e., the generator's output). In physical attacks, the author printed out the adversarial examples and retook them every 15 degrees in the range of $[-60^{\circ}, 60^{\circ}]$. Experiment results suggest that their attack outperforms the EOT and remains valid under different visual angles. However, their attack is the full pixel attack, which is impractical in practice.

\subsubsection{MetaAttack}
Feng \textit{et al.} \cite{feng2021meta} observed that the previous work \cite{jan2019connecting} requires enormous amounts of manually collected data and proposed to utilize the meta-learning technique to solve the problem. Specifically, the author proposed to use GAN to generate adversarial examples. The generator's parameters are trained via the meta-learning technique: for each task, the training set was split into the support set and query set, where the former is used to train the GAN, and the latter is used to validate it. The generator is trained on multi-tasks and fine-tuned on the new tasks in a few-shot learning manner. Moreover, EOT is adopted to enhance physical robustness. The physical experiments suggested that the captured adversarial image obtained a success rate of 95.2\%.

\begin{figure}[t]
	\centering
	\begin{minipage}{.18\linewidth}
		\centering
		\includegraphics[width =1\linewidth]{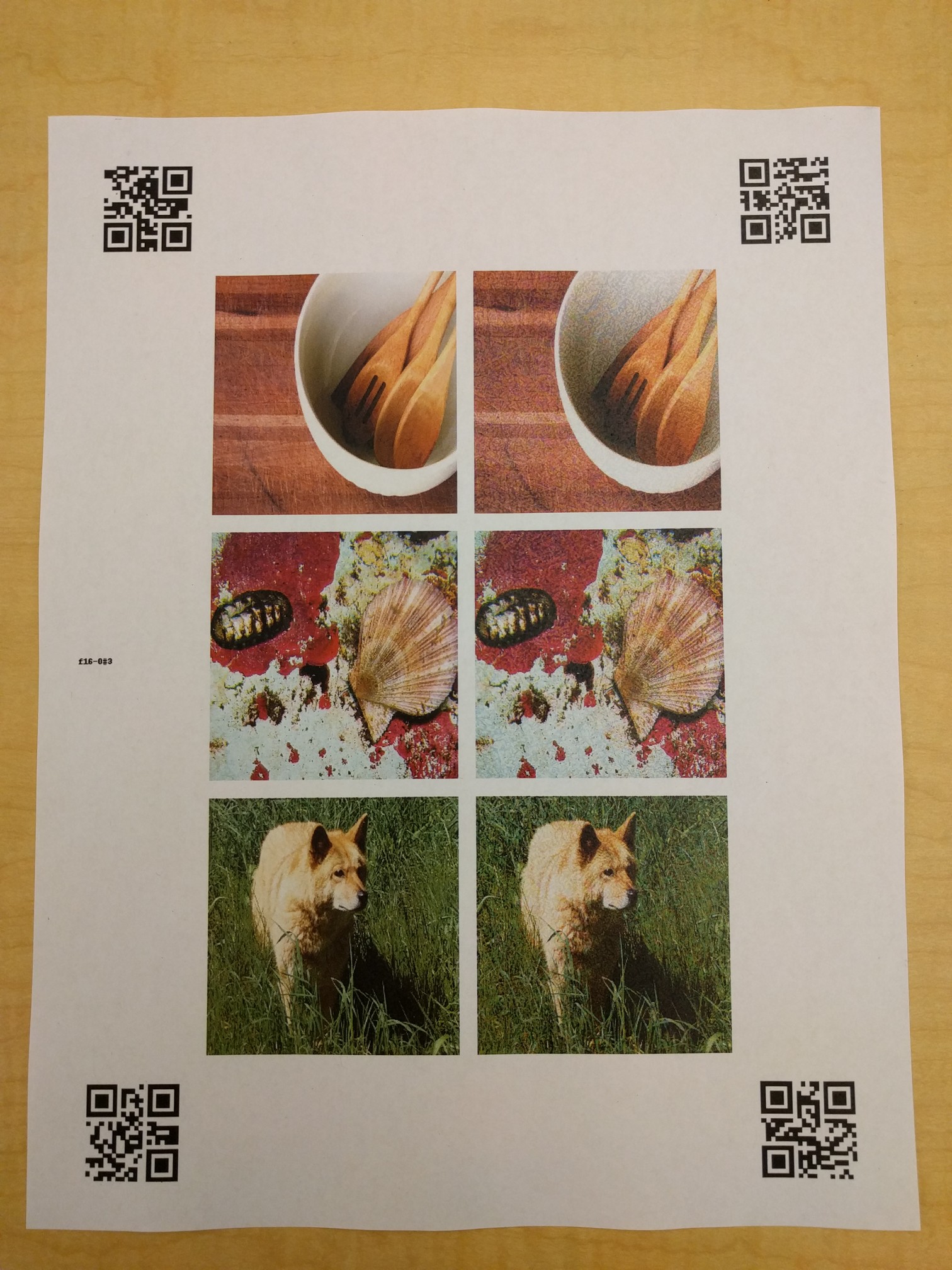}
		\centerline{\footnotesize I-FGSM \cite{ifgsm2018adversarial}} 
	\end{minipage}
	\begin{minipage}{.18\linewidth}
		\centering
		\includegraphics[width =1\linewidth]{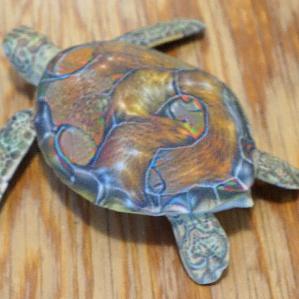}
		\centerline{\footnotesize EOT \cite{athalye2018synthesizing}} 
	\end{minipage}
	\begin{minipage}{.18\linewidth}
		\centering
		\includegraphics[width =1\linewidth]{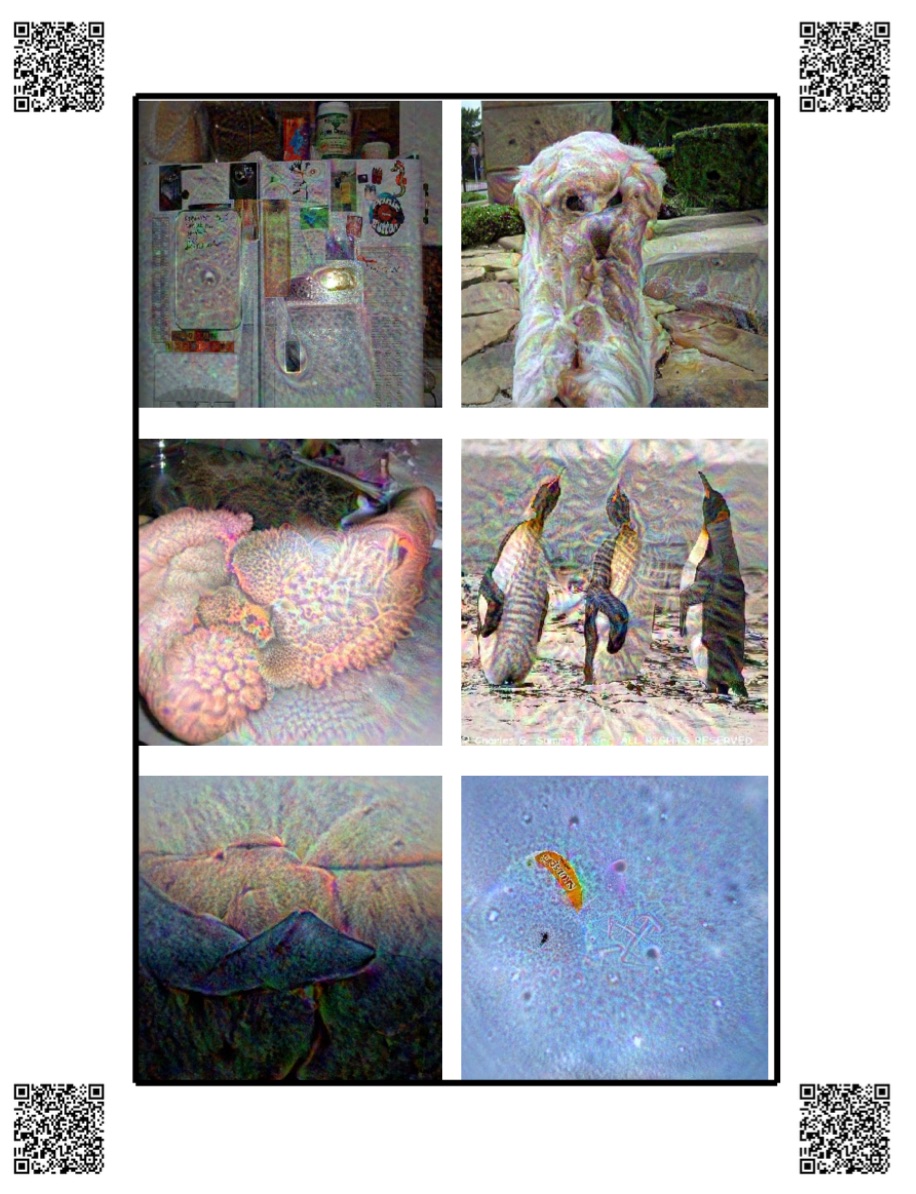}
		\centerline{\footnotesize D2P \cite{jan2019connecting}} 
	\end{minipage}
	\begin{minipage}{.18\linewidth}
		\centering
		\includegraphics[width =1\linewidth]{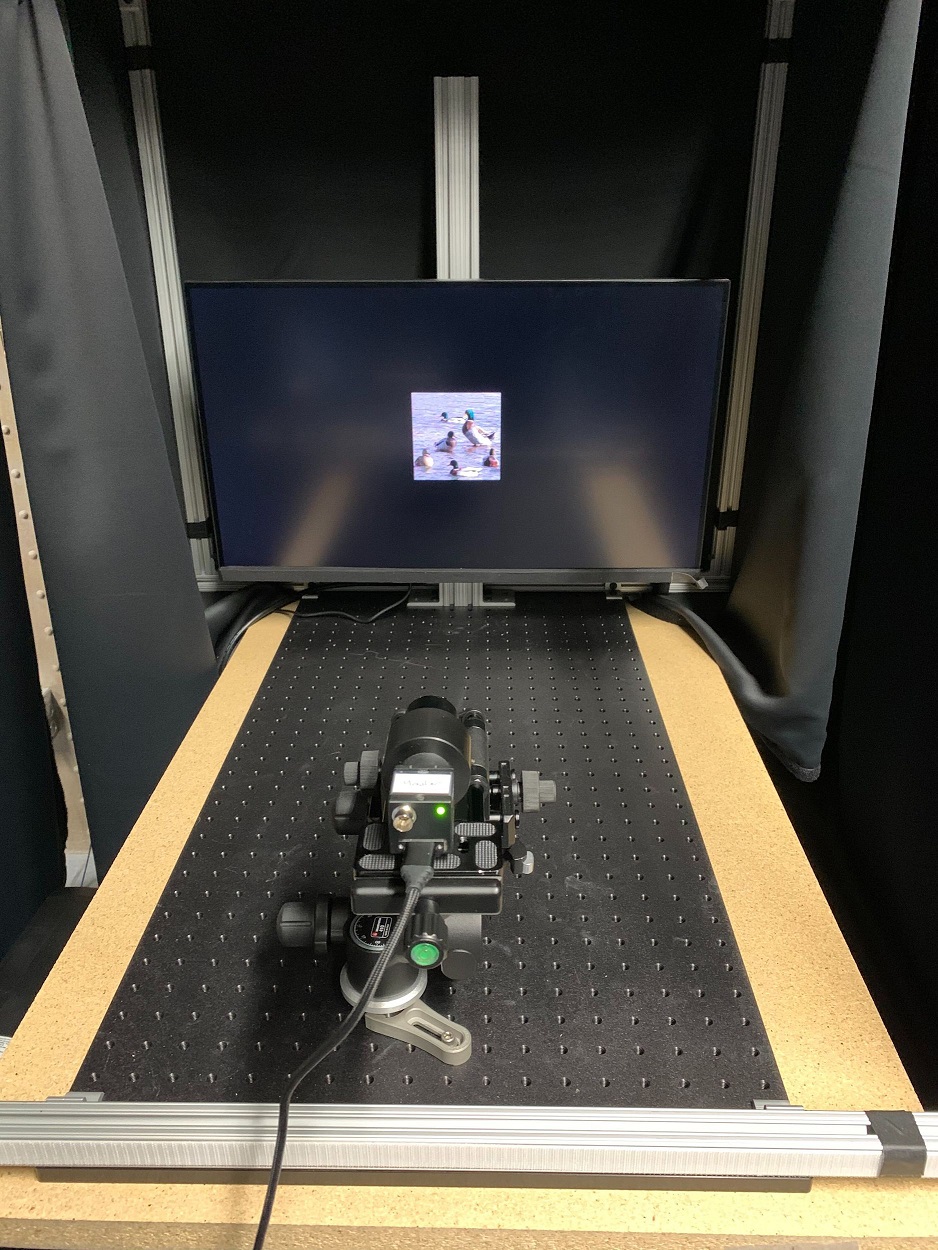}
		\centerline{\footnotesize ISPAttack\cite{phan2021adversarial}} 
	\end{minipage}
	\begin{minipage}{.18\linewidth}
		\centering
		\includegraphics[width =1\linewidth]{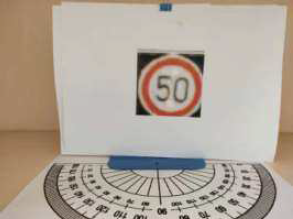}
		\centerline{\footnotesize MetaAttack \cite{feng2021meta}} 
	\end{minipage}
	\caption{Examples of physical adversarial example for general classification.}
	\label{fig:general_physical}
\end{figure}

\subsubsection{ISPAttack}
Unlike modifying the pixel of the RGB images, Phan \textit{et al.} \cite{phan2021adversarial} proposed to attack the imaging processing pipelines (ISP) by modifying the RAW image that outputs from the camera device directly, which makes it robust to compression (i.e., JPEG). They demonstrated that perturbations devised for RAW images remain physically aggressive when printed out and retaken by the camera. 

\subsubsection{Adversarial patch (AdvPatch)}
Concurrent to \cite{athalye2018synthesizing}, Brown \textit{et al.} \cite{brown2017adversarial} developed a general framework to construct the robustness adversarial patch, which remains valid under a set of transformations. Given a small perturbation cannot deceive the model equipped with adversarial defense, the author discarded the constraint on the magnitude of the adversarial patches. In contrast, they adopted the large perturbation to break defense techniques. Specifically, they replaced a partial pixel of the image $x$ with the transformation $t$ optimized adversarial patch $\delta$  at the location $l$, which is reduced as \textit{patch application operator $A(\delta,x,l,t)$}. Mathematically, the above problem is modeled as 

\begin{equation}
\hat{\delta} = \operatorname*{argmin}_{\delta} \mathbb{E}_{x \sim X, t \sim \mathcal{T}, l \sim L}  \left[\log \Pr(y_{t} \lvert A(\delta, x, l, t)) \right] + \|\delta\|_\infty.
\end{equation} 
By solving the above problem, the optimized adversarial patch can successfully mislead the DNN-based application in the smartphone to output the specific class $y_t$. However, lacking constraint on perturbation would produce a visually strange adversarial patch.

\subsubsection{ACOsAttack}
The previous physical attacks failed to explore the attack against the DNN-based application in real life, while Liu \textit{et al.} \cite{liu2020bias,wang2021universal} proposed to attack the automatic check-out system (ACOs) in the shopping center with a universal adversarial patch. Specifically, the author exploits models' semantic and perceptual biases to optimize the universal adversarial patch. The author exploited the semantic bias by training the adversarial patch with the collected samples with the most representative class-related semantic,  alleviating the heavy dependency on large-scale training data. Regarding the perceptual bias, the author extracted the textural patch prior in terms of the style similarities from hard examples, which convey strong model uncertainties.  Moreover, the EOT is adopted to improve the robustness of the adversarial patch. In physical attacks, the author printed the adversarial patch and stuck them on the commodities, then captured it from different environmental conditions (i.e., angle in $\left\{-30^\circ, -15^\circ,0^\circ,15^\circ,30^\circ\right\}$ and distance in $\left\{ 0.3m, 0.5m, 0.7m, 1m\right\}$). Experiment results suggested that the captured adversarial images can easily deceive the system. Additionally, their attack could degrade the recognition accuracy of the commonly used E-commerce platforms (i.e., JD and Taobao) by 43.75\% and 40\%.

\subsubsection{TnTAttack} 
Recently, Bao \cite{doan2022tnt} \textit{et al.} argued that the existing physical adversarial patches are visually conspicuous. To this end, they proposed to generate naturalistic adversarial patches with a generator. Unlike crafting full-pixel adversarial examples with GAN \cite{xiao2018generating}, the author proposed a two-stage optimized method. In the first stage, they trained a generator to map the latent variable $z$ (i.e., random noise) to the naturalistic image (i.e., flower). In the second stage, the author froze the generator's parameters and regarded the latent variable input of the generator as the optimization variable. The generator output an image patch that is pasted on the clean image and fed into the target to calculate the adversarial loss. Then, the latent variable is optimized via the gradient-based method to generate a naturalistic adversarially image patch. Their approach is based on the fact that the trained generator could map the randomly sampled latent variable to the naturalistic image distribution, in which must exist one image with adversarially. In such a way, the author significantly reduced the optimization variable (only 100-dimensional latent variables $z$). In physical attacks, the author printed the patch image and stuck it on the front of clothes, then captured a 1-minute video clip of the person who wore the clothes with the adversarial patch. Experiment results show that over 90\% of captured frames are misclassified by the target network.

\subsubsection{Copy/Paste attack}
Inspired by the ringlet butterfly's predators being stunned by adversarial ``eyespots" on its wings, Casper \cite{casper2022robust} speculated that the DNNs are fooled by interpretable features in the real world. To explore such phenomena, the author proposed to reveal the weakness of the victim network through adversarial perturbation. Specifically, the author used GAN to create the patch, which is pasted on the target image that is classified as the specific target class by the target model. Additionally, EOT is adopted to improve the adversarial patch's robustness, and TV loss is adopted to suppress high-frequency patterns. In physical attacks, the author printed the adversarial patch and pasted it on the object. Evaluation results on the recaptured image verify the effectiveness of their method in the real scenario.

\subsubsection{Renderer-based attack}

Mathov \etal \cite{mathov2022enhancing} pointed out physical attack performance of adversarial patch would be influenced by the following factors: 1) deformation and occlusion of the adversarial patch during physical applying; 2) color distortion caused by shadows and environmental conditions (e.g., light). To mimic the physical conditions during optimization, the author proposed to leverage the physical renderer tools to wrap the adversarial patch over the surface of 3D object.

Unlike generating the adversarial perturbation by modifying the pixel of images, Dong \textit{et al.} \cite{dong2022viewfool} suggested the existence of adversarial viewpoints in the real world, which means that the image taken at a specific view is sufficient to fool the model. To search such viewpoints, the author treated the spacial coordinate and orientation (i.e., yaw, roll, pitch) of the camera in the world axis as the optimization variables and exploited the NeRF \cite{mildenhall2020nerf} to render the object with the searched coordinate and orientation. In physical attacks, the author took pictures from adversarial viewpoints and then verify their attack.

\subsubsection{Optical attacks}
Unlike adversarial patch attacks, optical attacks are more stealthy and controllable as the attacker is not required to modify the target object. One representative characteristic of optical attacks involves the projector-camera model, a process for improving the robustness of physical attacks. It refers to projecting the adversarial perturbation into the real world by the projector and recapturing it by the camera for evaluation. For this reason, Nichols \textit{et al.} \cite{nichols2018projecting} first collected hundreds of real projected and captured image pairs as the training dataset, then they got inspiration from the one-pixel attack and adopted the differential evolution (DE) to find the position of the project point in the image. In contrast to \cite{nichols2018projecting}, Man \textit{et al.} \cite{man2019poster} projected a light source over the whole image to perform attacks. Specifically, to ensure the emitted light can impact the image to fool the classifier, the author treated the color distribution (the means and the standard deviations of the Gaussian distribution) of light as an optimization problem. In physical attacks, the author modulated the projector's color (the value of RGB) to emit the light to the displayed image, then took the photo for testing. Gnanasambandam \textit{et al.} \cite{gnanasambandam2021optical} considered the projector factor and devised a transform function consisting of a matrix and an offset to mimic the process of converting images to the physical scene, where the matrix is used to transfer the image to projector space; and the offset is used to counteract the background illumination. Huang \textit{et al.} \cite{huang2022spaa} formulated the physical projector attack as an end-to-end differentiable process, which was realized by a network that approximates the project-and-capture process, learning the adversarial perturbation on the projected-captured image pairs under the guidance of adversarial loss and stealthiness loss. 

Concurrent to \cite{gnanasambandam2021optical}, Kim \textit{et al.} \cite{kim2021light} modulated the phase of the light in the Fourier domain using a spatial light modulator (SLM) to yield the adversarial attack, where the modulator's parameters are optimized with gradient-based algorithms. Sayles \textit{et al.} \cite{sayles2021invisible} modulated a light signal to illuminate the object then its surface displays striping patterns, which could mislead the classifier to make a mistake in physical attacks. Duan \textit{et al.} \cite{duan2021adversarial} modulated a laser beam to perform adversarial attacks, where the laser's wavelength, layout, width, and intensity are optimized by random search. Hu \etal \cite{hu2022advlen} explored the zoom in/out of the camera on the performance of DNNs. Further, they also exploited the film effect \cite{hu2022advfilm} to perform physical adversarial attacks against DNNs by using color transparency glass. 

In addition to utilizing the lighting source to perform physical attacks, some researchers attempt to leverage the natural phenomenon to perform physical attacks. Guo \cite{guo2020watch} proposed a novel motion attack, which exploits the blurring effect caused by object motion to fool the DNNs. Specifically, the author optimized the specific transform parameters for the object and background to create the blurring effect. The background is extracted by a binary mask generated with a saliency detection model. Moreover, the author used a spatial transformer network \cite{jaderberg2015spatial} to enhance the robustness of adversarial examples. Zhong \textit{et al.} \cite{zhong2022shadows} argued that the pattern of perturbations generated by prior approaches is conspicuous and attention-grabbed for human observers and proposed to utilize the natural phenomenon (i.e., shadow) to perform physical attacks, where the position and shape of the shadow is optimized by PSO. Recently, Wang \etal \cite{wang2023rfla} took a step further than \cite{zhong2022shadows} that has the limited search space, leveraged the reflected light to perform physical adversarial attacks, where the position, shape, and color of reflected light are modeled as in a circle in the image and optimized.

\begin{figure}[t]
	\centering
	\begin{minipage}{.18\linewidth}
		\centering
		\includegraphics[width =1\linewidth]{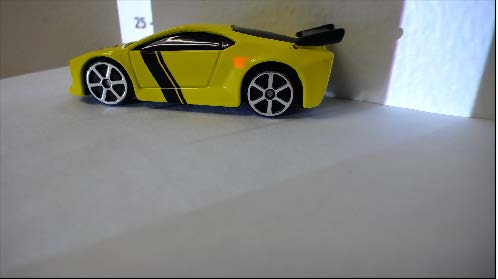}
		\centerline{\tiny LightAttack \cite{nichols2018projecting}} 			
	\end{minipage}
	\begin{minipage}{.18\linewidth}
		\centering
		\includegraphics[width =1\linewidth]{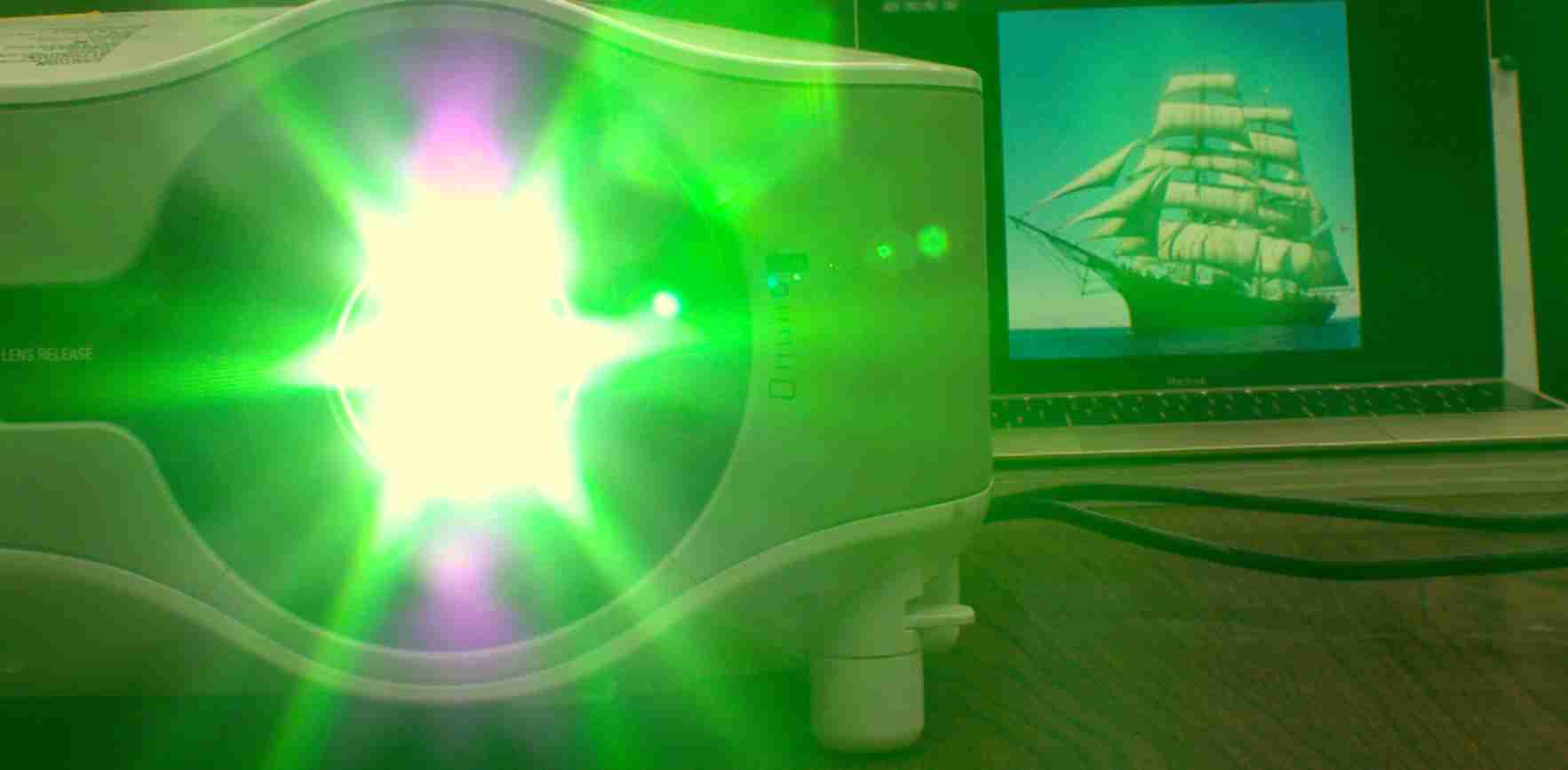}
		\centerline{\tiny ProjectorAttack\cite{man2019poster}} 					
	\end{minipage}
	\begin{minipage}{.18\linewidth}
		\centering
		\includegraphics[width =1\linewidth]{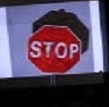}
		\centerline{\tiny OPAD\cite{gnanasambandam2021optical}} 					
	\end{minipage}	
	\begin{minipage}{.18\linewidth}
		\centering
		\includegraphics[width =1\linewidth]{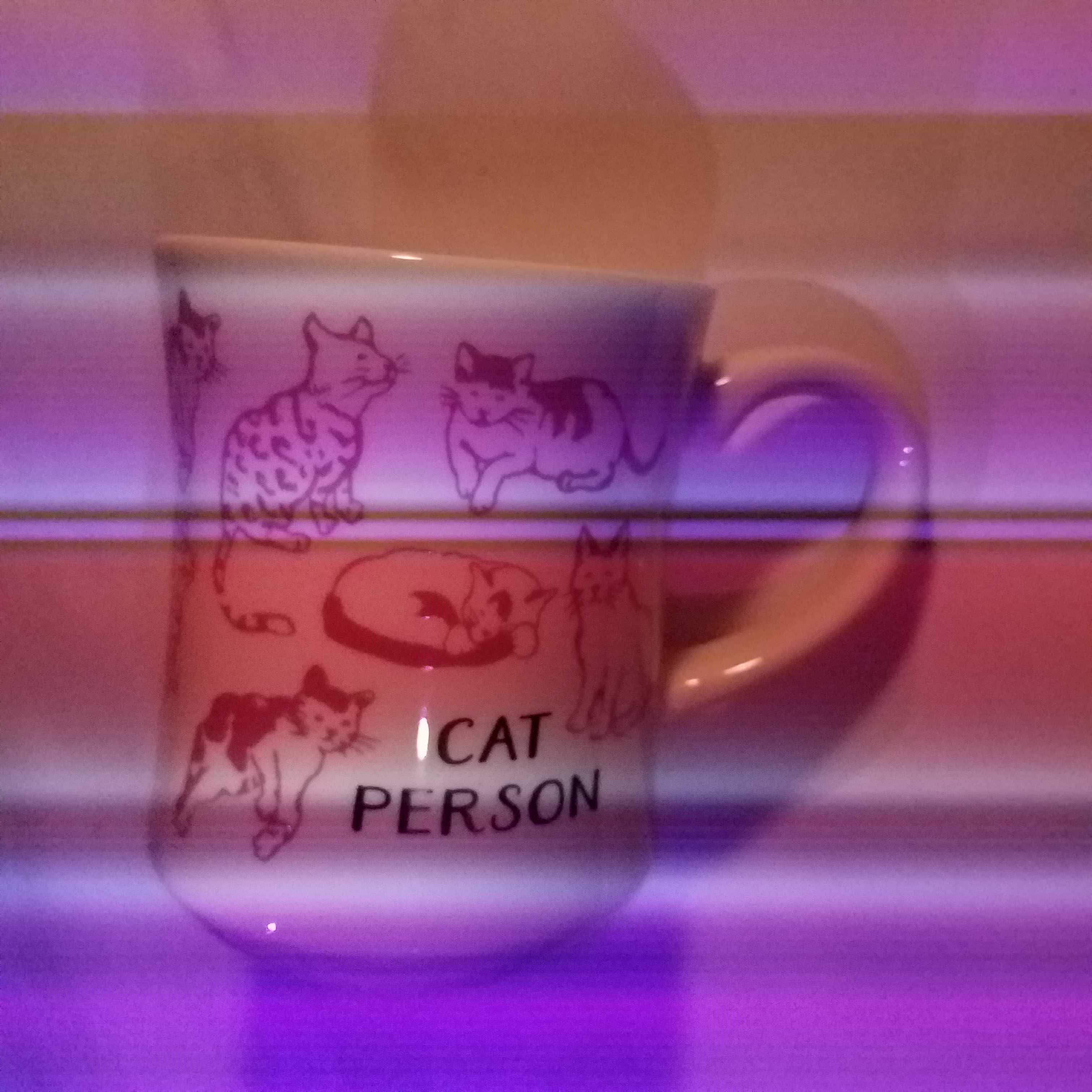}
		\centerline{\tiny LEDAttack\cite{sayles2021invisible}} 	
	\end{minipage}
	
	\begin{minipage}{.18\linewidth}
		\centering
		\includegraphics[width =1\linewidth]{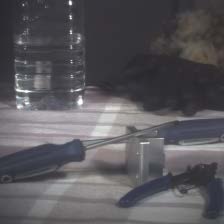}
		\centerline{\tiny SLMAttack\cite{kim2021light}} 	
	\end{minipage}
	\begin{minipage}{.18\linewidth}
		\centering
		\includegraphics[width =1\linewidth]{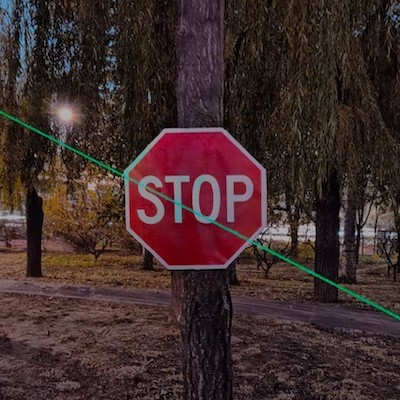}
		\centerline{\tiny AdvLB\cite{duan2021adversarial}} 	
	\end{minipage}
	\begin{minipage}{.18\linewidth}
		\centering
		\includegraphics[width =1\linewidth]{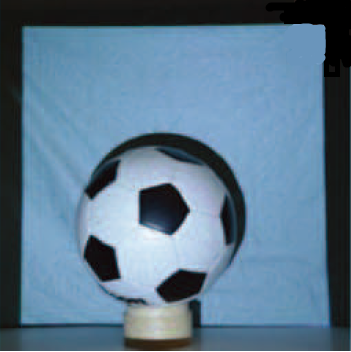}
		\centerline{\tiny SPAA \cite{huang2022spaa}} 	
	\end{minipage}
	\begin{minipage}{.18\linewidth}
		\centering
		\includegraphics[width =1\linewidth]{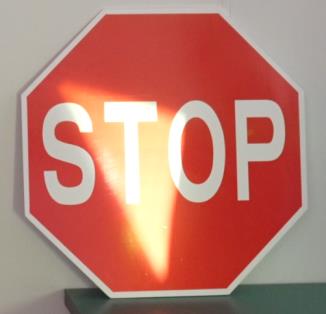}
		\centerline{\tiny RFLA \cite{wang2023rfla}}
	\end{minipage}
	\caption{Examples of optical adversarial examples.}
	\label{fig:optical_attacks}
\end{figure}

Figure \ref{fig:optical_attacks} illustrates various optical adversarial examples. However, despite its stealthiness and controllability, optical attacks are constrained by many factors, such as the object's surface material, the color's saturation, and the scene's light intensity.

\subsection{Facial Recognition}
\begin{figure}[t]
	\centering
	\begin{minipage}{.11\linewidth}
		\centering
		\includegraphics[width =1\linewidth]{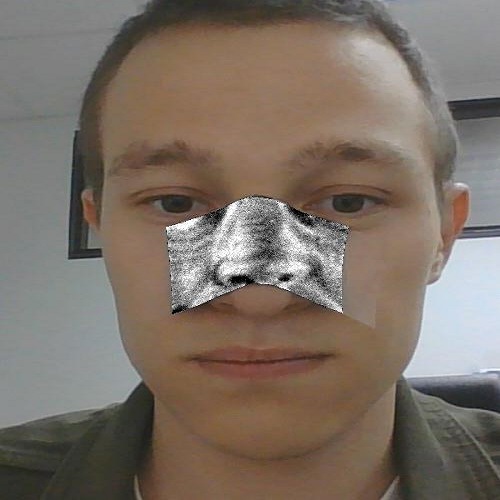}
		\centerline{\footnotesize \cite{pautov2019adversarial}} 
	\end{minipage}
	\begin{minipage}{.11\linewidth}
		\centering
		\includegraphics[width =1\linewidth]{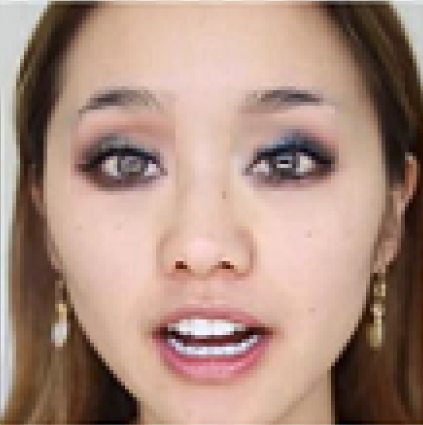}
		\centerline{\footnotesize \cite{zhu2019generating}} 
	\end{minipage}
	\begin{minipage}{.11\linewidth}
		\centering
		\includegraphics[width =1\linewidth]{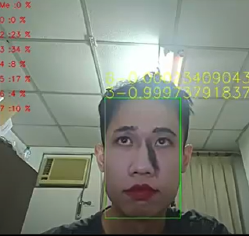}
		\centerline{\footnotesize \cite{lin2022real}}
	\end{minipage}
	\begin{minipage}{.11\linewidth}
		\centering
		\includegraphics[width =1\linewidth]{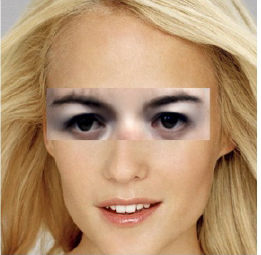}
		\centerline{\footnotesize \cite{xiao2021improving}} 
	\end{minipage}
	\begin{minipage}{.11\linewidth}
		\centering
		\includegraphics[width =1\linewidth]{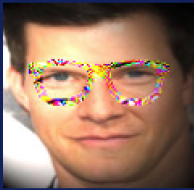}
		\centerline{\footnotesize \cite{singh2021brightness}} 
	\end{minipage}
	\begin{minipage}{.11\linewidth}
		\centering
		\includegraphics[width =1\linewidth]{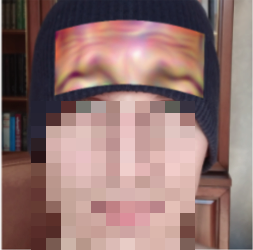}
		\centerline{\footnotesize \cite{komkov2021advhat}} 
	\end{minipage}
	\begin{minipage}{.11\linewidth}
		\centering
		\includegraphics[width =1\linewidth]{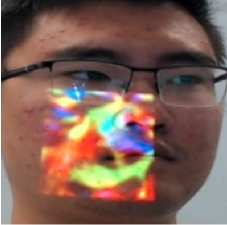}
		\centerline{\footnotesize \cite{nguyen2020adversarial}} 
	\end{minipage}
		
	\begin{minipage}{.11\linewidth}
		\centering
		\includegraphics[width =1\linewidth]{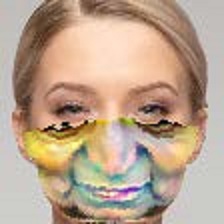}
		\centerline{\footnotesize \cite{zolfi2021adversarial}} 
	\end{minipage}
	\begin{minipage}{.11\linewidth}
		\centering
		\includegraphics[width =1\linewidth]{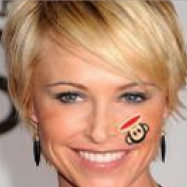}
		\centerline{\footnotesize \cite{wei2022adversarial}}
	\end{minipage}
	\begin{minipage}{.11\linewidth}
		\centering
		\includegraphics[width =1\linewidth]{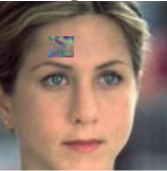}
		\centerline{\footnotesize \cite{wei2022simultaneously}}
	\end{minipage}
	\begin{minipage}{.11\linewidth}
		\centering
		\includegraphics[width =1\linewidth]{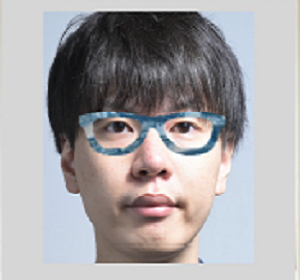}
			\centerline{\footnotesize \cite{singh2022powerful}}
	\end{minipage}
	\begin{minipage}{.11\linewidth}
		\centering
		\includegraphics[width =1\linewidth]{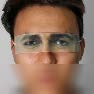}
		\centerline{\footnotesize \cite{yang2022controllable}}
	\end{minipage}
	\begin{minipage}{.11\linewidth}
		\centering
		\includegraphics[width =1\linewidth]{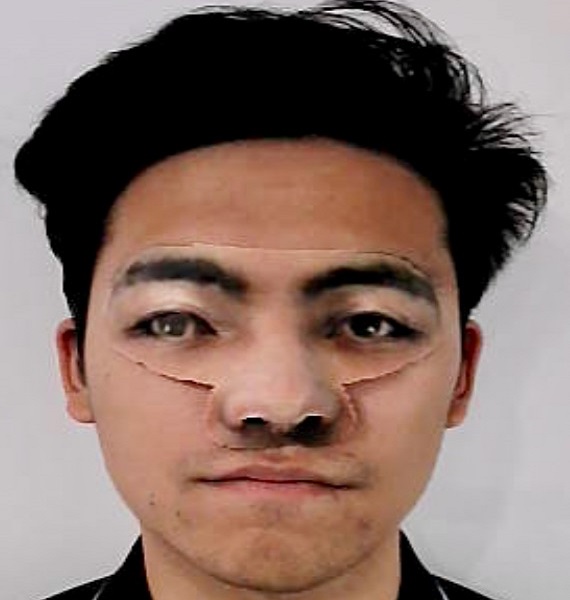}
		\centerline{\footnotesize \cite{yang2023towards}}
	\end{minipage}
	\begin{minipage}{.11\linewidth}
		\centering
		\includegraphics[width =1\linewidth]{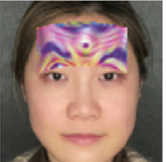}
		\centerline{\footnotesize \cite{zheng2023robust}}
	\end{minipage}
	\caption{Examples of adversarial stickers or light projections.}
	\label{fig:fr_model}
\end{figure}

Facial recognition is a technique to judge whether the given human face is in the face database. In general, the first step of facial recognition is to extract the face region from the input image and feed it to the facial recognition system to predict/match the concrete person. Thus, the adversarial attack against facial recognition can be categorized into impersonation and dodging attacks; the former indicates that the adversary aims to mislead the perturbed face to the specific face, while the latter requires the attacked face to be misidentified as another face except for the original one. Actually, the dodging and impersonation attack is analogous to the untargeted and targeted attacks, respectively. Moreover, physical attacks against the facial recognition system are usually implemented by modifying the accessorize of the face, such as eyeglasses frame \cite{sharif2016accessorize,sharif2019general}, hat \cite{komkov2021advhat}, sticker \cite{wei2022adversarial}, or even makeup \cite{zhu2019generating}. Figure \ref{fig:fr_model} displays the examples generated by different methods.

\subsubsection{Eyeglass attack}
The first physical adversarial attack against face recognition models is developed by Sharif \textit{et al.} \cite{sharif2016accessorize}. To attack the face recognition model, they took the facial accessory (i.e., eyeglasses) as the perturbation carrier. The robustness of the single perturbation is enhanced by training it on a set of images, making it still works in different environmental conditions. Additionally, the TV loss  \cite{mahendran2015tvloss} is adopted to improve the smoothness of perturbation. Furthermore, they devised the NPS loss to avoid producing unprintability color. Finally, the author obtained the perturbation by solving the following optimization problem

\begin{equation}
\operatorname*{argmin}_{\delta} ((\sum_{x \in X} softmaxloss(x+\delta, y_t)) + \lambda_1 \cdot TV(\delta) + \lambda_2 \cdot NPS(\delta)),
\end{equation}
where $softmaxloss(\cdot)$ is the adversarial loss for the face recognition, $\lambda_1$ and $\lambda_2$ are the weight to balance the objectives. To conduct the physical experiment, the author printed and cropped out the adversarial eyeglass frame and stuck it on the real eyeglasses. Then, the author asked the participant to wear adversarial eyeglasses and stand at a fixed position in the indoor environment and then gathered $30 \sim 50$ images for each participant. The evaluation result on those images indicated that adversarial eyeglass frames could effectively deceive the face recognition system.

\begin{figure}[t]
	\centering
	\begin{minipage}{.18\linewidth}
		\centering
		\includegraphics[width =1\linewidth]{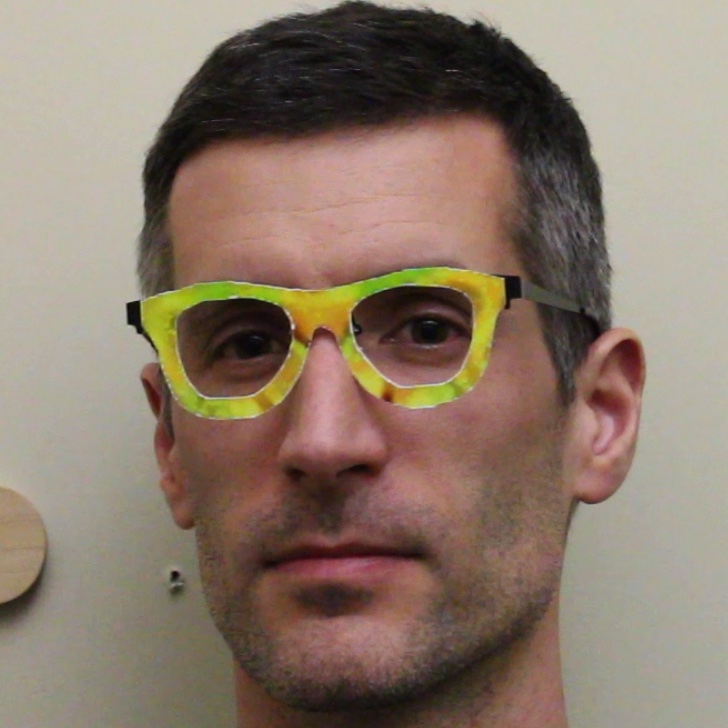}
	\end{minipage}
	\begin{minipage}{.18\linewidth}
		\centering
		\includegraphics[width =1\linewidth]{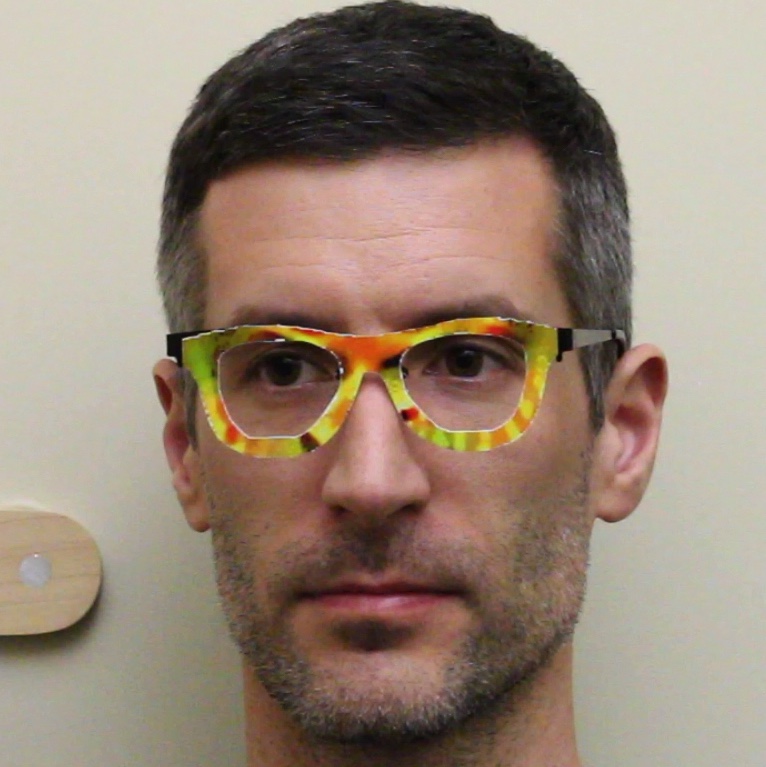}
	\end{minipage}
	\begin{minipage}{.18\linewidth}
		\centering
		\includegraphics[width =1\linewidth]{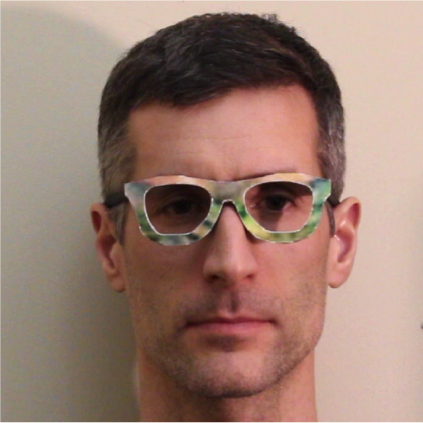}
	\end{minipage}
	\begin{minipage}{.18\linewidth}
		\centering
		\includegraphics[width =1\linewidth]{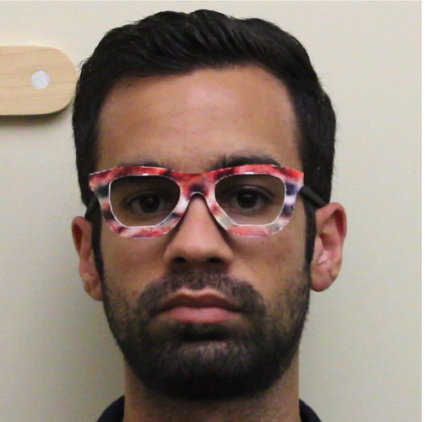}
	\end{minipage}

	\begin{minipage}{.18\linewidth}
		\centering
		\includegraphics[width =1\linewidth]{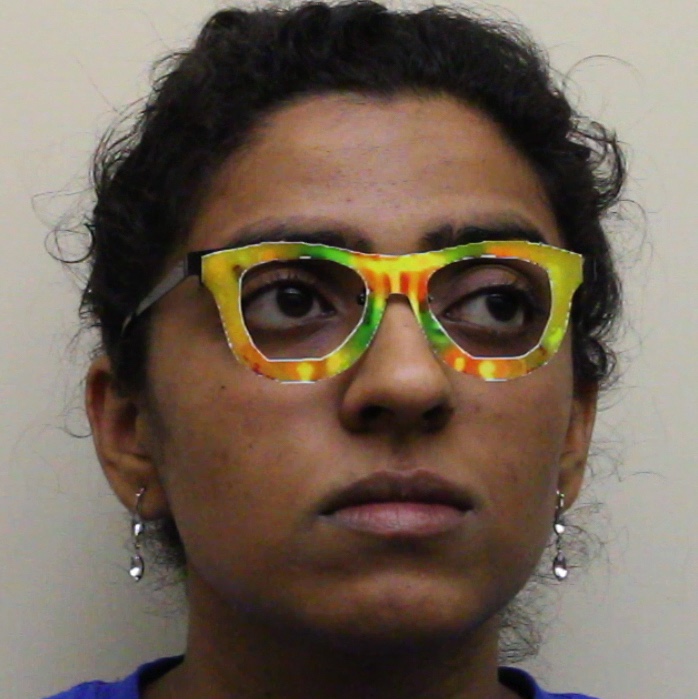}
		\centerline{\footnotesize (a)} 	
	\end{minipage}
	\begin{minipage}{.18\linewidth}
		\centering
		\includegraphics[width =1\linewidth]{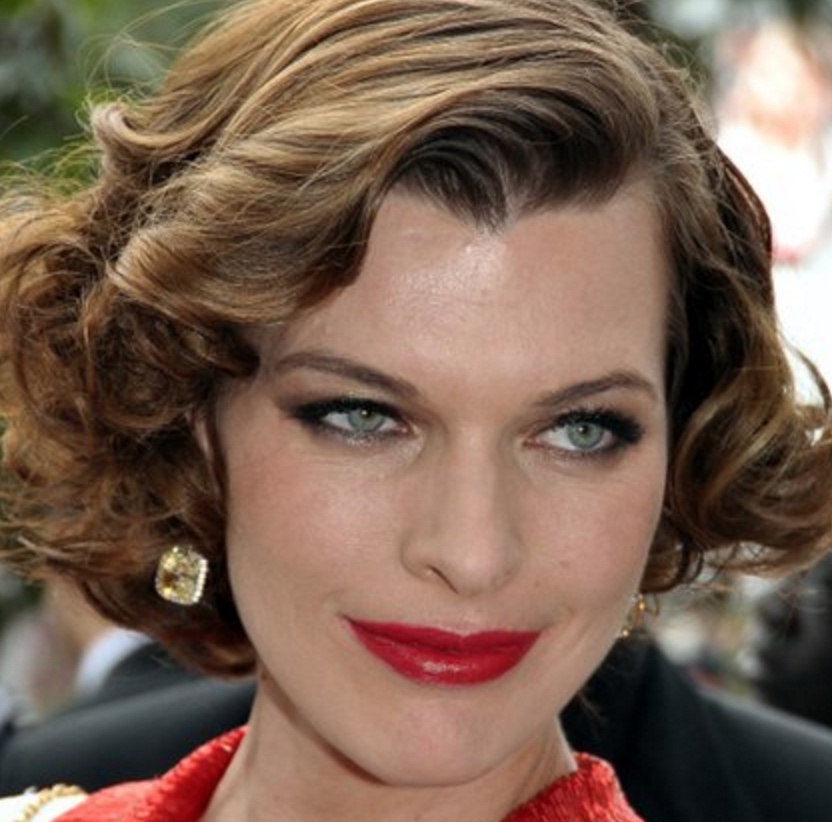}
		\centerline{\footnotesize (b)}
	\end{minipage}
	\begin{minipage}{.18\linewidth}
		\centering
		\includegraphics[width =1\linewidth]{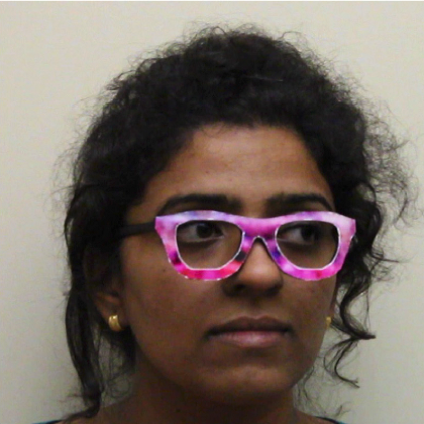}
		\centerline{\footnotesize (c)}
	\end{minipage}
	\begin{minipage}{.18\linewidth}
		\centering
		\includegraphics[width =1\linewidth]{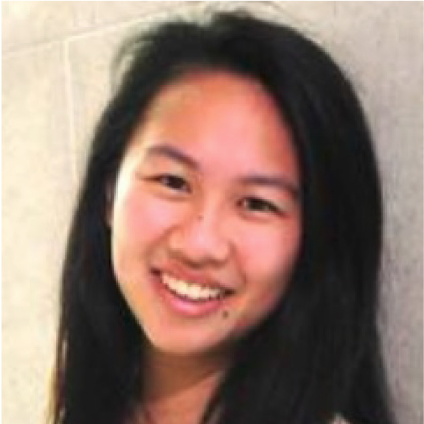}
		\centerline{\footnotesize (d)}
	\end{minipage}
	\caption{Examples of adversarial eyeglasses for dodging and impersonation attacks of \cite{sharif2016accessorize} (a,b) and \cite{sharif2019general} (c,d).}
	\label{fig:eyeglass}
\end{figure}

In their latter works \cite{sharif2019general}, they devoted to improving the naturalness of adversarial eyeglasses. Thus, the author developed a generative-based method to craft the adversarial glasses frame with a natural pattern. Specifically, the author first collected enormous images of real glasses frames as the training set, then trained an adversarial generative network (AGN) to learn how to synthesize natural eyeglass frames. To further enhance the robustness of the adversarial glasses in physical attacks, the author introduced the following three methods. The first one is to utilize multiple images of the attacker to improve the robustness of adversarial glasses under different conditions. The second is to collect various pose images to make the attacks robust to pose changing. The third is to use the Polynomial Texture Maps approach \cite{malzbender2001polynomial} to confine the RGB value of eyeglass under baseline luminance to values under a specific luminance, making the attacks robust to varying illumination conditions. In physical attacks, the author printed out the eyeglasses frame and adversarial pattern and then stuck the pattern over the eyeglasses frame. After that, the author captured the image for evaluation. Experiment results show that their method outperforms \cite{sharif2016accessorize} by the improvement of 45\% under different physical conditions. Figure \ref{fig:eyeglass} displays the printed adversarial eyeglasses generated by \cite{sharif2016accessorize} and \cite{sharif2019general}.

Lately, Singh \textit{et al.} \cite{singh2021brightness} leveraged the concept of curriculum learning to promote the robustness of attack to changing brightness against the facial recognition model. In optimization, the author constantly changes the brightness of face images stuck with adversarial eyeglasses, which makes their attack robust to brightness changes. Recently, Singh \etal \cite{singh2022powerful} combined the visible adversarial eyeglasses frame and full pixel-wise imperceptible to attack face recognition system, a threshold version of TV loss is adopted to improve the naturalness of adversarial examples. However, the author printed the adversarial example rather than the adversarial eyeglasses to perform physical attacks, which is impractical.

\subsubsection{Sticker}
Unlike treating the eyeglasses as the perturbation carrier, Pautov \textit{et al.} \cite{pautov2019adversarial} proposed a simple but effective method to craft the adversarial sticker to deceive face recognition. To apply the adversarial sticker over the facial area, the author applied the projective transformation over the adversarial sticker, then stuck the sticker on the nose or eye of the adversary. To optimize the sticker, the author treated the cosine similarity between the image embedded with the patch and the original clean image as adversarial loss, which is expressed as follows.

\begin{equation}
	\mathcal{L}(X, \delta) = \mathbb{E}_{t \in \mathcal{T}, x \in X} \left[ cos(e_{t(x_{y_t})}, e_{t(x_{adv})})\right],
\end{equation}
where $e_{x_t}$ and $e_{x'}$ are the embedding corresponding to the desired person ${x_t}$ and the photo $x_{adv}$ of the attacker with the applied patch. In addition, TV loss is adopted to constrain the visual effect of the adversarial sticker. To avoid the color discrepancy between the physical and the digital perturbation, the author instead optimizes the sticker on a grayscale. They conduct the physical attack by printing out various shapes (i.e., eyeglasses and stickers) of stickers and stick on a specific area of the face. Experiment results show that their sticker can dramatically deteriorate the accurate of Face ID in iPhone devices.

Zhang \textit{et al.} \cite{zhang2020adversarial} generated a physical adversarial patch optimized with EOT that covers the whole face to break the face authentication system equipped with the DNN-based spoof detection module.

\begin{figure}[h]
	\centering
	\begin{minipage}{.5\linewidth}
		\centering
		\includegraphics[width =1\linewidth]{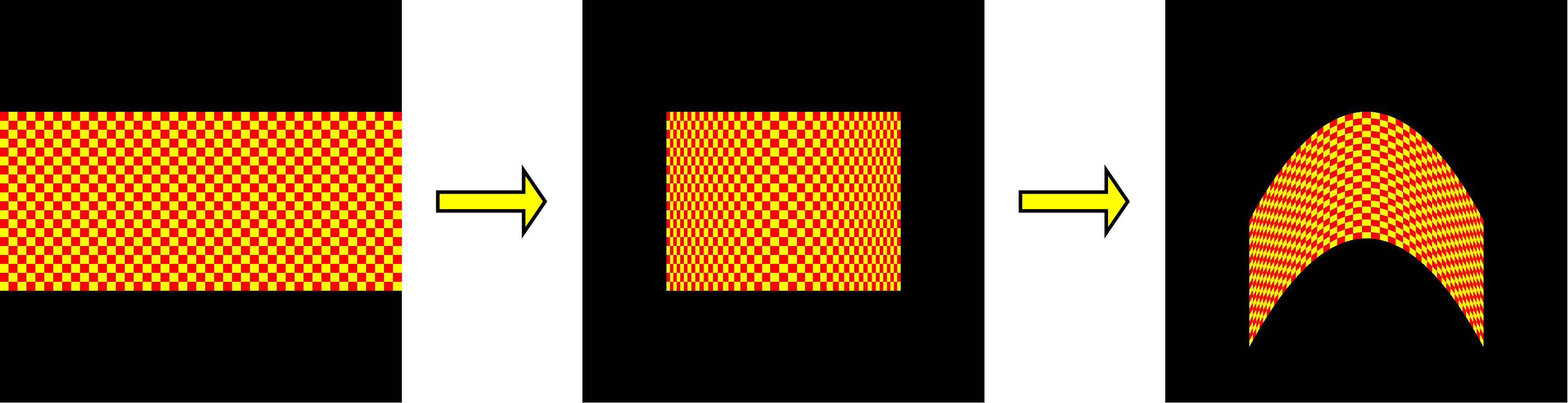}
	\end{minipage}
	\begin{minipage}{.2\linewidth}
		\centering
		\includegraphics[width =1\linewidth]{figures/fig_adv_hat.png}
	\end{minipage}
	\caption{Bend and rotation over the rectangle (left) and the stuck example (right)\cite{komkov2021advhat}.}
	\label{fig:adv_hat_transfer}
\end{figure}

Komkov \textit{et al.} \cite{komkov2021advhat} devised the adversarial sticker with the rectangle shape, which is stuck on the hat to attack the Face ID model. To optimize the adversarial sticker for the hat, the author first applied the bend and rotation (see Figure \ref{fig:adv_hat_transfer}) transformation over the sticker to approximate the actual sticker process, ensuring the length of the rectangle does not change, which is realized by performing a differentiable formula on the pixel. The adversarial loss (i.e., cosine similarity distance) and TV loss are adopted to optimize the adversarial sticker. In physical attacks, the author stuck the adversarial sticker on the forehead area of the hat. Experiment results show that the person wearing the generated adversarial sticker easily bypasses the Face ID system (e.g., ArcFace).

Concurrent to \cite{komkov2021advhat}, Xiao \textit{et al.} \cite{xiao2021improving} proposed to improve the transferability of the adversarial patch for face recognition systems. The author analyzes the drawback of existing transferability methods: sensitive to the initialization (e.g., MIM \cite{mim2018BoostingAA}) and exhibits severe overfitting to the target model when the magnitude of perturbation is substantial. To address these issues, the author regularizes the adversarial patch in a low-dimensional manifold represented by a generative model regularization. In optimization, the adversarial patch is bent and pasted on the eye area of the face. The author conducts the physical attack by printing and photographing the adversarial patch stuck in the human eye and achieves successful attacks. 

Unlike optimizing the pattern or color of adversarial perturbation, Wei \textit{et al.} \cite{wei2022adversarial} demonstrated that the natural image patch is sufficient to attack facial recognition models. Specifically, the author proposed a general framework for searching the most sensitive area of the model that is susceptible to the sticker by traversing the face image with a patch image, accelerating the search speed. In optimization, the author used the bend and rotation transformation in the 3D space to transform the adversarial patch to mimic the deformation of the sticker on the real face. In physical attacks,  the author stuck the adversarial patch on the attacker's face according to the searched coordinate. In their late work \cite{wei2022simultaneously}, they found that the position and the pattern of adversarial patches are equally significant to a successful attack. Thus, the author optimized the positions and simultaneously perturbation against the face recognition model. Specifically, the author used a UNet architecture to generate the position and the parameter (i.e., attack step) of the existing gradient-based algorithm (e.g., I-FGSM \cite{ifgsm2018adversarial}) to reduce the optimization variables. In optimization, the author adopted the reinforcement learning framework to model and solve the above problem. In physical attacks, the author printed the adversarial patch with photo paper as it can reduce the color discrepancy as much as possible. Experiment results on their method achieved an average success rate of 66.18\% on captured images.

In the prevalence of the COVID-19 pandemic, wearing masks has become common sense. Zolfi \textit{et al.} \cite{zolfi2021adversarial} got inspiration from this observation and generated a mask-like universal adversarial patch, which makes the attack more stealthy. The adversary who wears the crafted adversarial mask can bypass the face recognition systems in some scenarios, e.g., airport. Specifically, the author proposed a mask projection method to convert the faces and masks into UV space to mimic actual wearing. The mask image is optimized by minimizing the cosine similarity loss and TV loss to ensure adversarial and smoothness. In addition, geometric transformations and color-based augmentations are adopted to improve the robustness of the mask. 

Similar to \cite{mathov2022enhancing}, Yang \etal \cite{yang2022controllable} developed a comprehensive tool based on the generative model to provide a fair evaluation of physical adversarial patches against face recognition. Moreover, they further proposed an attack method to optimize the eye texture (which will be cropped out as a sticker to perform physical attacks) of the 3D face. In their latter work \cite{yang2023towards}, they pointed out the existing patch-based physical attack may fail toward facial recognition equipped with the face anti-spoof system as the patch cannot align the human face well. Thus, the author proposed to synthesize a 3D adversarial face mask by optimizing the face latent variable extracted by a 3D MM model \cite{blanz1999morphable}, engendering a more natural face mask.

Zheng \etal \cite{zheng2023robust} systematically analyze the factor that impacts the physical adversarial attacks on the face recognition system, which includes 1) deformation and position disturbance caused by wearing the adversarial sticker; 2) chromatic aberration of the sticker caused by the printers and cameras; 3) photographing variations caused by different camera pose and environment light conditions. 4) different facial expressions and movements. To address the above issues, the author adopted the D2P to solve 1) and 2), EOT and off-plane blend to solve 3), and adopted the curriculum adversarial attack to solve 4).

\subsubsection{Makeup}
Zhu \etal \cite{zhu2019generating} proposed to attack the face recognition system with makeup generated by a generator, which gets the input of a clean face image and adversarial eye region patch generated by a CycleGAN \cite{zhu2017unpaired}. Despite their imperceptible while lack of physical attack tests. Recently, similar to \cite{zhu2019generating}, Lin \etal \cite{lin2022real} adopted a similar technique to generate adversarial makeup. The difference to \cite{zhu2019generating} is that they trained the generator to synthesize the adversarial makeup face directly with CycleGAN, and the Gaussian blur is introduced to eliminate the distortion caused by the manual application of makeup. Their effectiveness is verified in the physical test.

\subsubsection{Light Projection Attack (LPA)}
Unlike the adversarial sticker approaches, Nguyen \textit{et al.} \cite{nguyen2020adversarial} exploited the adversarial light projection to attack the face recognition systems. The critical technique of the light project attack is calibrating the adversarial region and color between the projector and digital images. To address the issue, the author proposed two calibration approaches (i.e., manual assign and automatic obtain with a landmark detection) to constrain the adversarial area. To calibrate the color between the digital adversarial pattern and the physical version of adversarial light emitted by the projector, the author presented a color transformation function $\gamma$ (Equation \ref{eq:calibrateion}) to implement the conversion between the camera ($C$) and projector ($\mathcal{G}$).

\begin{equation}
\begin{split}
C(\hat{x}_{adv} + \mathcal{G}(\gamma(h(x)))) = x + h(x) \\
\Longleftrightarrow C(\hat{x}_{adv}) + C(\mathcal{G}(\gamma(h(x)))) = x + h(x) \\
\Longleftrightarrow C(\mathcal{G}(\gamma(h(x)))) = h(x) \\
\Longleftrightarrow \gamma = (C \odot \mathcal{G})^{-1},
\end{split}
\label{eq:calibrateion}
\end{equation}
where $h(x)$ represents the algorithm to search adversarial perturbation. Additionally, the author pointed out that $\gamma$ in LAB color space can better approximate the physical domain RGB color space. In physical attacks, the author projects the adversarial light on the adversary's face to attack the facial recognition model. 

To sum up, the main attack pipeline against facial recognition can be briefly described as follows. First, apply the adversarial perturbation precisely over the face area using bend and rotation or 2D/3D conversion via rendered. Meanwhile, a set of transformations or data augmentation are utilized to improve the robustness. Finally, the cosine similarity-based adversarial loss \cite{pautov2019adversarial,komkov2021advhat,zolfi2021adversarial,nguyen2020adversarial,singh2021brightness} and TV loss or NPS loss to ensure the physical deployable.

\subsection{Traffic sign recognition}
\begin{figure}[t]
	\centering
	\begin{minipage}{.13\linewidth}
		\centering
		\includegraphics[width =1\linewidth]{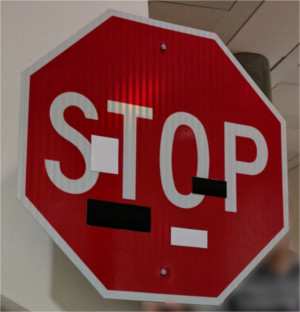}
		\centerline{\tiny $RP_2$ \cite{eykholt2018robust}} 
	\end{minipage}
	\begin{minipage}{.13\linewidth}
		\centering
		\includegraphics[width =1\linewidth]{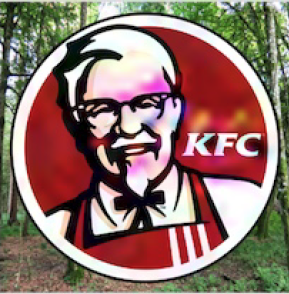}
		\centerline{\tiny AdvLogo\cite{sitawarin2018rogue}} 
	\end{minipage}
	\begin{minipage}{.13\linewidth}
		\centering
		\includegraphics[width =1\linewidth]{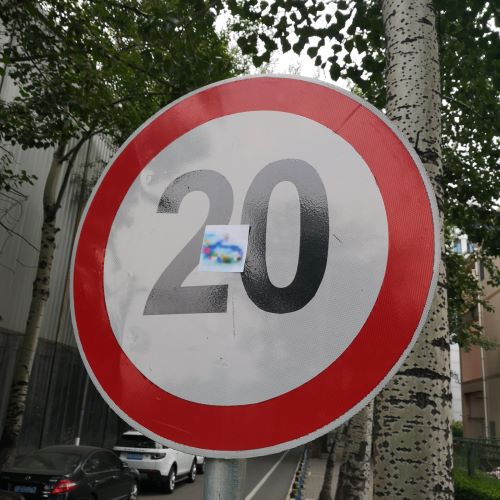}
		\centerline{\tiny PS\_GAN \cite{liu2019perceptual}} 
	\end{minipage}
	\begin{minipage}{.13\linewidth}
		\centering
		\includegraphics[width =1\linewidth]{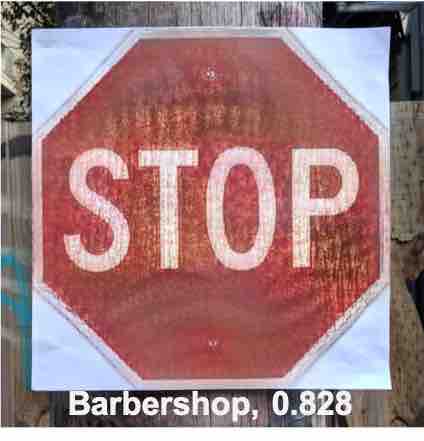}
		\centerline{\tiny AdvCam\cite{duan2020adversarial}} 
	\end{minipage}
	\begin{minipage}{.13\linewidth}
		\centering
		\includegraphics[width =1\linewidth]{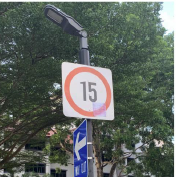}
		\centerline{\tiny IAP \cite{bai2021inconspicuous}} 
	\end{minipage}
	\begin{minipage}{.13\linewidth}
		\centering
		\includegraphics[width =1\linewidth]{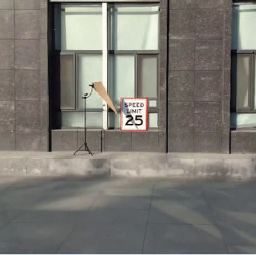}
		\centerline{\tiny ShadowAttack\cite{zhong2022shadows}} 
	\end{minipage}
	\begin{minipage}{.13\linewidth}
		\centering
		\includegraphics[width =1\linewidth]{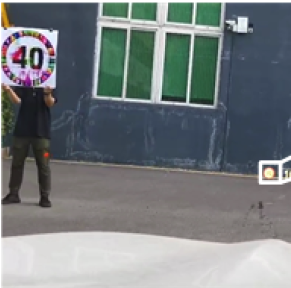}
		\centerline{\tiny 4AAttack\cite{jia2022fooling}} 
	\end{minipage}
	\caption{Examples of adversarial examples against traffic sign recognition.}
	\label{fig:traffic_sign_cls}
\end{figure}

Traffic sign recognition is one of the widely used applications of the DNN model, which can be deployed in various scenarios, such as road sign recognition in automatic driving or license plate recognition in auto-checkout parking lots. Therefore, a line of work has been proposed to explore the security risk of these models. The examples of various physical attack against the traffic sign recognition model is provided in Figure \ref{fig:traffic_sign_cls}.

\subsubsection{RP$_2$}
Eykholt \textit{et al.} \cite{eykholt2018robust} first demonstrated that the traffic sign recognition model is vulnerable to simple white-black block physical adversarial attack. The author proposed to optimize multiple white-black blocks with rectangle shapes. Specifically, the author first collected considerable physical traffic sign images as the training set to expect to learn the ability to resist various physical environmental conditions. Moreover, they adopt the NPS loss to constrain the color of the adversarial patches, eliminating the color discrepancy between the digital adversarial patch and the printed physical adversarial patch. Finally, the author formulated the above problem as follows

\begin{equation}
\operatorname*{argmin}_{\delta}~~ \mathbb{E}_{x_i \sim X} J(f(x_i + t(M_x \cdot \delta)), y) + \lambda \|M_x \cdot \delta \|_p + NPS,
\label{eq:eykholt_v1}
\end{equation}
where the $M_x$ is the binary mask indicating the patch location; $X$ is the training dataset consisting of digital and physical examples, and $t(M_x \cdot \delta)$ indicates the transformation function for the perturbed object. In physical attacks, the author performs the experiments under various camera distances (i.e., $\left\{1.5m, 3m, 12m\right\}$) and angles (i.e., $0^\circ \sim 15^\circ$) on the adversarial patch. Additionally, they also conduct dynamic physical attacks by recording the video while driving the car toward the adversarial traffic sign. Experiment results suggested that their method can dramatically deceive the traffic sign recognition model. Additionally, the author empirically found that 1) adversarial performance is easily impacted by the location of the mask; 2) the mask optimized by $L_1$  regularization show more sparse that represents the most vulnerable of the model, while the $L_2$ result in successful attacks at an earlier iterative step.

Concurrent to \cite{eykholt2018robust}, Sitawarin \textit{et al.} \cite{sitawarin2018rogue} proposed a general attack pipeline, which can generate the adversarial logo, traffic sign, and custom adversarial sign. In optimization, the author used a circle mask to guide the perturbation located in the target area (e.g., traffic sign). In addition, a set of transformations is used to ensure robustness. Finally, the adversarial perturbation is optimized by minimizing the C\&W adversarial loss \cite{carlini2017towards}. To perform dynamic physical attacks, the author drove a car and recorded the video to capture the adversarial sign from far (80 feet away) to near. Experiment results showed their attack could achieve a 95.74\% of attack success rate.

\subsubsection{PS\_GAN}
Unlike the optimization-based attacks \cite{eykholt2018robust,sitawarin2018rogue}, Liu \textit{et al.} \cite{liu2019perceptual} proposed to utilize the generative model to construct inconspicuous adversarial patches. Specifically, the author exploited the attention map extracted by the Grad-CAM \cite{selvaraju2017grad} to guide the paste location of adversarial patches for better attack performance. Meanwhile, they minimized the discrepancy between the adversarial patch and the seed patch for better visual fidelity. In physical attacks, the author printed the adversarial patch and stuck it on the traffic sign (i.e., ``Speed Limit 20") in the street with varying angles (e.g., $\left\{0^\circ, 15^\circ,30^\circ,-15^\circ,-30^\circ \right\}$) and distances (e.g., $\left\{1m, 3m, 5m\right\}$). Experiment results suggested that most of the captured traffic sign images pasted with adversarial patches are misclassified by the model (from 86.7\% to 17.2\%). 

\subsubsection{AdvCam}
Take further step than \cite{liu2019perceptual}, Duan \textit{et al.} \cite{duan2020adversarial} proposed a novel approach to camouflage the adversarial perturbation (i.e., AdvCam) into a natural style that appears legitimate to human observers. Specifically, they exploited the style transfer technique \cite{gatys2016image} to hide the large magnitude perturbations into customized styles, which makes attacks more stealthy. They adopt the following adversarial loss to maintain the aggressive

\begin{equation}
\mathcal{L}_{adv} = \left\{
\begin{aligned}
& \log(\Pr(y_{adv} \mid x_{adv})), ~~\rm for~targeted~attack \\
& -\log(\Pr(y\mid x_{adv})), ~~\rm for~untargeted~attack,
\end{aligned}
\right.
\end{equation}
where $\Pr(y_{adv}\mid \cdot)$ is the probability output (softmax on logits) of the target model on input image with respect to class $y_{adv}$. The style loss and content loss were adopted to ensure style transfer and the TV loss is introduced to reduce the variations between adjacent pixels. Additionally, the author utilized various physical conditions (i.e., rotation, scale size, and lighting change) to boost the robustness of adversarial patterns in the real world. Moreover, the author experimentally demonstrated that the location and shape of the adversarial pattern had limited influence on the attack performance, e.g., the attack was valid even though the adversarial patch was far from the object. However, although the stealthiness of the generated adversarial examples, the constructed adversarial object is irrational to perform the attack in the real world as placing the traffic sign on the roadside without authorization would violate the traffic rule. 

\subsection{IAP}
Bai \etal \cite{bai2021inconspicuous} pointed out that the previous generative-based patch attack \cite{liu2019perceptual} involved large-scale training data and resulted in a conspicuous pattern. To address the issues, the author focused on generating an inconspicuous patch for traffic signs by only using one single image. Specifically, they first located the perceptual sensitivity region of the victim model, then trained a multiple-scale generator to synthesize the inconspicuous patch. Despite its effect, the training of multiple generators is time-cost and hardy.

Additionally, the license plate recognition (LPR) model has been demonstrated to be vulnerable to physical adversarial attacks \cite{qian2020spot}. Rather than optimize the perturbation over the while image pixel, Qian \textit{et al.} \cite{qian2020spot} found the best position where to paste the adversarial patch. Specifically, the author first cut out characters of the plate and selected the victim character. Then the position to be pasted with the rectangle black block is optimized by the genetic algorithm. The author revealed that different license plate characters show different attack sensitivity. To perform physical attacks, the author first collected the car plate from the real world, then optimize the optimal position for attack. After that, the author pasted the black sticker on the real car plate at the optimized location and captures the image for evaluation. Experiment results show the vulnerability of the LPR model to physical adversarial attacks.

In summary, the technique in developing an attack against the  Traffic sign recognition task can be described as follows: first, locate the traffic sign to be attacked in the given image; then control the position of the allowable region of adversarial perturbation, which usually realized by a binary mask. Finally, robustness should be considered to improve adversarial performance in the physical world. Although the existing method obtains certain success, these approaches mainly create the adversarial perturbation with a fixed binary mask with a rectangle shape; the adaptive mask and deformable shape would further boost the attack performance and delivery attack method.

\section{Physical Adversarial Attack on Object Detection Task}
\label{sec:object_det}
Object detection is a technique to locate and classify the object in the image, which is widely applied in video surveillance and automatic driving, where the target object is the vehicle, pedestrian, traffic sign, and so on. Existing object detectors can be categorized into single-stage detectors and two-stage detectors. The former output the detection results directly and typically include the YOLO series\cite{redmon2016you,redmon2018yolov3}, SSD \cite{liu2016ssd} and RetinaNet \cite{lin2017focal}. In contrast, the latter first extracts the feature with respect to input via a backbone and then performs classification and regression over the feature to obtain the categories and bounding boxes. The typically two-stage detector includes Faster RCNN \cite{ren2015faster} and Master RCNN \cite{he2017mask}. Nonetheless, both two types of object detectors have three outputs: bounding boxes, objectness, and category. Bounding boxes locate the possible object in the image, where objectness gives the confidence that the located area contains the object, and category gives the classification results. 

Performing adversarial attacks against object detection at least needs to modify one of three outputs, making it more challenging than attacking the image recognition tasks. The main way to perform a physical adversarial attack against object detection is to modify the attributes (e.g., appearance) of the object, including the adversarial patch, adversarial camouflage, and other novel approaches (e.g., emitting the laser toward the object). Although the pixel-wise adversarial examples can also realize physical adversarial attacks by printing the image with adversarial perturbation, we ignore them because it is unrealistic for the adversary to modify the whole environment in practice when attacking the object detector. Table \ref{tab:overview_2} lists the corresponding physical attacks against image recognition tasks.

\begin{table*}[!h]
\setlength\tabcolsep{1pt}
\centering
\caption{Physical adversarial attacks against object detection tasks. We list them by time and task, aligning with the discussed order.}
\label{tab:overview_2}
\begin{threeparttable}
\begin{tabular}{ccccccccc}
\hline
\textbf{Method}                                           & \textbf{Year-Venue}          &  \textbf{\makecell{Adversarial's \\ Knowledge}} & \textbf{\makecell{Threat \\ Model}}  & \textbf{\makecell{Robust \\ Technique}}          & \textbf{\makecell{Physical \\Test Type}}   & \textbf{Space} & \textbf{Remark}         & \textbf{Code} \\ 
\midrule
AdvDetector\cite{lu2017adversarial}             & 2017-ArXiv           & White-box               & OD             & Illumination               & Dynamic              & 2D    & Pixel-wise     & $\times$    \\
ShapeShifter\cite{chen2018shapeshifter}         & 2018-ECMLPKDD        & White-box               & OD             & EOT                        & Dynamic              & 2D    & Pixel-wise     & \checkmark    \\
LPAttack\cite{yang2020beyond}                   & 2020-AAAI            & White-box               & OD             & EOT,TV,NPS                 & Static               & 2D    & Pixel-wise     & $\times$    \\
4AAttack\cite{jia2022fooling}                   & 2022-arXiv           & White-box               & OD             & EOT,D2P                   & Static               & 2D    & Pixel-wise     & $\times$    \\
RP2+\cite{song2018physical}                     & 2018-USENIX-W        & White-box               & OD             & TV,NPS,D2P                 & Static               & 2D    & Patch          & $\times$    \\
AdvPatch\cite{thys2019fooling}                  & 2019-CVPR-W          & White-box               & OD             & EOT,TV,NPS                 & Dynamic              & 2D    & Patch          & \checkmark    \\
NestedAE\cite{zhao2019seeing}                   & 2019-CCS             & White-box               & OD             & D2P, Alignment  		   & Dynamic              & 3D    & Patch          & $\times$    \\
DPatch2\cite{lee2019physical}                   & 2019-ArXiv           & White-box               & OD             & EOT                        & Static               & 2D    & Patch          & $\times$    \\
ScreenAttack\cite{hoory2020dynamic}             & 2020-ArXiv           & White-box               & OD             & TV                         & Static               & 2D    & Patch          & $\times$    \\ 
TranslucentPatch\cite{zolfi2021translucent}  & 2021-CVPR               & White-box               & OD             & Affine,NPS 				   & Static               & 2D    & Patch          & $\times$    \\
Daedalus\cite{wang2021daedalus}                 & 2021-IEEE TC         & White-box               & OD             & EOT,NPS 				   & Static               & 2D    & Patch          & $\times$    \\
SwitchPatch\cite{shapira2022attacking}		    & 2022-ArXiv            & White-box               & OD             & TV                         & Static               & 2D    & Patch          & $\times$    \\
AdvBgImage\cite{xu2022universal}		        & 2022-ACNS            & White-box               & OD             & NPS                         & Static               & 2D    & Patch          & $\times$    \\
UAVPatch\cite{wen2023light}				        & 2023-ICASSP          & White-box               & OD             & EOT,TV                      & Static               & 2D    & Patch          & $\times$    \\
T-PATCH\cite{zhu2023tpatch}				        & 2023-USENIX Security & White-box               & OD             & EOT                         & Static               & 2D    & Patch          & $\times$    \\
Invisible   Cloak1\cite{yang2018building}       & 2018-UEMCON          & White-box               & OD             & EOT,TV                     & Static               & 3D    & Wearable & $\times$    \\
Adversarial   T-Shirt\cite{xu2020adversarial}   & 2020-ECCV            & White-box               & OD             & EOT,TPS                    & Static               & 2D    & Wearable  & $\times$    \\
Invisible   Cloak2\cite{wu2020making}           & 2020-ECCV            & White-box               & OD             & TPS                        & Static               & 2D    & Wearable  & \checkmark    \\
LAP\cite{tan2021legitimate}                     & 2021-ACM MM          & White-box               & OD             & TV,NPS                     & Static               & 2D    & Wearable  & $\times$    \\
NaturalisticPatch\cite{hu2021naturalistic}      & 2021-ICCV            & White-box               & OD             & TV                         & Static               & 2D    & Wearable  & \checkmark    \\
AdvTexture\cite{hu2022adversarial}              & 2022-CVPR            & White-box               & OD             & EOT,TPS                    & Static               & 2D    & Wearable & \checkmark    \\ 
T-SEA\cite{huang2023t}				            & 2023-CVPR			   & White-box               & OD             & EOT, CutOut                 & Static               & 2D    & Wearable       & $\times$    \\
AdvART\cite{guesmi2023advart}				    & 2023-ArXiv		   & White-box               & OD             & EOT, TV                     & Static               & 2D    & Wearable       & $\times$    \\
MeshAdv\cite{xiao2019meshadv}                   & 2019-CVPR            & White-box               & OD             & -                          & Static               & 3D    & Shape          & $\times$    \\ 
CAMOU\cite{zhang2019camou}                      & 2019-ICLR            & White-box               & OD             & EOT                        & Static               & 2D    & Camouflage     & $\times$    \\
ER\cite{wu2020physical}                         & 2020-arXiv           & Black-box               & OD             & -                          & Static               & 3D    & Camouflage     & $\times$    \\
UPC\cite{huang2020universal}                    & 2020-CVPR            & White-box               & OD             & EOT                        & Static               & 3D    & Camouflage     & \checkmark    \\
DAS\cite{wang2021dual}                          & 2021-CVPR            & White-box               & OD             & TV                         & Static               & 3D    & Camouflage     & \checkmark    \\
FCA\cite{wang2022fca}                           & 2022-AAAI            & White-box               & OD             & TV                         & Static               & 3D    & Camouflage     & \checkmark    \\
CAC\cite{duan2022learning}                      & 2022-IJCAI           & White-box               & OD             & EOT                        & Static               & 3D    & Camouflage     & $\times$    \\
DPA\cite{suryanto2022dta}                       & 2022-CVPR            & White-box               & OD             & EOT                        & Static               & 3D    & Camouflage     & $\times$    \\ 
ASA\cite{zhang2023boosting}                     & 2022-PR              & White-box               & OD             & -                          & Static               & 3D    & Camouflage     & \checkmark    \\
AdvCaT\cite{hu2023physically}                   & 2023-CVPR            & White-box               & OD             & EOT,TPS                    & Static               & 3D    & Camouflage     & \checkmark    \\
SLAP\cite{lovisotto2021slap}                    & 2021-USENIX Security & White-box               & OD             & EOT,TV                     & Static               & 2D    & Optical        & \checkmark  \\ \hline
BulbAttack\cite{zhu2021fooling}                 & 2021-AAAI            & White-box               & IR OD          & EOT,TV                     & Static               & 2D    & Optical        & $\times$    \\
QRAttack\cite{zhu2022infrared}                  & 2022-CVPR            & White-box               & IR OD          & EOT                        & Static               & 2D    & Patch          & $\times$    \\ 
HOTCOLD Block\cite{wei2023hotcold}              & 2023-AAAI            & Black-box               & IR OD          & -                          & Static               & 2D    & Patch          & $\times$    \\ 
InfraredPatch\cite{wei2023physically}         & 2023-CVPR            & White-box               & IR OD          & -                          & Static               & 2D    & Patch          & \checkmark    \\ \hline
Type ON-OFF Attack\cite{du2022physical}               & 2022-WACV            & White-box               & AI OD           & EOT,TV,NPS                 & Dynamic              & 2D    & Patch          & \checkmark    \\
UAVAttack\cite{zhang2022adversarial}            & 2022-Remote Sensing  & White-box               & AI OD           & EOT,TV,NPS                 & Static               & 2D    & Patch          & $\times$    \\ 
AP-PA\cite{lian2022benchmarking}                & 2022-TGRS            & White-box               & AI OD           & EOT,TV,NPS                 & Static               & 2D    & Patch          & \checkmark    \\ \hline
\end{tabular}
\begin{tablenotes}
\footnotesize
\item[*] {\textbf{OD}: Object Detection. \textbf{IR OD}: Infrared Object Detection. \textbf{AI OD}: Aerial Images Object Detection.}
\end{tablenotes}
\end{threeparttable}
\end{table*}

\subsection{Patch-based physical adversarial attack}
Patch-based physical attacks mainly focus on generating the adversarial patch for the plane object. The most remarkable advantage is that it is easily deployable in physical attacks as the adversary can perform attacks by only printing it out and hanging it. Such a characteristic makes it be suitable carrier for perturbation in physical attacks.

Lu \etal \cite{lu2017adversarial} first investigated the adversarial attack against the object detector by training the adversarial texture of the stop sign on physically collected video frames. They optimize the adversarial patch by minimizing the mean score instead of the maximum score, as it can bring more gains. Moreover, the average illumination intensity of the stop sign is adopted to constrain the optimized texture for naturalness. However, the physical attack result is not satisfying. 

Chen \textit{et al.} \cite{chen2018shapeshifter} first demonstrated that the EOT could be applied to improve the robustness of adversarial patches in attacking two-stage object detectors (i.e., Faster RCNN \cite{ren2015faster}). The optimization based on C\&W attack \cite{carlini2017towards} is formulated as follows.

\begin{equation}
\operatorname*{argmin}_{x_{adv}\in \mathbb{R} ^{h\times w\times 3}} \mathbb{E}_{x \sim X, t \sim \mathcal{T}} [ \frac{1}{m} \sum_{r_i \in rpn(M_t(x_{adv}))} L_{f_i}(M_t(x_{adv}), y)] + c \cdot \|\tanh(x_{adv}) - x\|^2_2,
\label{eq:shapeshift}
\end{equation}
where $M_t(x_{adv}) = M_t(x, {\rm tanh}(x_{adv}))$ is a function that exerts transform $t$ on the object ${\rm tanh}(x_{adv})$, and then paste it on the background image $x$ with a binary mask; $rpn$ indicates the region proposal extracted by Faster RCNN at the first stage; the $L_{f_i}(M_t(x_{adv}), y)$ is the loss function of classification in the $i$-th region proposal; the last term is a regularization term that is used to balance the attack strength and the quality of the adversarial patch. In physical attacks, the author conducted extensive static and dynamic attacks under multi-view conditions, and the result verified their method.

Following up the physical adversarial against image recognition model \cite{eykholt2018robust},  Eykholt \textit{et al.} \cite{song2018physical} extended the RP2 algorithm to attack the object detector. Specifically, they modify the adversarial loss for adapting to attack object detectors under two strategies creation and disappearance attack. According to the observation that the detector tends to focus more on contextual areas for prediction, thus they take the object's position and size into account, which is realized by performing rotation and transformation. Additionally, they find the $L_\infty$ lead to a pixelated perturbation that hurts the attack performance. Then they adopt the TV loss instead of $L_\infty$ to address this issue. The final optimization objective is expressed as follows.
\begin{equation}
\operatorname*{argmin}_{\delta}~~ \mathbb{E}_{x_i \sim X} J(f(x_i + t(M_x \cdot \delta)), y) + \lambda \cdot TV(M_x \cdot \delta ) + NPS,
\label{eq:eykholt_v2}
\end{equation}
where $J(\cdot, y)$ is the adversarial loss function devised for the object detector, which considers the classification output. In physical attacks under indoor and outdoor conditions, the author recorded the video of the adversarial object beginning away from 9 meters for the disappearance attack. By contrast, in the creation attack setting, they placed the adversarial patch on the wall or cupboard and captured the video from toward the adversarial object, away from 3 meters. The experiment result shows that the adversarial patch can mislead the YOLOv2 in most cases, demonstrating its fragility in practice. In addition, the experiment suggests that the indoor attack performs well than the outdoor attack, which can be attributed to the complex environment in the outdoor conditions.

Zhao \textit{et al.} \cite{zhao2019seeing} systemically investigated the adversarial patch against two detectors (i.e., YOLOv3 and Faster RCNN) in both digital and physical environments. Specifically, the author first exploited the intermediate features to improve the transferability of adversarial examples. To enhance the physical adversarial attack, the author constructed a physical dataset by collecting the real-world background and traffic sign image, respectively; then, the traffic sign image is embedded into the background image. Furthermore, the author adopted the EOT to guarantee the robustness of the adversarial patch. In physical attacks, the author places the camera device inside the car and toward the printed adversarial patch. They evaluated different distances between the car and the adversarial patch by controlling the car's speed (i.e., $6km/h \sim 30km/h$), the attack success rate over 72\%. 

Unlike scrawling the object with the adversarial patch at the fixed position for misclassified or misdetection, Lee \textit{et al.} \cite{lee2019physical} introduced an approach to craft the adversarial patch that can place anywhere (even far away from the object) in the image, and suppress all the detected objects. Specifically, the author customized the adversarial loss against the YOLO serial detectors and applied the EOT on images represented in HSV color space to improve the physical robustness. In physical attacks, the author printed out the adversarial patch and recorded the video under natural lighting conditions. The YOLOv3 is adopted to evaluate the effect of physical attacks. Experiment results show that the patch is invariant to location. However, when placing the adversarial patch away from the object, the size of the patches needs to be enlarged.

Rather than crafting the adversarial patch pasted on the object's surface, Zolfi \textit{et al.} \cite{zolfi2021translucent} instead proposed a novel adversarial attack that crafts the adversarial patch, which was stuck on the camera lens. To avoid the camera lens being overlapped by the adversarial patch, the author took the image's alpha channel into account during the optimization. Moreover, The NPS loss is also adopted to ensure the printability of the adversarial patch. The author conducted the physical attack by first printing out the adversarial patch on transparent paper and placing it on the camera's lens, then capturing the video. The physical attack evaluated on YOLOv2/5 and Faster RCNN achieves the average fooling rate of 35.48\%.

Recently, Shapira \textit{et al.} \cite{shapira2022attacking} developed a targeted attack method to generate the adversarial patch that is pasted on the car hood to attack object detectors (i.e., YOLOv3 and Faster RCNN) in a more realistic monitor scenario. Specifically, the author devised a tailored projection method to adaptively adjust the position and shape of the adversarial patch in terms of the camera orientation. To performs the targeted attack, the author considers all candidates that surpass the specific threshold before the NMS component to ensure the relevant candidates can be classified as the target label. Moreover, TV loss is adopted to improve the physically deployable of the adversarial patch. In physical attacks, the author stuck the printed patch on the toy car, achieving an attack success rate of 95.9\% on recaptured images.

\begin{figure}[t]
	\centering
	\begin{minipage}{.18\linewidth}
		\centering
		\includegraphics[width =1\linewidth]{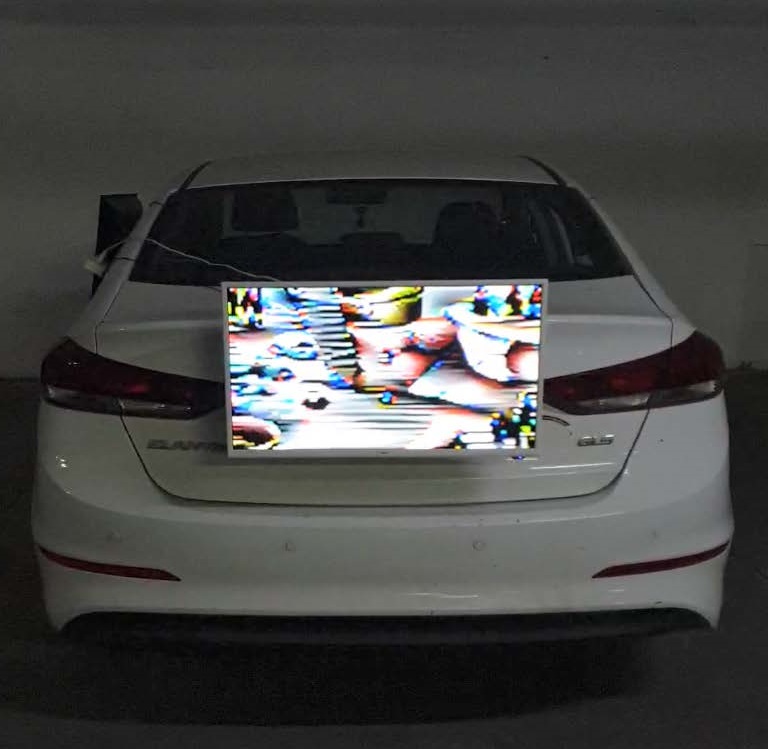}
		\centerline{\footnotesize \cite{hoory2020dynamic}}
	\end{minipage}
	\begin{minipage}{.18\linewidth}
		\centering
		\includegraphics[width =1\linewidth]{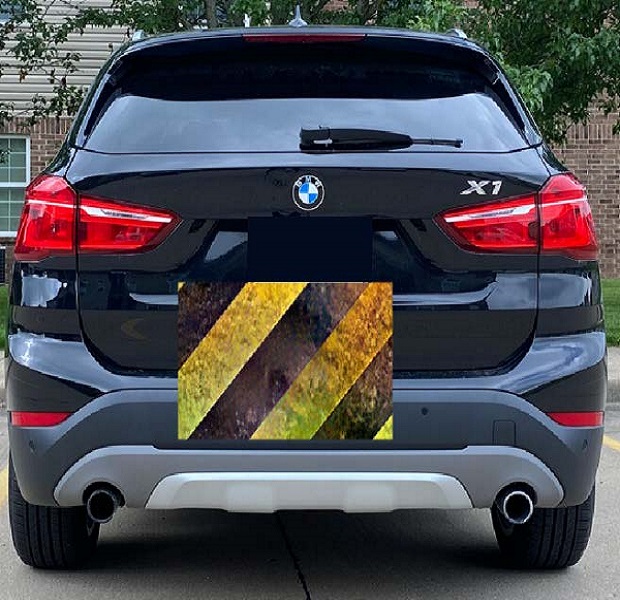}
		\centerline{\footnotesize \cite{cheng2022physical}}
	\end{minipage}
	\begin{minipage}{.18\linewidth}
		\centering
		\includegraphics[width =1\linewidth]{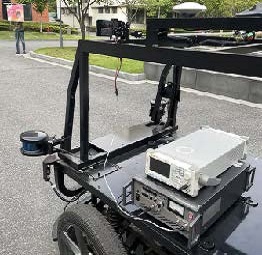}
		\centerline{\footnotesize \cite{zhu2023tpatch}}
	\end{minipage}
	\begin{minipage}{.18\linewidth}
		\centering
		\includegraphics[width =1\linewidth]{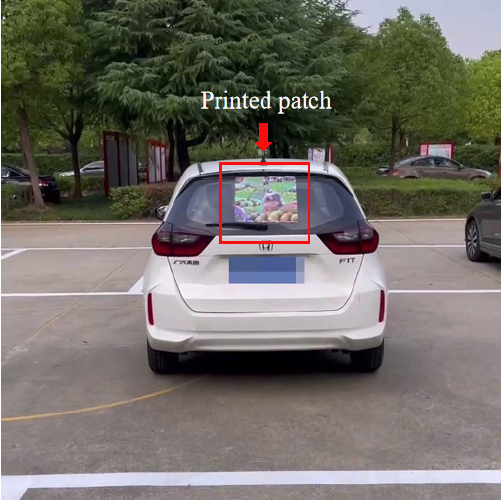}
		\centerline{\footnotesize \cite{wen2023light}}
	\end{minipage}
	\caption{Visualization of adversarial examples.}
	\label{fig:screen}
\end{figure}

The first step of current patch-based physical attacks is to print out the adversarial patch, which is only suitable for the static scenario in practice, ignoring the constantly changing character of the physical world. To address this challenge, Hoory \textit{et al.} \cite{hoory2020dynamic} proposed a dynamic adversarial patch for evading object detection models by displaying the adversarial patch with a digital screen (see Figure \ref{fig:screen}). The author adopted the loss function as the same to \cite{thys2019fooling} to train the adversarial patch. Additionally, the author pointed out that the adversarial patch crafted by loss function \cite{thys2019fooling} is easily misdetected to the semantically related category, which makes it inconsistent with background objects. To this end, the author considered the semantically-related class (e.g., bus or truck) of the original class (i.e., car) as the target and devised the following loss function

\begin{equation}
J(\delta, x_{adv}, C) = \sum_{y_i \in C}\mathcal{L}(f(x_{adv}),y_i) + \lambda \cdot TV(\delta),
\end{equation}
where $C$ is the set of semantically-related labels used to avoid being detected as car-related classes. Furthermore, the author split the train set into multi subsets and trained an adversarial patch for each subset. In physical attacks, the author placed a digital screen on the side and rear of the car, where the adversarial patch is chosen from the patch subsets to dynamic display in terms of the changing environmental conditions. Evaluation results on YOLOv2 suggested that 90\% video frames with the adversarial perturbed screen can not detect the target object (i.e., car).

Unlike sticking the adversarial patch on the target object to perform physical adversarial attacks, Xu \etal \cite {xu2022universal} placed the target object on the printed optimized adversarial background image to mislead the object detector. Despite its novelty but the actual use scene is not clear. Jia \etal \cite{jia2022fooling} devoted to physically fooling the automatic car equipped with object detectors, they extended the EOT with the transformation of random blur caused by the camera shake and resolution changing derived by car moving. Recently, Zhu \etal \cite{zhu2023tpatch} proposed a novel patch-based physical attack triggered by acoustic signals, which is sent to impact the imaging process of the capture camera. The patch works only when the acoustic signal is sent, or else the patch is no harm. Wen \etal \cite{wen2023light} used the projector carried by a drone to project the optimized adversarial patch toward the rear windshield to fool the vehicle detection model. To accurately project the adversarial patch on the windshield, the author took the perspective deformation, double image, and partial reflectance of light caused by photoing process and the glass material into account during optimization. Nonetheless, such an attack is easily perceived by the victim. 

In addition to object detection on common object, Yang \textit{et al.} \cite{yang2020beyond} proposed to generate the adversarial license plate to fool the SSD. Specifically, the author devised a mask to guide the perturbation region, optimized by the C\&W adversarial loss and color regularization consisting of NPS and TV loss to ensure adversarially and eliminate the color discrepancy between the digital and physical worlds. In optimization, EOT is used to enhance physical robustness. In physical attacks, the author created the adversarial license plate and placed it on the real car. Experiment results suggested that the detection performance on captured images average degraded 76.9\% in indoor and outdoor conditions.

\subsection{Wearable adversarial perturbation}
\begin{figure}[h]
	\centering
	\begin{minipage}{0.14\linewidth}
		\centering
		\includegraphics[width =1\linewidth]{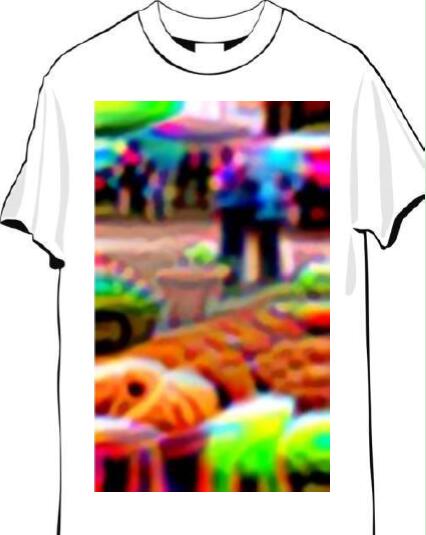}
		\centerline{\footnotesize \cite{yang2018building}}
	\end{minipage}
	\begin{minipage}{0.14\linewidth}
		\centering
		\includegraphics[width =1\linewidth]{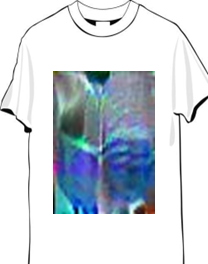}
		\centerline{\footnotesize \cite{wang2019advpattern}}
	\end{minipage}
	\begin{minipage}{0.14\linewidth}
			\centering
			\includegraphics[width =1\linewidth]{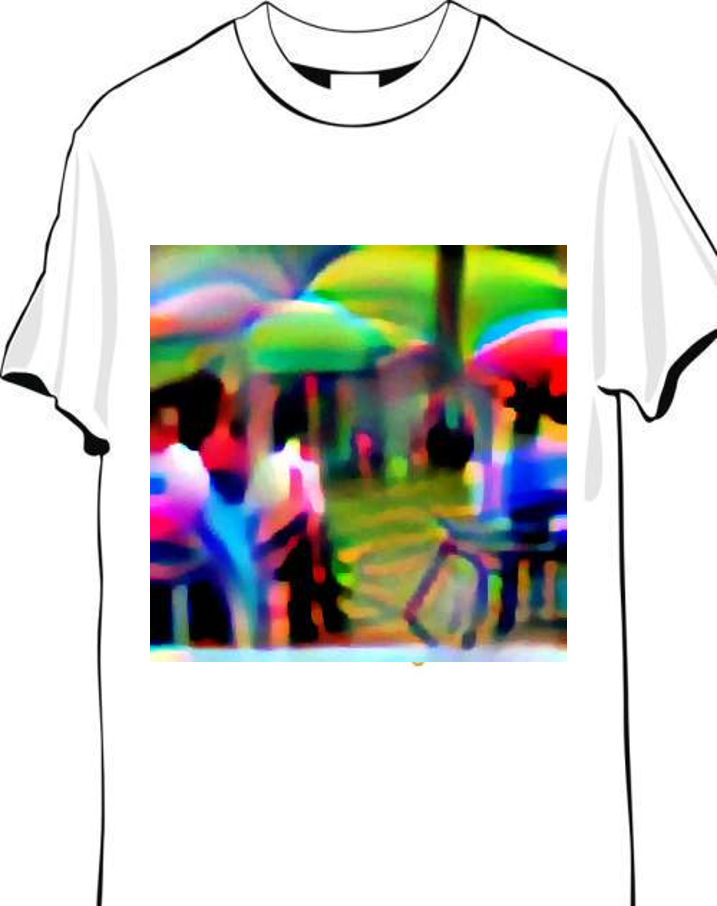}
			\centerline{\footnotesize \cite{thys2019fooling}}
		\end{minipage}
	\begin{minipage}{0.14\linewidth}
		\centering
		\includegraphics[width =1\linewidth]{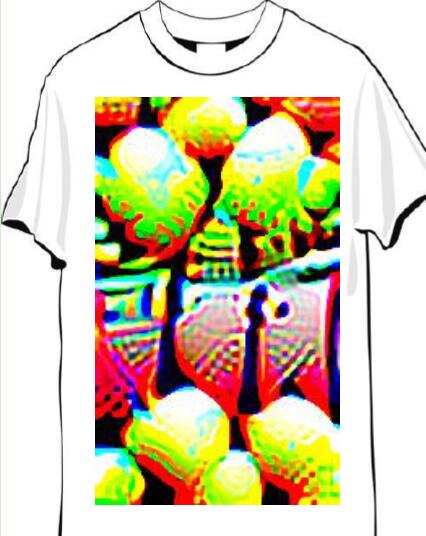}
		\centerline{\footnotesize \cite{xu2020adversarial}}
	\end{minipage}
	\begin{minipage}{0.14\linewidth}
		\centering
		\includegraphics[width =1\linewidth]{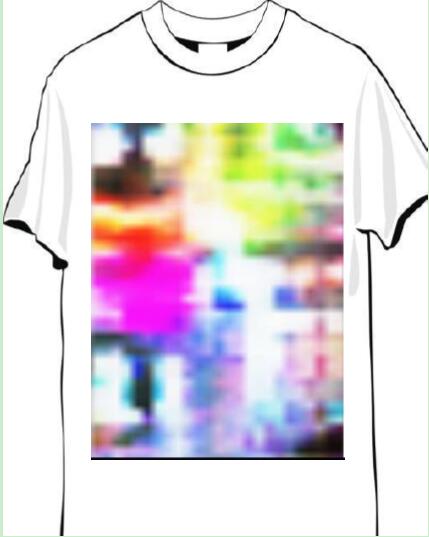}
		\centerline{\footnotesize \cite{wu2020making}}
	\end{minipage}
	\begin{minipage}{0.14\linewidth}
		\centering
		\includegraphics[width =1\linewidth]{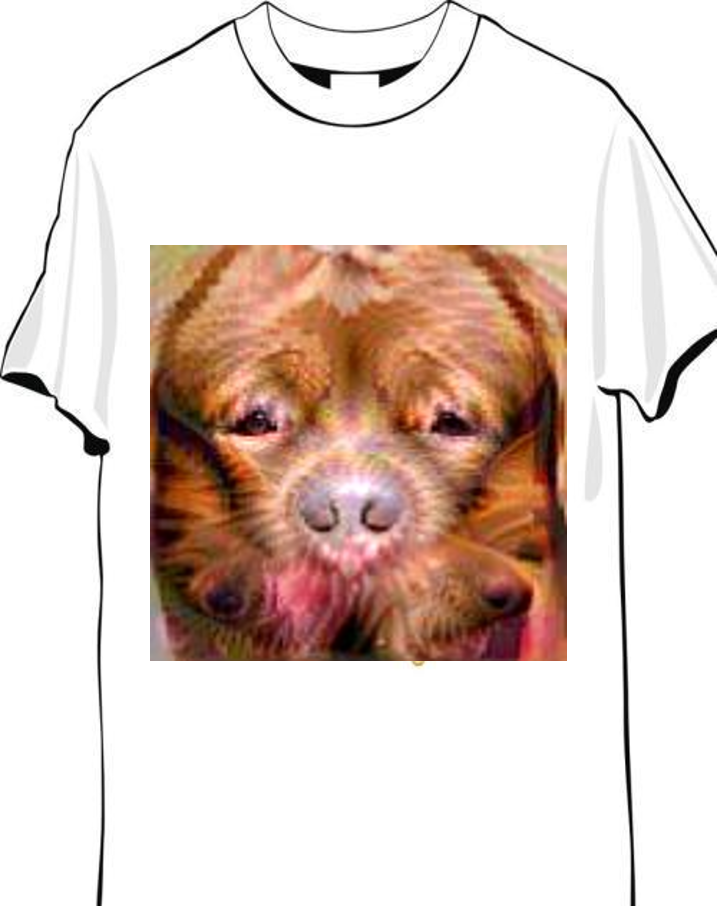}
		\centerline{\footnotesize \cite{huang2020universal}}
	\end{minipage}
	
	\begin{minipage}{0.14\linewidth}
		\centering
		\includegraphics[width =1\linewidth]{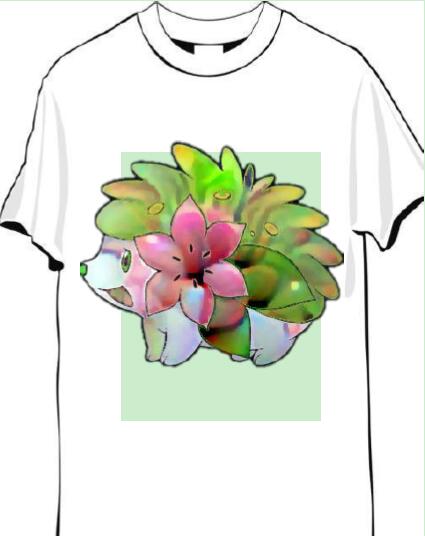}
		\centerline{\footnotesize \cite{tan2021legitimate}}
	\end{minipage}
	\begin{minipage}{0.14\linewidth}
		\centering
		\includegraphics[width =1\linewidth]{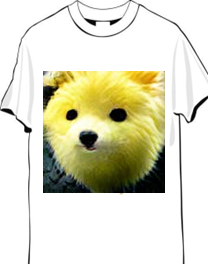}
		\centerline{\footnotesize \cite{hu2021naturalistic}}
	\end{minipage}
	\begin{minipage}{0.14\linewidth}
		\centering
		\includegraphics[width =1\linewidth]{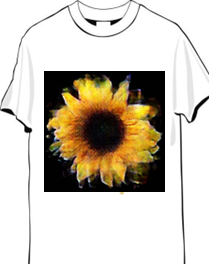}
		\centerline{\footnotesize \cite{doan2022tnt}}
	\end{minipage}
	\begin{minipage}{0.14\linewidth}
		\centering
		\includegraphics[width =1\linewidth]{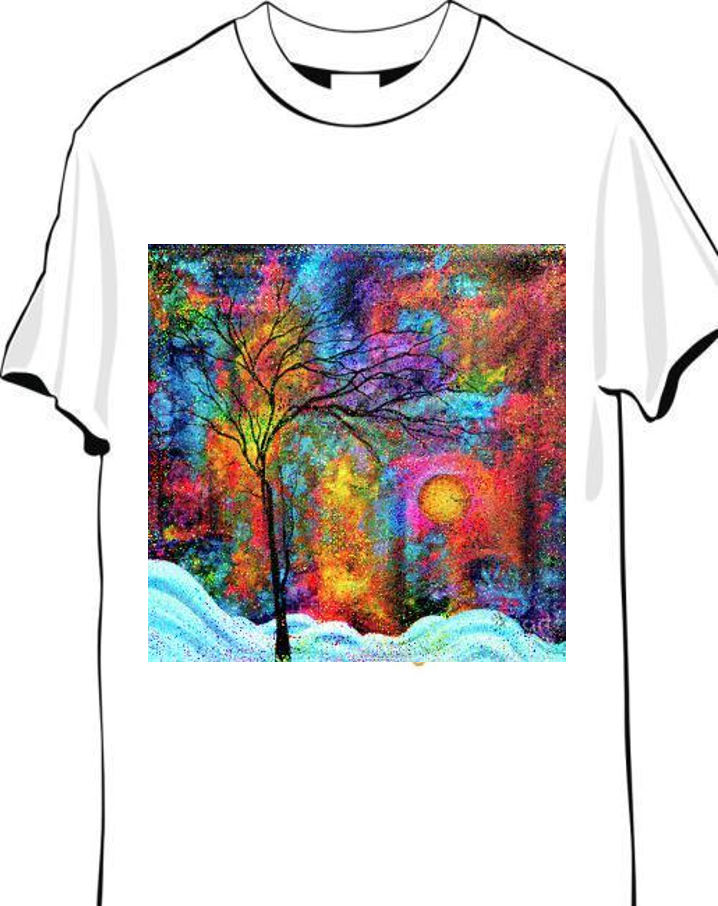}
		\centerline{\footnotesize \cite{guesmi2023advart}}
	\end{minipage}
	\begin{minipage}{0.14\linewidth}
		\centering
		\includegraphics[width =1\linewidth]{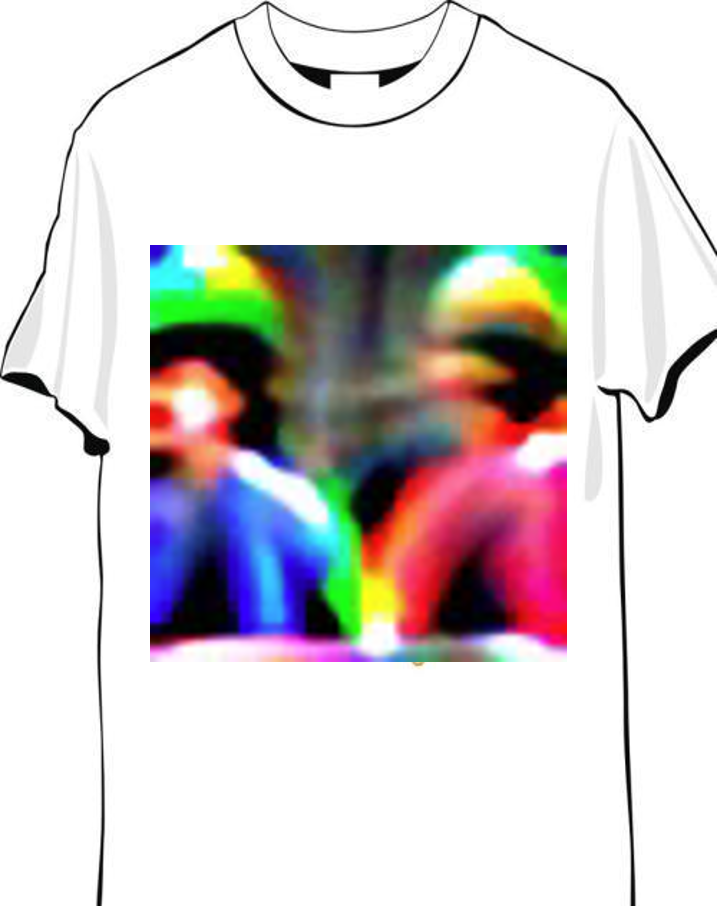}
		\centerline{\footnotesize \cite{huang2023t}}
	\end{minipage}
	\caption{Visualization examples of various wearable adversarial patches.}
	\label{fig:adv_clothes}
\end{figure}

Compared with the simple plane (non-rigid) object, the non-rigid object (i.e., person) shows more difficulty as its exhibits the deformable character of the target object. Wearable adversarial perturbation approaches \cite{yang2018building,xu2020adversarial,xu2020adversarial,wu2020making} were designed to solve the problem. Figure \ref{fig:adv_clothes} illustrates some wearable patch-based adversarial examples.

Yang \textit{et al.} \cite{yang2018building} extended EOT from 2D space to 3D space by building up a pinhole camera model to mimic the patch's transformation in the physical world. In optimization, the author adopted the C\&W-based adversarial loss, TV loss, and Euclidean distance between the clean image and the adversarial image with patch to optimize the adversarial patch. The author performs the physical attack by holding the laptop that displays the adversarial patch. Experiment results on the captured image show that the YOLO's detection performance degraded from 100\% to 28\%.

Thys \textit{et al.} \cite{thys2019fooling} proposed to generate adversarial patches for intra-class variety (i.e., persons). Specifically, they aimed to evade the detection of object detection applications by holding the printed adversarial patch cardboard plate. In optimization, the author designed an algorithm to place the adversarial patch in the center of the detected object, then adopted the objectness output as the adversarial loss, which experimentally shows it superior to the classification output; NPS and TV loss are also used to guarantee the printability and naturalness of the adversarial patch. To improve the physical robustness of the adversarial patch, the author adopted a set of random transformations to the optimization process. Experiment results demonstrated that the generated adversarial patch could hide the person in front of the surveillance camera (based on YOLOv2) in most cases.

Xu \textit{et al.} \cite{xu2020adversarial} devised an adversarial T-shirt to fool the pedestrian object detector in the physical world. The generated adversarial pattern is attached to the surface of the T-shirt to construct the adversarial T-shirt. To mimic the cloth deformation in the physical world, the author applied the Thin Plate Spline Mapping (TPS) transformation over the adversarial perturbation. Specifically, the author adopted the following three approaches to enhance the performance of the adversarial attack: 1) approximate the cloth's deformation with TPS; 2) approximate the color discrepancy between digital and physical space with a quadratic polynomial regression. 3) approximate physical transformation with EOT. By performing the above operations, the T-shirt stuck with the adversarial patch degrades the person detector significantly under indoor and outdoor physical conditions.

Concurrent to \cite{xu2020adversarial}, Wu \textit{et al.} \cite{wu2020making} proposed to devise an adversarial patch to implement the disappear attack. To this end, the author devised the objectness disappear loss, which is expressed as follows

\begin{equation}
\mathcal{L}_{obj} = \mathbb{E}_{t\sim \mathcal{T}} \sum_{i}\max \left\{S_i(A(x, \delta, t)) +1 ,0\right\}^2,
\end{equation}
where $A(x, \delta, t)$ is the patch apply function, which pastes the adversarial patch $\delta$ on the image with the transformation process $t$; $S_i$ indexes the objectness score of $i$-th proposal. Additionally, the TV loss is adopted to ensure the visual effect. By minimizing this loss function, the adversarial examples with the devised patch could suppress the positive scores. Moreover, the author empirically finds that the TPS would detriment the adversarial performance.

Recently, Huang \etal \cite{huang2023t} focused on enhancing the transferability of patch-based attacks by devising a novel ensemble strategy by randomly dropping a subset of layers of the network. Moreover, the author performed the augmentation consisting of the cutout on the adversarial patch and constrained data transformation to further enhance the patch's robustness.

In addition to the above methods, some researchers attempted to make natural adversarial patches or devoted to making the adversarial clothes with the optimized texture, which will be discussed separately.

\subsubsection{Natural-orient methods}

The existing works mainly focus on attack performance, ignoring the imperceptible of the adversarial patch. Huang \textit{et al.} \cite{huang2020universal} proposed to craft the universal physical adversarial camouflage (UPC) with the natural image initialization, engendering a visual inconspicuous patch. Moreover, a set of transformations is performed on the adversarial patch to mimic deformable properties. Then, the author devised specific two-stage attack procedures against Faster RCNN: the RPN attack and the classifier and regressor attack, which are implemented by the following loss functions.

\begin{equation}
\begin{split}
\mathcal{L}_{rpn} = &\mathbb{E}_{p_i \sim \mathcal{P}}(\mathcal{D}(s_i, s_t) + s_i\|\vec{b_i} - \nabla \vec{b_i}\|_1), \\
\mathcal{L}_{cls} = &\mathbb{E}_{p_i \sim \hat{\mathcal{P}}}f_{cls}(p)_y + \mathbb{E}_{p_i \sim \mathcal{P^\star}}\mathcal{L}(f_{cls}(p), y), \\
\mathcal{L}_{reg} = &\sum_{p_i \sim \mathcal{P^\star}} \|f_{reg}(p)-\nabla \vec{b}\|_2,
\end{split}
\label{eq:upc}
\end{equation}
where $s_i$ and $\vec{b_i}$ are the confidence score and coordinates of $i$-th bounding box; $s_t$ is the target score, where $s_1$ and $s_0$ represent the background and foreground. $\nabla \vec{b_i}$ is the predefined vector that guides the optimization direction. $f_{cls}$ and $f_{reg}$ are the prediction output of the classifier and the regressor. $\mathcal{P}$ is the output proposal, $\hat{\mathcal{P}}$ is the top-k proposal filtered by confidence score and $\mathcal{P}^\star$ is the proposal with respect to the true label $y$. Additionally, the TV loss is also adopted to produce more natural patterns. In physical attacks, the author printed the adversarial patterns and stuck them over the person or car (i.e., non-plane objects) and captured the picture from different distances (8 $\sim$ 12 m) and angles ($-45^\circ \sim 45^\circ$) and the results indicate their adversarial camouflage could reduce the detector's performance by a large margin.

To construct the inconspicuous adversarial patch, Tan \textit{et al.} \cite{tan2021legitimate} proposed two-stage strategies to optimize the legitimate adversarial patch that can escape both human eyes and object detector models in the physical world. In the first stage, the author began with a random cartoon image and optimized an adversarial patch that could deceive the detector. In the second stage, the optimized patch crafted in the first stage is required to be closer to the clean cartoon image for the purpose of evading human attention.  Moreover, TV loss and the Euclidean distance between the adversarial patch and the cartoon image were utilized to ensure its naturalness; NPS loss is used to confine the patch color to printable colors. The author conducted the physical attack by attaching the printed adversarial patch to the white T-shirts, and the result demonstrated their adversarial patch can be successful physically against YOLOv2.

Concurrently to \cite{tan2021legitimate}, Hu \textit{et al.} \cite{hu2021naturalistic} exploited the learned image manifold of a pre-trained GAN (e.g., BigGAN and StyleGAN) to guide the generation of the adversarial patch. Rather than directly optimizing the adversarial patch from scratch, the author treats the image sample sampling from the generator as the adversarial patch. Then the patch is applied to the target object inside the image to construct the adversarial examples. The TV loss, and the adversarial loss that calculates the maximum of objections score and classification probabilistic are used to guide the patch's optimization. Physical evaluation of the printed patch show that the physically recaptured image could degrade the detection recall of YOLOv4tiny by approximately  23.80\% and 43.14\% (for indoor and outdoor, respectively). Moreover, they exhaustly investigate the influence of different transformations, patch sizes, and target classes on attack performance.

Guesmi \etal \cite{guesmi2023advart} followed the common pipeline of patch-based attack \cite{thys2019fooling} but devoted to synthesizing the natural-looking adversarial patch by using the natural image as initialization, and a similarity loss is used to constrain it.

\subsubsection{Texture-orient methods}

Hu \textit{et al.} \cite{hu2022adversarial} got inspiration from cloth making and proposed to treat the adversarial patch as the whole cloth to be cut. To generate a naturalistic patch, the author utilized the generative model to generate the adversarial patch and then overlap them with the object. To simulate the physical deformation of the cloth, the author applied the TPS transformation over the adversarial patch. TV loss is adopted to ensure the smoothness of the patch's pattern. In their recent work \cite{hu2023physically}, they focused on generating natural clothes texture by initializing it with camouflage texture, which is optimized by leveraging the differentiable-modified Voronoi diagram. The optimized camouflage is then rendered and wrapped on the 3D object. In physical attacks, the author made real clothes in terms of the optimized texture.

\begin{figure}[h]
	\centering
	\begin{minipage}{.20\linewidth}
		\centering
		\includegraphics[width =1\linewidth]{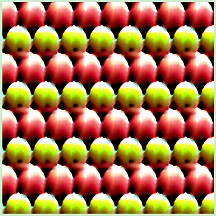}
		\centerline{\footnotesize (a) \cite{hu2022adversarial}}
	\end{minipage}
	\begin{minipage}{.20\linewidth}
		\centering
		\includegraphics[width =1\linewidth]{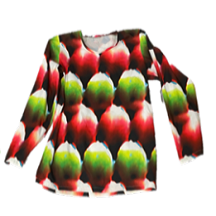}
		\centerline{\footnotesize (b) \cite{hu2022adversarial}}
	\end{minipage}
	\begin{minipage}{.20\linewidth}
		\centering
		\includegraphics[width =1\linewidth]{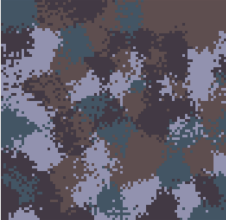}
		\centerline{\footnotesize (c) \cite{hu2023physically}}
	\end{minipage}
	\begin{minipage}{.20\linewidth}
		\centering
		\includegraphics[width =1\linewidth]{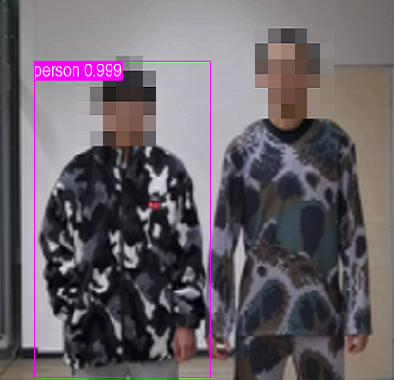}
		\centerline{\footnotesize (d) \cite{hu2023physically}}
	\end{minipage}
	\caption{Adversarial Clothes. Adversarial material (a, c) v.s. adversarial clothes (b,d).}
	\label{fig:advclothes}
\end{figure}

In summary, the existing wearable adversarial attack aims to improve stealthiness and robustness against multi-view by utilizing the natural image prior and full coverage pattern. Some texture-orient methods \cite{hu2022adversarial,hu2023physically} attempt to generate the texture of clothes, making it valid in multi-view angles. Therefore, devising the make sense and natural adversarial clothes requires further study.

\subsection{Camouflage-based physical adversarial attack}
\begin{figure}[h]
	\centering
	\begin{minipage}{.16\linewidth}
		\centering
		\includegraphics[width =1\linewidth]{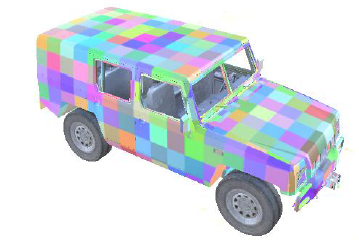}
		\centerline{\footnotesize Random}
	\end{minipage}
	\begin{minipage}{.16\linewidth}
		\centering
		\includegraphics[width =1\linewidth]{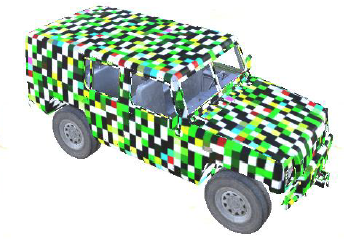}
		\centerline{\footnotesize CAMOU \cite{zhang2019camou}}
	\end{minipage}
	\begin{minipage}{.16\linewidth}
		\centering
		\includegraphics[width =1\linewidth]{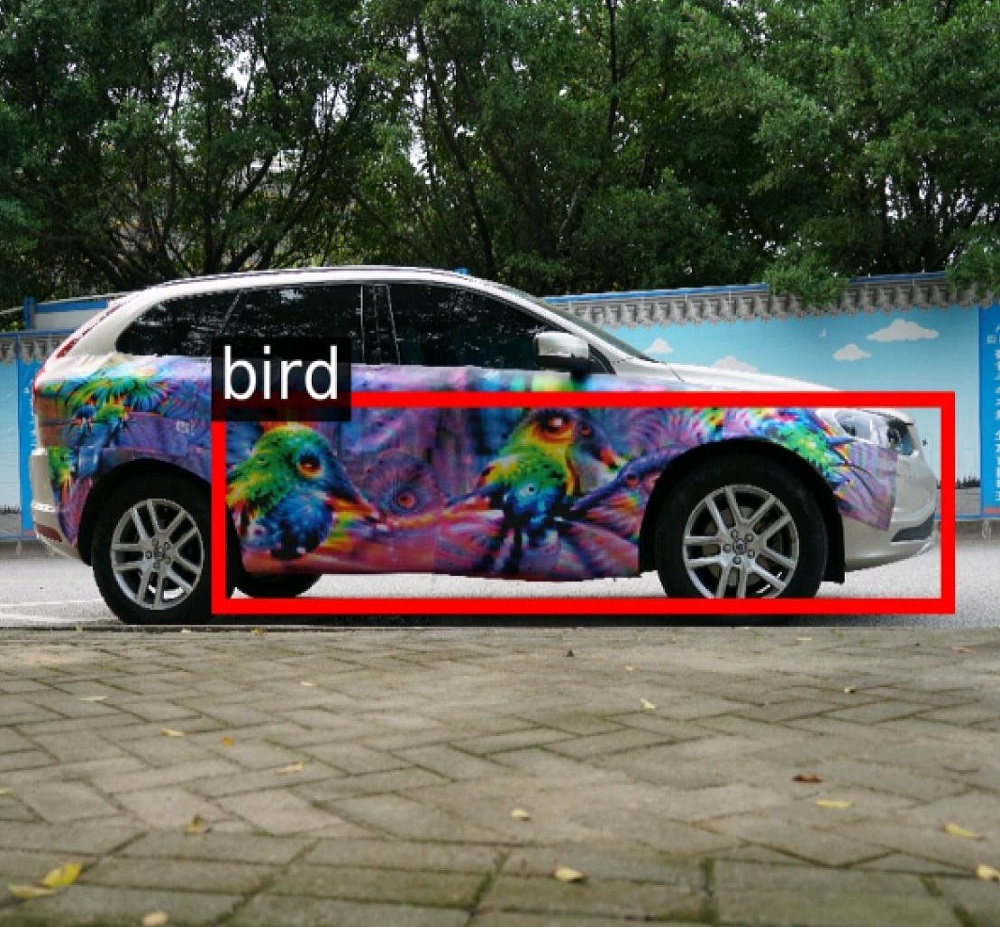}
		\centerline{\footnotesize UPC \cite{huang2020universal}}
	\end{minipage}
	\begin{minipage}{.16\linewidth}
		\centering
		\includegraphics[width =1\linewidth]{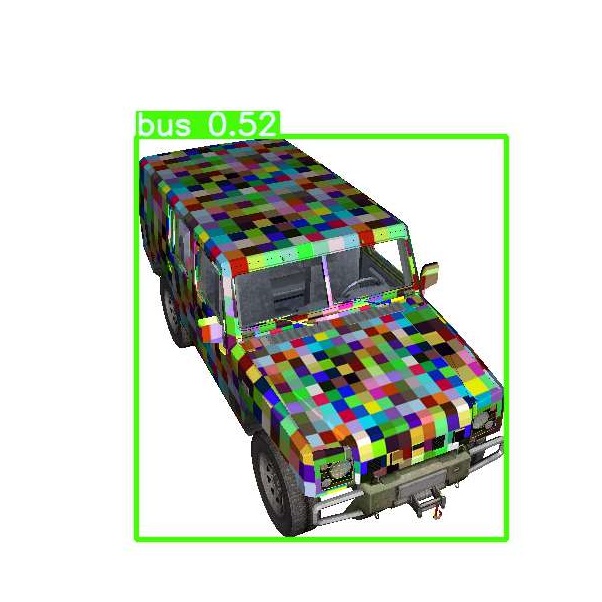}
		\centerline{\footnotesize ER \cite{wu2020physical}}
	\end{minipage}
	\begin{minipage}{.16\linewidth}
		\centering
		\includegraphics[width =1\linewidth]{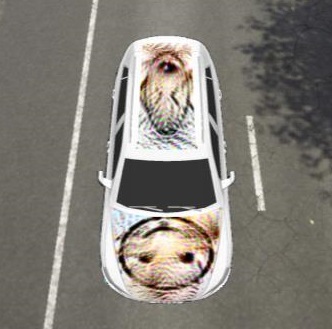}
		\centerline{\footnotesize DAS \cite{wang2021dual}}
	\end{minipage}
	
	\begin{minipage}{.16\linewidth}
		\centering
		\includegraphics[width =1\linewidth]{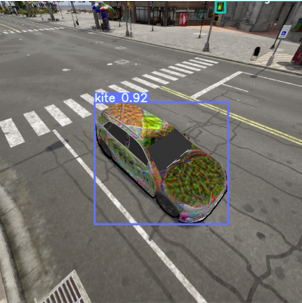}
		\centerline{\footnotesize FCA \cite{wang2022fca}}
	\end{minipage}
	\begin{minipage}{.16\linewidth}
		\centering
		\includegraphics[width =1\linewidth]{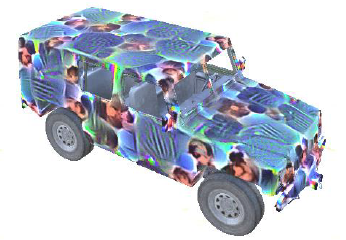}
		\centerline{\footnotesize CAC \cite{duan2022learning}}
	\end{minipage}
	\begin{minipage}{.16\linewidth}
		\centering
		\includegraphics[width =1\linewidth]{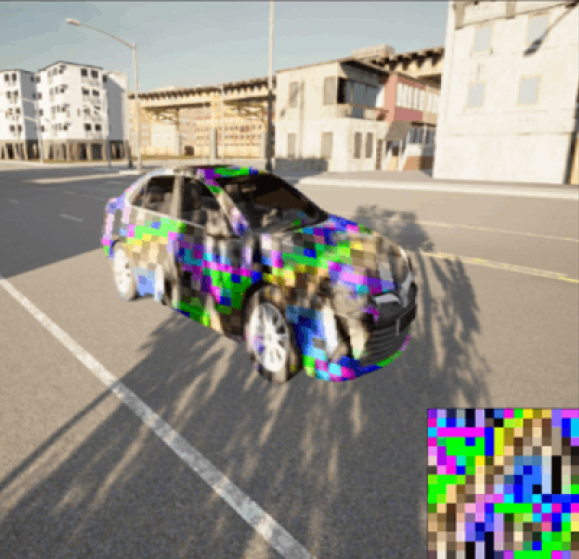}
		\centerline{\footnotesize DTA \cite{suryanto2022dta}}
	\end{minipage}
	\begin{minipage}{.16\linewidth}
		\centering
		\includegraphics[width =1\linewidth]{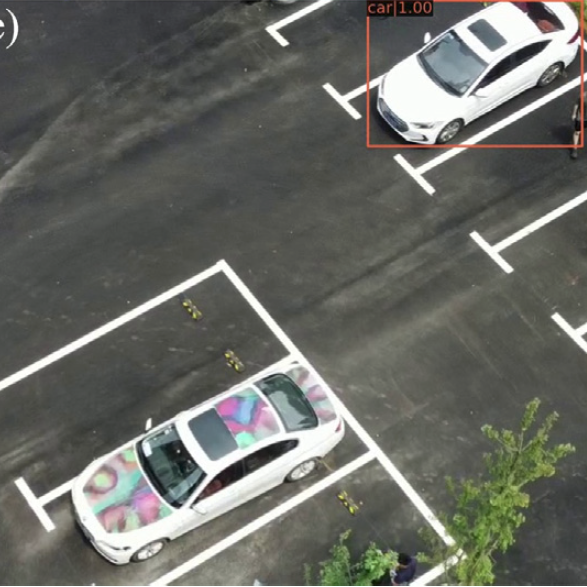}
		\centerline{\footnotesize CAC \cite{zhang2023boosting}}
	\end{minipage}
	\caption{Examples of various camouflage textures rendered vehicles.}
	\label{fig:varoious_camou}
\end{figure}

The works discussed above are mainly focused on generating the adversarial patch on the 2D image, where each pixel can be modified independently. Recently, researchers have increasingly paid attention to more realistic attacks in 3D space, developing adversarial camouflage-based attacks that focus on modifying the appearance (i.e., shape or texture) of the 3D object. Figure \ref{fig:varoious_camou} provides examples of camouflage-based attacks.

\subsubsection{Shape-orient methods}
Zeng \textit{et al.} \cite{zeng2019adversarial} first proposed a method to generate adversarial examples against 3D shape classification and a visual question answer (VQA) system. Specifically, they crafted the adversarial example by adopting the neural renderer technique, which allows them to perform the 3D transformation to better approximate real-world environments. Concurrent to Zeng \cite{zeng2019adversarial}, Xiao \textit{et al.} \cite{xiao2019meshadv} developed an approach to generate an adversarial object by modifying the mesh of a 3D object, namely MeshAdv. The adversarial mesh is optimized on the loss consisting of classification loss, objectness loss, and perceptual loss. Moreover, the author extends the TV loss from pixel-wise in the 2D image to the vertex in the 3D image. 

\subsubsection{Texture-orient methods}
Unlike the works \cite{zeng2019adversarial,xiao2019meshadv} optimize the object's mesh, Zhang \textit{et al.} \cite{zhang2019camou} proposed to optimize the texture of the 3D object. The author presented a clone network to mimic the entire processing containing the adversarial texture render and detector prediction. Then apply the white-box attack on the clone network to optimize the adversarial texture. Although the camouflaged vehicle achieved good attack performance on Mask RCNN, the optimized texture is a mosaic-like pattern, which is conspicuous to the human observer. 

Huang \textit{et al.} \cite{huang2020universal} optimized the adversarial texture from the patch seed and stick the adversarial patch over the partial vehicle's surface to obtain well appearance. It is worth noting that the author conducted the experience by printing out the adversarial texture and sticking it on the real car. 

Rather than construct the adversarial patch under the white-box setting, Wu \textit{et al.} \cite{wu2020physical} considered a more challenging black-box attack. The author utilized the genetic algorithm to optimize the adversarial texture. To address the dimension disastrous in optimizing the adversarial texture, the author instead optimized small-size adversarial texture ($64 \times 64$), then enlarged or repeated (ER) to a large scale ($2048 \times 2048$), dramatically reducing the optimization variable. The adversarial texture is rendered wrapped over the vehicle in the simulator environment via Carla \cite{dosovitskiy2017carla} PythonAPI, then recorded the adversarial examples from the simulator environment and used to query the detector (i.e., Light head RCNN \cite{li2017light}). They showed the adversarial texture generated by the black-box attack could significantly reduce the performance (i.e., 52\%) of the detector in the simulator environment but ignore evaluation in the physical environment.

Wang \textit{et al.} \cite{wang2021dual} pointed out that the adversarial camouflage generated by the previous works \cite{zhang2019camou,wu2020physical} ignored the model's attention, resulting in worse transferability. To this end, the author proposed simultaneously suppressing the model and human attention. To suppress the model's attention, the author proposed dispersing the intensity attention map by disconnecting the connected graph generated from the attention map. To suppress human attention, the author adopted the seed content patch as initialization and encouraged the adversarial texture to remain similar to it. Moreover, the author optimizes the specific faces of the 3D model as the textures via neural renderer \cite{kato2018neural}. To optimize the adversarial texture, the author constructed a dataset sampled from the Carla simulator at different distances and angles. The author conducts the physical attack by printing out the camouflaged rendered car on paper and then cropping the camouflage texture and sticking it on the toy car. Experiment results suggested that the average performance reduction of four detectors (i.e., YOLOv5, SDD, Faster RCNN, and Mask RCNN) is 26.05\%.

Take further step than \cite{wang2021dual}, Wang \textit{et al.} \cite{wang2022fca} proposed to craft the full-coverage adversarial camouflage texture to overcome more complex environment conditions (e.g., multi-view and occlusion), which are less explored in the current patch-like \cite{thys2019fooling} and partial camouflage work \cite{huang2020universal,wang2021dual}. To maintain the effectiveness of the small ratio of adversarial camouflaged vehicles in the image, the author used multi-scale IoU loss to constrain the adversarial texture. Finally, the author combined the TV loss and adversarial loss (i.e., including IoU, classification, and objections) to optimize the adversarial camouflage texture. They adopt a similar physical attack process with \cite{wang2021dual}, and the experiment shows their method's effectiveness under complex environments, degrading the average reduction of the detector performance of 53.99\% in physical attacks.

Unlike \cite{wang2022fca} optimize the full coverage faces of the 3D model that encountered the difficult issue of hard implementation, Duan \textit{et al.} \cite{duan2022learning} proposed to optimize the 2D texture of the 3D model directly. They proposed to optimize the texture by minimizing all classification output of the RPN module at the first stage of Faster RCNN. Although they adopted various physical settings (e.g., angles, brightness, etc.), the author trained the texture on the image with pure-color background. In physical attacks, the author print out the entire vehicle model with the 3D printer, and the experiment demonstrates their method's effectiveness in the physical environment under $360^\circ$ free viewpoint settings.  

Suryanto \textit{et al.} \cite{suryanto2022dta} pointed out the previous works \cite{wang2021dual,wang2022fca} fail to consider more specific physical environmental conditions, e.g., shadow. To this end, the author used a network to approximate the shadowing effect. Rather than adopt the neural renderer to optimize the adversarial texture, the author optimizes the 2D image as the texture but transforms them with the same camera matrix as the vehicle. In physical attacks, the author printed out the camouflaged 3D object with a 3D printer and captured the photo every $45^\circ$. Experiment results suggested that the average reduction of two detectors (i.e., EfficientDetD0 \cite{tan2020efficientdet} and YOLOv4) model is 47.5\%.

Zhang \etal \cite{zhang2023boosting} improved the transferability of adversarial texture-based physical attack against object detector on UAV angle by dispersing the fusion attention map of the augmented input, but the effective on monitor angle is not evaluated.

In summary, the camouflage-based adversarial texture has attracted increasing attention in recent years because it can better mimic physical transformations (e.g., shadowing and brightness). For example, recently, Byun \textit{et al.} \cite{byun2022improving} proposed to adopt the neural renderer as the data augmentation technique to boost the robustness of adversarial examples. Additionally, these works conducted the physical attack by printing a small-size car model rather than the real car. It is far away from a realistic attack on the vehicle. Alcorn \textit{et al.} \cite{alcorn2019strike} argued that the reason for the vulnerability of detectors might be attributed to these simulated images being strange to the detector. In other words, the detector was not trained on such simulated data, resulting in bad generalization. Moreover, there are many issues to be solved, including but not limited to small-size adversarial objects and the complex physical environment condition (e.g., occlusion and lighting) in the real world.

\subsection{Projector's projection}
Unlike the adversarial patch, Lovisotto \textit{et al.} \cite{lovisotto2021slap} presented a novel adversarial attack by modulating the projector's parameters. The author first collected the projectable colors as the following steps: 1) collect the image of the target object as the projection surface $S$; 2) choose a color $c_p=[r,g,b]$, which is projected over the projection surface, resulting in the output $O_{c_p}$. 3) repeat the above two steps until collecting sufficient diversified data. Moreover, the author took the camera noise and the projector's work mechanism into account to collect as realistic color as possible. After that, the author fitted a projection model as follows

\begin{equation}
\mathcal{L}_{\mathcal{P}} = \operatorname*{argmin}_{w} \sum_{\forall c_s, c_p} \|\mathcal{P}(c_s, c_p) - c^{(s,p)}_o\|_1,
\end{equation}
where $\mathcal{P}$ is the projection model, ${c_s, c_p, c^{(s,p)}_o}$ is the training data triple, consists of projecting $c_p$ on pixels of color $c_s$, and the output $c^{(s,p)}_o$; $w$ is the parameters of the projection model. To optimize the adversarial examples, the author combined the EOT and treated the objectness and classification output of the detector as the adversarial loss. The author conducted the physical attack by placing the projector away from the target object from 1 to 12 meters and viewing angles $-30^\circ \sim 30^\circ$, achieving the success rate 100\% for the traffic sign classification model and 77\% for object detectors (i.e., Mask RCNN and YOLOv3).

\subsection{Attack infrared image detector}
\begin{figure}[h]
	\centering
	\begin{minipage}{.2\linewidth}
		\centering
		\includegraphics[width =1\linewidth]{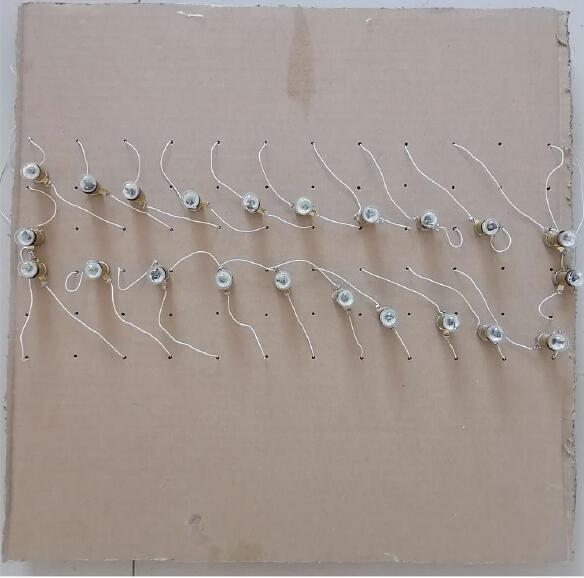}
		\centerline{\footnotesize  BulbAttack \cite{zhu2021fooling}}
	\end{minipage}
	\begin{minipage}{.2\linewidth}
		\centering
		\includegraphics[width =1\linewidth]{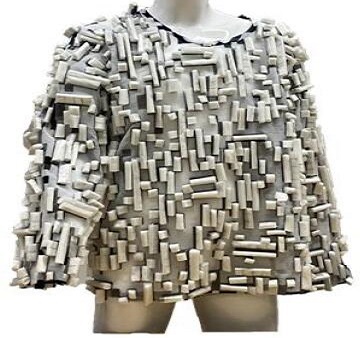}
		\centerline{\footnotesize QRAttack \cite{zhu2022infrared}}
	\end{minipage}
	\begin{minipage}{.2\linewidth}
		\centering
		\includegraphics[width =1\linewidth]{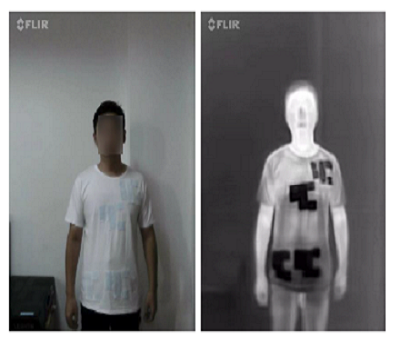}
		\centerline{\footnotesize HOTCOLD Block \cite{zhu2022infrared}}
	\end{minipage}
	\begin{minipage}{.2\linewidth}
		\centering
		\includegraphics[width =1\linewidth]{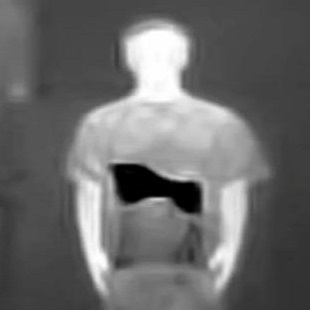}
		\centerline{\footnotesize InfraredPatch \cite{zhu2022infrared}}
	\end{minipage}
	\caption{Adversarial against infrared thermal imaging detector.}
	\label{fig:infrare}
\end{figure}

The existing physical adversarial attack mainly focused on the RGB image in visible light, while visible light has its limitations in dark environments (e.g., darkness); for example, the image content taken in the evening is hardly identified. However, infrared thermal imaging has its advantage because it is free from lighting changes, enabling them adopted widely in the real world. 

Zhu \textit{et al.} \cite{zhu2021fooling} proposed a novel adversarial method to attack the infrared thermal imaging detector. The author devised a board with multi bulbs, where the positions of bulbs are the optimization variables. To mimic the shining of the bulb, the author applied the gaussian function on the pixel of the bulb's position. Finally, the author treated the combination of objectness output of detector and TV loss as their loss function to optimize the adversarial patch. The author conducted the physical attack by placing the bulb in the optimized positions and photoed the people who held the board. Experiment results show that the adversarial bulb board degrades the detector by 34.48\%.

Although \cite{zhu2021fooling} demonstrated the fragility of the infrared detector, the attack merely works at a specific angle. To address such issues, their follow-up work \cite{zhu2022infrared} constructs the adversarial construct the adversarial clothes covered with adversarial texture made by aerogel. More specifically, the author adopted a similar idea with \cite{hu2022adversarial} to optimize the adversarial patch image. To simplify the physical implementation, the author regarded the generation parameter of the QR code as the optimization variable, which determines the different QR code textures. The generated QR code texture is tiled to texture with a large shape, and then the actual adversarial patch is randomly cropped from it and pasted on the target object in the image. The TPS and various transformations are used to improve the robustness of the adversarial patch. The author made the clothes by attaching the black block to the aerogel in terms of the optimized QR code texture. The participant who wore the adversarial patch could successfully bypass the detector (i.e., YOLOv3), where the detection performance degraded by 64.6\%.

Recently, Wei \etal \cite{wei2023hotcold} realized stealthy physical attacks against the infrared detector by wearing the warming paste and cooling paste inside the clothes, where the wearing position and pattern are optimized. Concurrent to \cite{wei2023hotcold}, Wei \etal \cite{wei2023physically} learned a deformable shape adversarial patch, which is physically constructed by aerogel material, and the paste position on the human body is also optimized.

\subsection{Attack aerial image detector}
Unmanned aerial surveillance has gained increasing attention with the development of drone techniques and object detection. Correspondingly, the adversarial attack against unmanned aerial surveillance aroused increasing attention in the military scenario. The conventional technique against unmanned aerial surveillance is to use the camouflage net, while a recent study shows that a partial imagery patch could deceive the unmanned aerial surveillance.

\begin{figure}[h]
	\centering
	\begin{minipage}{.18\linewidth}
		\centering
		\includegraphics[width =1\linewidth]{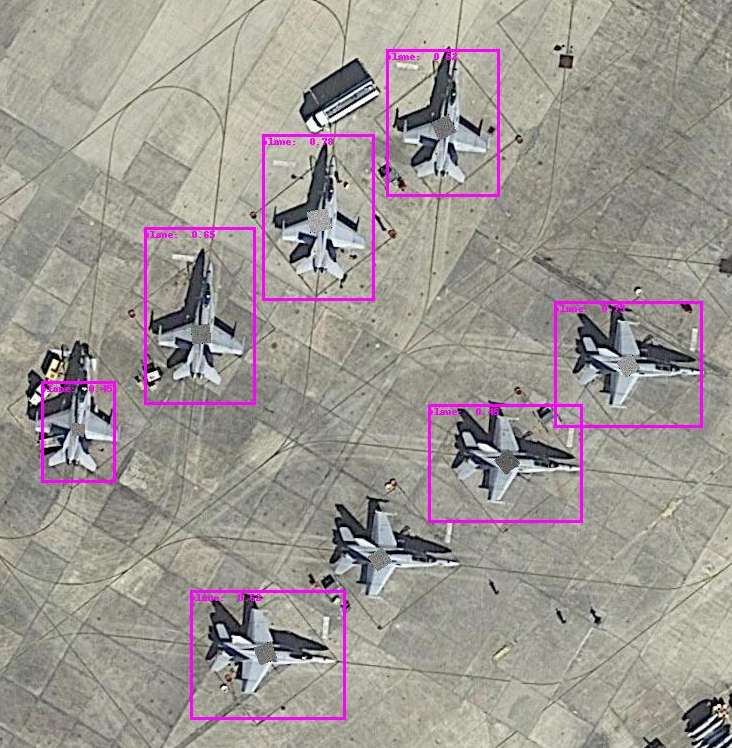}
		\centerline{\footnotesize  Random patch \cite{den2020adversarial}}
	\end{minipage}
	\begin{minipage}{.18\linewidth}
		\centering
		\includegraphics[width =1\linewidth]{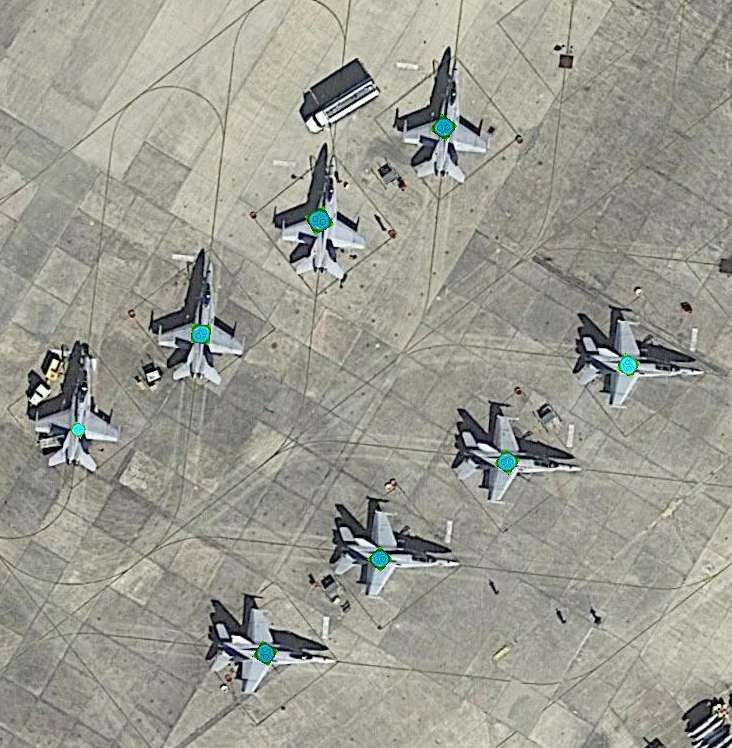}
		\centerline{\footnotesize Adversarial patch \cite{den2020adversarial}}
	\end{minipage}
	\begin{minipage}{.18\linewidth}
		\centering
		\includegraphics[width =1\linewidth]{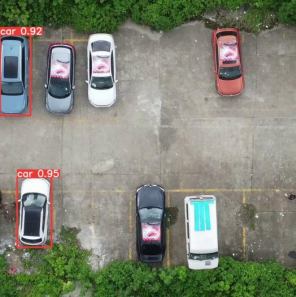}
		\centerline{\footnotesize  UAVAttack \cite{zhang2022adversarial}}
	\end{minipage}
	\begin{minipage}{.18\linewidth}
		\centering
		\includegraphics[width =1\linewidth]{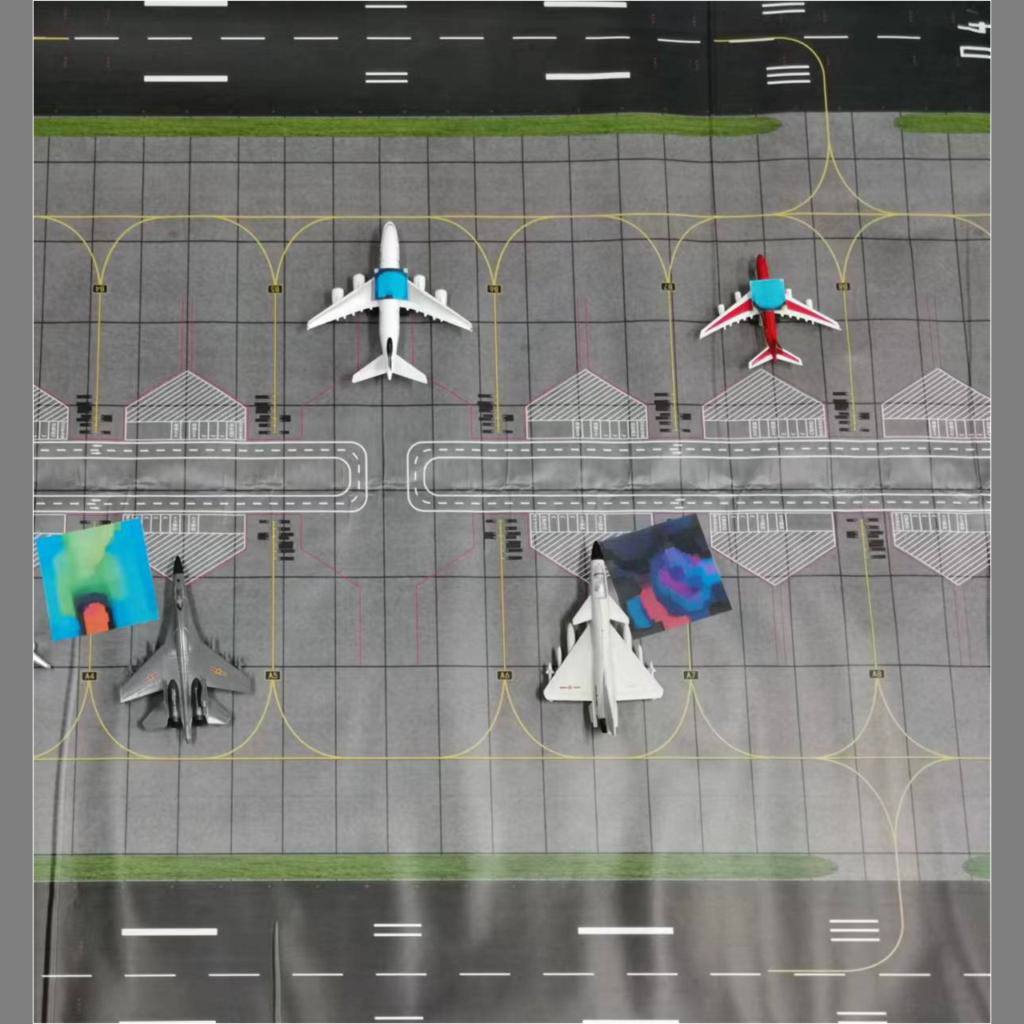}
		\centerline{\footnotesize AP-PA \cite{lian2022benchmarking}}
	\end{minipage}
	\begin{minipage}{.18\linewidth}
		\centering
		\includegraphics[width =1\linewidth]{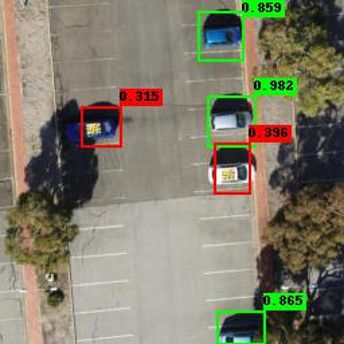}
		\centerline{\footnotesize Type ON-OFF Attack \cite{du2022physical}}
	\end{minipage}
	\caption{Visualization examples of adversarial patch against aerial image detector.}
	\label{fig:aerial_examples}
\end{figure}

Den Hollander \textit{et al.} \cite{den2020adversarial} assumed unmanned aerial surveillance is the YOLOv2 trained on the aerial imagery dataset. To attack the object detector, the author combined the common loss consisting of the adversarial loss, NPS loss, and the objectness loss. Moreover, the author proposed a novel colorfulness metric as a saliency loss to train the adversarial patch. The saliency loss is represented as follows

\begin{equation}
\begin{split}
\mathcal{L}_{sal} = \sqrt{\sigma_{rg}^2 + \sigma_{yb}^2} + 0.3 * \sqrt{\mu_{rg}^2 + \mu_{yb}^2} \\
rg = R - G, yb = 0.5 * (R + G) -B,
\end{split}
\end{equation}
where $\mu$ and $\sigma$ are the mean and standard deviation of the auxiliary variables, R, G, and B denote the patch's color channel values. The author investigated the impact of the patch's size, position, number, and saliency on the adversarial performance. Experiment results demonstrate that the optimized patch could reduce the detector performance by 47\%.

Recently, Du \textit{et al.} \cite{du2022physical} implement the physical adversarial attack against the aerial imagery detection. Considering the realistic scenario, the author designed two types of adversarial patches to accommodate the different scenarios. Type ON refers to the adversary placing the adversarial patch on the top of the car, which is suitable for dynamic scenarios (i.e., the car is driving). By contrast, Type OFF refers to the adversary placing the adversarial patch around the car, which is suitable for stationary scenarios (e.g., parking lot). The author adopted objectness, NPS, and TV loss to train the adversarial patch. The author conducted the physical attack by printing out the adversarial patch and placing them around or near the target objects under different lighting conditions. Experiment results demonstrated that the printed-out adversarial patch could significantly reduce objectness score (25\% to 85\%).

Concurrent to \cite{du2022physical}, Zhang \textit{et al.} \cite{zhang2022adversarial} focused on crafted adversarial patch attack against the object detection in the scenario of multi-scale objects of remote sensing. Specifically, the author first collected numerous aerial images by a drone, then optimized the adversarial patch under a set of transformations with the devised loss function, composed of the YOLO's training loss, TV loss, and NPS loss. In physical attacks, the author printed the adversarial patch with a size of $1.1m \times 1.1m$ and then placed it on the car's roof. Experimenting with the captured image with a drone suggested that they achieve a 30\% success rate ranging from 25 to 120 meters high. 

Lian \textit{et al.} \cite{lian2022benchmarking} proposed a general framework against various object detection models by extending the prior works \cite{thys2019fooling} to aerial images. On one side, the author devised a scale-adaptive by controlling the ratio of the patch size and the detected bounding box. On the other side, the author also placed the patch on top of the target object to suppress all objects in scenarios. In optimization, the author used the loss function consisting of mean objections score loss, TV loss, and NPS loss under the augmentation of a set of transformations. In physical attacks,  the author constructed a simulated environment by placing the plane stuck with the printed adversarial patch on the created airport sandbox. Experiment results suggested that the detection performance would significantly degrade under physical attacks. Figure \ref{fig:aerial_examples} illustrates the captured physical adversarial examples.

Apart from two adversarial patch attacks against the object detector on the aerial imagery dataset, there are a number of attempts to attack the remote sensing image recognition, and detection \cite{chen2019adversarial,chen2020attack,xu2020assessing}.

\section{Others}
\label{sec:other}
\begin{figure}[h]
	\centering
	\begin{minipage}{.22\linewidth}
		\centering
		\includegraphics[width =1\linewidth]{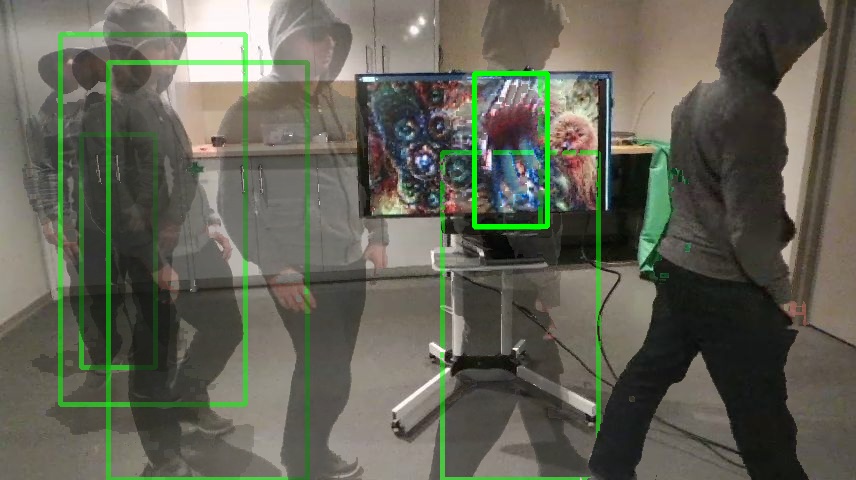}
		\centerline{\scriptsize PATAttack \cite{wiyatno2019physical}}
	\end{minipage}
	\begin{minipage}{.22\linewidth}
		\centering
		\includegraphics[width =1\linewidth]{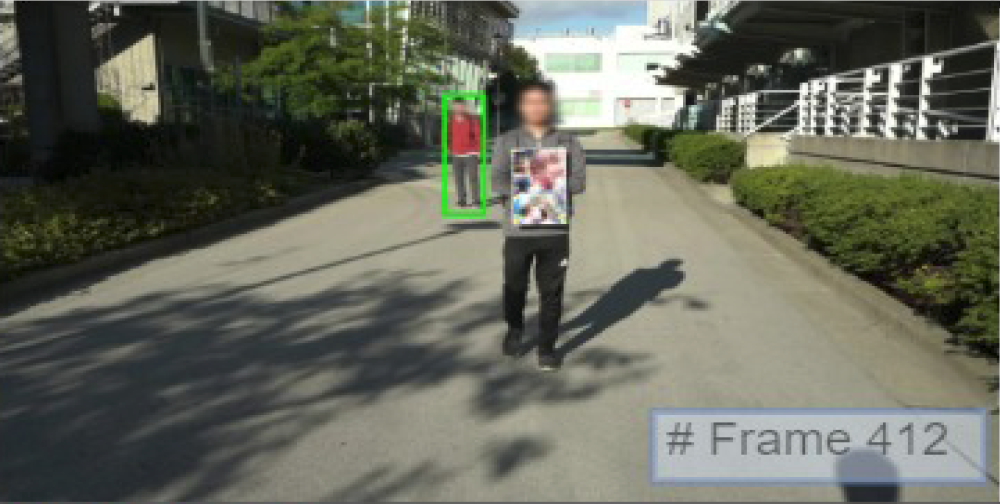}
		\centerline{\scriptsize MTD \cite{ding2021towards}}
	\end{minipage}
	\begin{minipage}{.22\linewidth}
		\centering
		\includegraphics[width =1\linewidth]{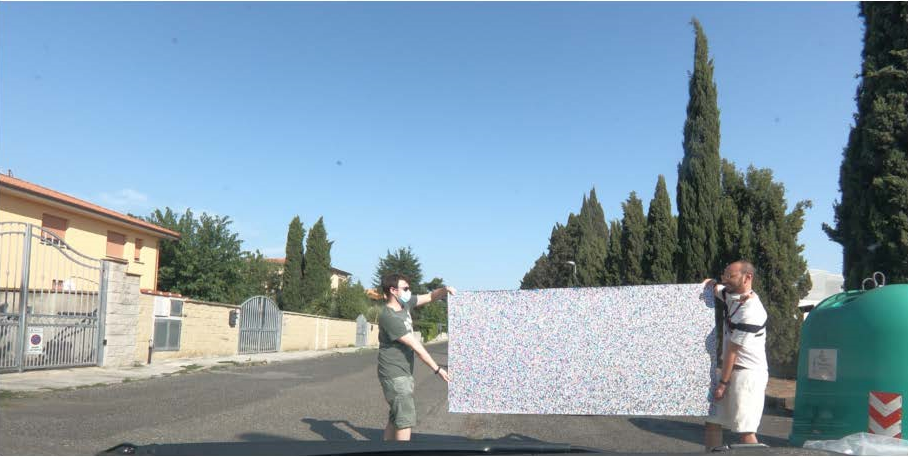}
		\centerline{\scriptsize RWAE \cite{nesti2022evaluating}}
	\end{minipage}
	
	\begin{minipage}{.25\linewidth}
		\centering
		\includegraphics[width =1\linewidth]{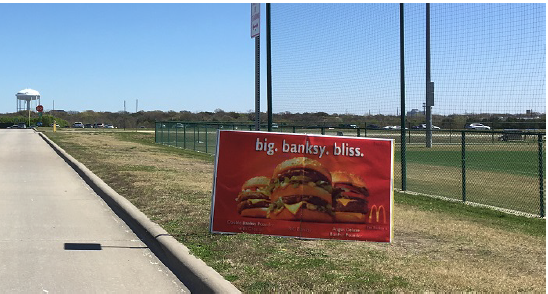}
		\centerline{\scriptsize PhysGAN \cite{kong2020physgan}} 
	\end{minipage}
	\begin{minipage}{.25\linewidth}
		\centering
		\includegraphics[width =1\linewidth]{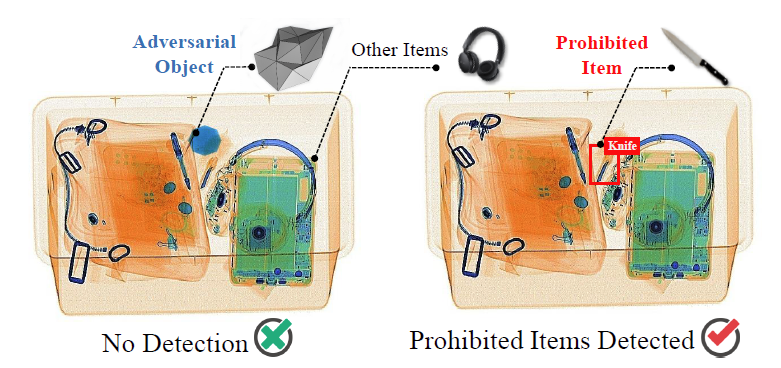}
		\centerline{\scriptsize X-rayAttack \cite{liu2023x}}
	\end{minipage}	
	\caption{Visualization adversarial example against other tasks.}
	\label{fig:others}
\end{figure}

This section reviews the literature besides the scope of image recognition and object detection, as the research field has less attention. Table \ref{tab:overview_3} lists the collected physical adversarial attack methods of other tasks. Figure \ref{fig:others} illustrates some physical adversarial examples. 

\begin{table*}[!h]
\setlength\tabcolsep{1pt}
\centering
\caption{Physical adversarial attacks against other tasks. We list them by time and task, aligning with the discussed order.}
\label{tab:overview_3}
\begin{threeparttable}
\begin{tabular}{ccccccccc}
\hline
\textbf{Method}                                           & \textbf{Year-Venue}          &  \textbf{\makecell{Adversarial's \\ Knowledge}} & \textbf{\makecell{Threat \\ Model}}  & \textbf{\makecell{Robust \\ Technique}}          & \textbf{\makecell{Physical \\Test Type}}   & \textbf{Space} & \textbf{Remark}         & \textbf{Code} \\ 
\midrule
AdvPattern\cite{wang2019advpattern}    & 2019-ICCV            & White-box          & ReID                  & EOT,TV             & Static               & 2D    & Patch           & \checkmark    \\ 
PhysGAN\cite{kong2020physgan}          & 2020-CVPR            & White-box          & SteeringModel         & D2P                & Dynamic              & 2D    & Pixel-wise     & $\times$    \\ \hline
PATAttack\cite{wiyatno2019physical}    & 2019-ICCV            & White-box          & Object Tracking       & EOT                & Dynamic              & 2D    & Patch          & $\times$   \\ 
MTD \cite{ding2021towards}             & 2021-AAAI            & White-box          & Object Tracking       & EOT,TV             & Dynamic              & 2D    & Patch          & $\times$   \\ \hline
RWAE\cite{nesti2022evaluating}         & 2022-WACV            & White-box          & Semantic Segmentation & EOT                & Static               & 2D    & Patch          & \checkmark    \\ \hline
MDEAttack\cite{cheng2022physical}      & 2022-ECCV            & White-box          & MDE                   & EOT                & Static               & 2D    & Patch          & $\times$   \\ \hline
X-rayAttack\cite{liu2023x}               & 2023-USENIX Security & White-box          & PID                   & -                  & Static               & 3D    & Mesh           & $\times$   \\ \hline
\end{tabular}
{\textbf{MDE}: Monocular depth estimation. \textbf{PID}: Prohibited Item Detection. }
\end{threeparttable}
\end{table*}

\subsection{Attack object tracking}
Compared to objection detection, object tracking is a task to track someone specified by a template, in which the people locked will constantly be located by the object tracker. Therefore, physical attacks against the object tracking system require adversarial perturbation that constantly works over time. 
Wiyatno \textit{et al.} \cite{wiyatno2019physical} presented a physical adversarial texture attack to make the object tracking model loss tracking target. Specifically, the author devised a set of adversarial loss functions against the object-tracking task. In addition, the perceptual similarity based on Euclidean distance is adopted to guarantee the imperceptible of the adversarial poster; EOT is adopted to improve physical robustness. In physical attacks, the displayed adversarial poster with a 50-inch TV monitor could make the object tracking system misbehave in an indoor condition with static lighting. Further, Ding \textit{et al.} \cite{ding2021towards} developed a physically deployable attack against single object tracking with the adversarial patch, which dilates or shrinks the detection bounding box with the adversarial patch.

\subsection{Attack semantic segmentation}
Semantic segmentation is similar to the object detection task but needs fine-grind classification for every pixel in the image. However, physical attacks against semantic segmentation are less explored, which will be discussed as follow.
\label{sec:ss}
Nesti \textit{et al.} \cite{nesti2022evaluating} proposed to optimize a scene-specific billboard patch to attack the semantic segmentation models. The author utilized the Carla simulator \cite{dosovitskiy2017carla} to simulate a more realistic real-world environment. Rather than adopt the EOT in the 2D image space, the author exploited various physical 3D transformations provided by Carla. More specifically, the author first applied the projective transformation with the extracted 3D rotation matrix to precisely paste the patch onto the target object's surface. Finally, the patch is optimized by the following devised loss function.

\begin{equation}
\mathcal{L}_M = \sum_{i \in \Omega} \mathcal{L}_{CE}(f_i(x_{adv}), y_i),
\mathcal{L}_{\bar{M}} = \sum_{i \not \in \Omega} \mathcal{L}_{CE}(f_i(x_{adv}), y_i),
\end{equation}
where $\mathcal{L}_M $ describes the cumulative cross-entropy loss for those pixels that have been misclassified with respect to the ground truth $y$, denoted as $\Omega$; while $\mathcal{L}_{\bar{M}}$ refers to all the others. The author printed out the billboard patch as a $1m \times 2m$  poster, which could successfully hide the target object. Figure \ref{fig:others} illustrates the physical billboard patch.

\subsection{Attack other tasks}
Additionally, physical adversarial attacks have been recently applied to other tasks, such as person re-identification (ReID), steering model \cite{kong2020physgan}, depth estimation \cite{cheng2022physical}, and X-ray prohibited item detection \cite{liu2023x}.

Wang \textit{et al.} \cite{wang2019advpattern} first attempted to develop the physical adversarial attack against the person re-identification (ReID). They devised a degradation function (i.e., a kind of data augmentation technique) that randomly changes the brightness or blurs of the image and then created an augmented generated set constructed by a multi-position sampling strategy, which boosts the physical attack performance under a multi-view. Unlike adopting EOT to improve the physical robustness of adversarial patches, Kong \textit{et al.} \cite{kong2020physgan} proposed a novel method to learn physical transformation from videos recorded in the real world by using a generative model (i.e., PhysGAN), which is then used to optimize the physical robust adversarial patch.
Similar to \cite{duan2020adversarial}, Cheng \etal \cite{cheng2022physical} leveraged the style transfer to synthesize a naturally styled camouflage patch, realizing an inconspicuous attack against a monocular deep estimation network.
Recently, the security of X-ray-based prohibited item detection system is raised attention. Liu \etal \cite{liu2023x} leveraged a neural renderer technique to optimize a 3D object that is constrained by a shape polishment method, engendering a reasonable 3D adversarial object. Additionally, the position of a 3D object in the suitcase is optimized by a policy-based reinforcement learning algorithm.

\section{Challenges and Future Work}
\label{sec:disccus}

Physical adversarial attacks impose potential risks for existing deployed DNN-based systems, appealing for much attention. Although a line of approach has been proposed, there still have improved space. In this section, we will summarize the common techniques used to develop robust physical attacks and their existing limitations from the perspective of the attack pipeline (i.e., input transformation, algorithm design, and evaluation). Structurally, we discuss the existing limitation and future work in separate paragraphs in the same subsection. Finally, we discuss the potentially positive application of the physical deployable adversarial perturbation.

\subsection{Input transformation}
Two represent robustness techniques during input transformation are EOT \cite{athalye2018synthesizing} and D2P \cite{jan2019connecting}. EOT is the most common technique to enhance the adversarial example's robustness both in the digital and physical world, which performs a set of predefined transformations during optimization. Recently, \cite{dong2022isometric} pointed out EOT can not cover all predefine transformations and easily results in performing poorly in some transformations. Thus,  MaxEOT \cite{dong2022isometric} improved EOT by using the worst-case (most harmful) transformations, making them work on all weaker transformations. However, the definition of worst-case transformations may impact the performance, and the search process is time intensity. In addition, existing approaches perform transformations on image space, which is impractical as the environment background is unchangeable as well as can not cover scale change or occlusion that occurs in the real world. The D2P method models all potential transformations involve in digital-to-physical and physical-to-digital and then trains the model with large-scale digital-physical image pair. Therefore, the transformations learned by the D2P model depend on the diversity of collected data and the training strategies. 

Based on the above discussion, some potential research directions can be derived: first, finding a more accurate definition of worst-case transformation that is suitable for physical adversarial attack and developing a more efficient search algorithm for worst-case. Second, developing a method to perform transformations on the object rather than the whole image, a promising method is to adopt the physical renderer. Third, collected larger diversity of digital-physical images and develop a pair-free D2P model, such as the style transfer can be performed to learn the latent style caused by different physical transformations. Finally, ensemble combining the above method may result in better robustness of physical attack.

\subsection{Attack algorithm design}
In addition to input transformation, the algorithm design is crucial to develop physical adversarial attacks. In this part, we discuss the limitation and potential future works from the form of adversarial perturbation and the generalization of physical adversarial attack.

\subsubsection{Form of adversarial perturbation}
The common form of physical adversarial perturbation against computer vision is patch-based, texture-based, and optical-based. Despite various approaches having been proposed for different forms of perturbation, the limitation still exists, which will be discussed as follows. As for the patch-based, it is common sense that the large size of the patch, the better the attack performance \cite{brown2017adversarial,lee2019physical}, which is expected. However, the large patch easily appeals to the human observer and results in a failed attack, while a small patch leads to unsatisfying attack performance. As for the texture-based, existing methods are prone to optimize conspicuous repeat adversarial patterns (mosaic-like) \cite{zhang2019camou,wu2020physical,suryanto2022dta}, which is unnatural. Furthermore, the manufacture of adversarial texture in the real world also impacts the attack performance as some mismatch between the made texture and the digital texture in size or color \cite{wang2021dual,wang2022fca}. As for the optical-based, one of the limitations of the existing method optical-based method is that the created light is easily influenced by environmental conditions, such as the strong sunlight would attenuate the created adversarial light \cite{huang2022spaa} and the moist air disturbs the light. Despite some work attempts to leverage the natural phenomenon (e.g., shadow \cite{zhong2022shadows} and reflected light \cite{wang2023rfla}) to perform physical attacks to avoid the environmental light impact, they may incur deformation when they produce the perturbation toward the target object during the deployment stage.

Some potential research directions can be used to solve the above issues. First, for the patch-based attack, the current methods are mainly focused on optimizing a rectangle shape patch, while developing the deformable patch may lead to better attack performance by replacing smaller regions of the target object. Moreover, another possible approach is to make the large patch as well as the texture more natural without satisfying attack performance; for example, a promising technique is style transfer and task-specific design. Second, for the texture-based attack, the mismatch of the manufactured texture and digital texture is mainly caused by the current renderer technique unsupporting the invertible renderer, resulting in unable to optimize the texture of the 3D object directly. Thus, a novel invertible renderer can optimize the texture of the 3D object directly is expected, then the texture can be made out according to the actual size of the 3D object. Finally, for the optical-based attack, a potential solution to solve the deformable during deployment is to consider the interaction (e.g., position, angular offset) of the light source and target object during optimization.

\subsubsection{Generalization of physical adversarial attack}
Generalization of the physical adversarial attack involves many crucial problems, including but not limited to developing practical physical attacks, improving the transferability of physical attacks, and breaking the actual deployed system. The many existing physical attacks belong to white-box attacks, but the deployed systems in the real world are inaccessible, which makes the existing white-box approaches invalid. Moreover, the image-specific perturbation is also unsuitable in practice. Furthermore, the actual deployed system may contain many components (e.g., multi-modal and multi-task model), while current physical attacks only focus on the specific modal or task.

Based on the aforementioned discussion, the potential research directions of physical adversarial attack include:

\textbf{Universal physical attacks under the black-box setting.} Although a line of black-box attacks has been proposed in the digital world, less explored in the physical world as there is a disconnect between the optimization of perturbation and performing a physical attack against the deployed system, which indicates that the system would not return any beneficial feedback. Therefore,  developing a practical physical attack is a challenging task. Moreover, the perturbation used to perform the physical attack should be universal, while only one approach attempt to optimize the universal perturbation under the black-box setting in the digital world \cite{ghosh2022black}.

\textbf{Transferability of physical attacks.} The transferability of physical attacks means that perturbation generated for one model can be used to attack other models without hurting attack performance significantly. Therefore, it can be a substitute approach to black-box attacks. Some approaches have been used to improve the transferability of physical attack, such as attention map \cite{zhang2023boosting} and model-ensemble \cite{huang2023t}. However, the physical attack performance of such attacks is not reported. Thus, it is necessary to investigate whether transferability that can work in the digital world also works in the physical world.

\textbf{Physical attacks against multi-modal and multi-tasks.} Physical attacks against multi-modal and multi-tasks. Current attacks are mainly focused on attacking the specific modal or task, while the actual deployed system is usually composed of many components (i.e., DNN models). Therefore, developing the physical attack against multi-modal or multi-task is more reality in the real world.

\subsection{Evaluation}

\subsubsection{Evaluation benchmark dataset and tool}
Although most physical attack methods were proposed, they are devised for different datasets, making them hard to evaluate different methods fairly. Although some works release their own customized dataset \cite{huang2020universal,wang2021dual}, the quality and description are deficient. On the other hand, the real environment is complex and unable to perform the physical attack in the same weather and light conditions as the previous work, which makes it hard to reproduce the result reported in the publications.

Therefore, future work can consider the following aspect: 1) It is urgent to develop a high-quality and labeled dataset for evaluating the effectiveness of the attack methods. 2) Recently, some research has attempted to use the simulator to develop adversarial attacks \cite{wang2021dual,wang2022fca,zhang2023boosting}, which gives a promising direction to develop an environment for evaluating physical attacks by using the simulator as it can provide reproducibility and definite environments by fixing the scene and parameters (e.g., weather and light conditions). For example, one can develop a physical testing environment using an open source simulator (e.g., Carla\cite{Dosovitskiy17} or AirSim \cite{airsim2017fsr}), then provide the physical testing interface, allowing the attacker to call the testing procedure with the optimized adversarial perturbation and report the attack results.

\subsubsection{Uniform physical test criterion}
Most physical attacks conduct physical experiments under stationary conditions, which means that the adversary takes photos of the adversarial perturbation disturbed object at a specific position and then evaluates the attack performance of the captured image. However, the target object moves in and out of the viewfinder of the system's sensor device in the actual physical environment, meaning that the ratio of the perturbed object is dynamically changing with respect to the fixed viewfinder. In the dynamically changing process, many factors caused by the background environment would impact the attack performance. The stationary test cannot include such a complex scenario. To this end, the dynamic physical test may better represent the performance of the attack method in the real world. 

In future work, uniform physical test criteria should be made under stationary and dynamic settings, including the print resolution of adversarial perturbation, environment conditions (e.g., weather and lighting), and the distance and position of the sampling camera to the target perturb object and so on (e.g., the moving speed of the camera or the target object in dynamic test).

\subsection{Potential positive applications}
The adversarial attack is a double-blade sword. On the one side, adversarial attacks would lead the DNN-based system to make mistakes, resulting in potential security risks. On the other side, adversarial attacks also have their benefits side, such as 1) the adversarial perturbed physical adversarial example could be used as the training data to boost the adversarial robustness of DNN further; 2) adversarial perturbation can be used to make the DNN system remain stable in complex environmental conditions \cite{salman2021unadversarial,wang2022defensive}. Therefore, potential applications exploiting adversarial perturbation to boost the performance of DNN in the real world is an interesting research direction.

\section{Conclusion}
\label{sec:conclude}

In this paper, we systemically discussed the physical adversarial attacks in computer vision tasks, including image recognition and object detector tasks. We argue that DNN-based systems still face enormous potential risks in both digital and physical environments. These risks should be thoroughly studied to avoid unnecessary loss as widely deployed in the real world. Specifically, we proposed a taxonomy scheme to categorize and summarize the current physical adversarial attack. By reviewing the existing physical adversarial attack, we summarize the common characteristic of physical adversarial attacks, which may be helpful to future physical adversarial attack research. Finally, we discuss the problems that need to be solved in physical adversarial attacks and provide some possible research directions. 

\section{Acknowledgments}
This work was supported by the National Natural Science Foundation of China (No.11725211 and 52005505).

\bibliographystyle{IEEEtran}.

\bibliography{references}

\end{document}